\documentclass[10pt,onecolumn,journal,compsoc]{IEEEtran}

\usepackage[OT1]{fontenc} 
\usepackage{adjustbox}
\usepackage[ruled, linesnumbered]{algorithm2e}
\usepackage{amsfonts}       
\usepackage{amsmath}
\usepackage{amssymb}
\usepackage{amsthm}
\usepackage{bbm}
\usepackage{bm}
\usepackage{booktabs}       
\usepackage[font=small]{caption}
\usepackage{color}
\usepackage[noadjust]{cite}

\usepackage{enumitem} 
\usepackage{float}
\usepackage{grffile} 
\usepackage{graphicx}
\usepackage{graphics}
\usepackage{microtype}      
\usepackage{makecell}
\usepackage{multirow}
\usepackage[subrefformat=parens,labelformat=parens]{subfig}
\usepackage{url}            
\usepackage[table]{xcolor} 
\usepackage{xparse}
\usepackage{colortbl}

\ExplSyntaxOn
\NewDocumentCommand{\longdash}{ O{2} }
 {
  --\prg_replicate:nn { #1 - 1 } { \negthinspace -- }
 }
\ExplSyntaxOff

\usepackage{hyperref}       

\newcommand\scalemath[2]{\scalebox{#1}{\mbox{\ensuremath{\displaystyle #2}}}}  

\newtheorem{theorem}{Theorem}
\newtheorem{lemma}{Lemma}
\theoremstyle{definition}
\newtheorem{definition}{Definition}

\newtheorem{remark}{Remark}

\allowdisplaybreaks 

\begin{document}
	
	
 	\title{Continuous Conditional Generative Adversarial Networks: Novel Empirical Losses and Label Input Mechanisms}

	\author{Xin~Ding,
    		Yongwei Wang,~\IEEEmembership{Student Member,~IEEE,}
    		Zuheng Xu,\\
    		William~J.~Welch, 
    		and~Z.~Jane~Wang,~\IEEEmembership{Fellow,~IEEE}
    		\IEEEcompsocitemizethanks{\IEEEcompsocthanksitem Xin Ding, Zuheng Xu and William J.\ Welch are with the Department of Statistics, University of British Columbia, Vancouver, BC, V6T 1Z4 Canada (e-mail: xin.ding@stat.ubc.ca, zuheng.xu@stat.ubc.ca, will@stat.ubc.ca) .\protect%
    		\IEEEcompsocthanksitem Yongwei Wang and Z.\ Jane Wang are with the Department of Electrical and Computer Engineering, University of British Columbia, Vancouver, BC, V6T 1Z4 Canada (e-mail: yongweiw@ece.ubc.ca, zjanew@ece.ubc.ca). \textit{(Corresponding author: Yongwei Wang)}}
	}
	\markboth{ACCEPTED BY IEEE TRANSACTIONS ON PATTERN ANALYSIS AND MACHINE INTELLIGENCE}%
	{Shell \MakeLowercase{\textit{et al.}}: Bare Demo of IEEEtran.cls for Computer Society Journals}

	\IEEEtitleabstractindextext{%
		\begin{abstract}

		This paper focuses on \textit{conditional generative modeling} (CGM) for image data with continuous, scalar conditions (termed regression labels). We propose the first model for this task which is called \textit{continuous conditional generative adversarial network} (CcGAN). Existing \textit{conditional GANs} (cGANs) are mainly designed for categorical conditions (e.g., class labels). Conditioning on regression labels is mathematically distinct and raises two fundamental problems: (P1) since there may be very few (even zero) real images for some regression labels, minimizing existing empirical versions of cGAN losses (a.k.a. empirical cGAN losses) often fails in practice; and (P2) since regression labels are scalar and infinitely many, conventional label input mechanisms (e.g., combining a hidden map of the generator/discriminator with a one-hot encoded label) are not applicable. We solve these problems by: (S1) reformulating existing empirical cGAN losses to be appropriate for the continuous scenario; and (S2) proposing a \textit{naive label input} (NLI) mechanism and an \textit{improved label input} (ILI) mechanism to incorporate regression labels into the generator and the discriminator.  The reformulation in (S1) leads to two novel empirical discriminator losses, termed the \textit{hard vicinal discriminator loss} (HVDL) and the \textit{soft vicinal discriminator loss} (SVDL) respectively, and a novel empirical generator loss. Hence, we propose four versions of CcGAN employing different proposed losses and label input mechanisms. The error bounds of the discriminator trained with HVDL and SVDL, respectively, are derived under mild assumptions. To evaluate the performance of CcGANs, two new benchmark datasets (RC-49 and Cell-200) are created. A novel evaluation metric (\textit{Sliding Fr\'echet Inception Distance}) is also proposed to replace Intra-FID when Intra-FID is not applicable. Our extensive experiments on several benchmark datasets (i.e., RC-49, UTKFace, Cell-200, and Steering Angle with both low and high resolutions) support the following findings: the proposed CcGAN is able to generate diverse, high-quality samples from the image distribution conditional on a given regression label; and CcGAN substantially outperforms cGAN both visually and quantitatively.  

		\end{abstract}
		
		}

	\maketitle
	
	\IEEEdisplaynontitleabstractindextext
	
	\IEEEpeerreviewmaketitle

	\IEEEraisesectionheading{\section{Introduction}\label{sec:introduction}}
	
	\textit{Conditional generative adversarial networks} (cGANs), first proposed in \cite{mirza2014conditional}, aim to estimate the distribution of images conditioning on some auxiliary information (a.k.a. \textit{conditional generative modeling} (CGM) for image data), especially class labels. Subsequent studies \cite{odena2017conditional, miyato2018cgans, brock2018large, SAGAN-zhang19d} confirm the feasibility of generating diverse, high-quality (even photo-realistic), and class-label consistent fake images from well-trained class-conditional GANs. Unfortunately, existing cGANs are not applicable for CGM with continuous, scalar conditions, termed \textit{regression labels}, due to two problems:
	
	{\setlength{\parindent}{0cm}\textbf{(P1)}} cGANs are often trained to minimize the empirical versions of their losses (a.k.a. empirical cGAN losses) on some training data, a principle also known as \textit{empirical risk minimization} (ERM) \cite{vapnik2000nature, mohri2018foundations, shalevshwartz2014understanding}. The success of ERM relies on a large sample size for each distinct condition. Unfortunately, we usually have only a few real images for some regression labels. Moreover, since regression labels are continuous, some values may not even appear in the training set. Consequently, a cGAN cannot accurately estimate the image distribution conditional on such missing labels.
	
	{\setlength{\parindent}{0cm}\textbf{(P2)}} In the class-conditional generative modeling, class labels are often encoded by one-hot vectors or label embedding and then fed into the generator and discriminator by hidden concatenation \cite{mirza2014conditional}, an auxiliary classifier \cite{odena2017conditional} or label projection \cite{miyato2018cgans}. A precondition for such label encoding is that the number of distinct labels (e.g., the number of classes) is finite and known. Unfortunately, in the continuous scenario, we may have infinitely many distinct regression labels.

	A naive approach denoted by \textbf{cGAN ($\bm{K}$ classes)} to solve \textbf{(P1)-(P2)} is to ``bin'' the regression labels into $K$ disjoint intervals and still train a cGAN in the class-conditional manner (these intervals are treated as independent classes) \cite{Greg2019}. Another naive approach denoted by \textbf{cGAN (concat)} for solving \textbf{(P2)} directly combines a regression label with the input or a hidden map of the generator and discriminator. However, the sampling results of both approaches in our empirical studies in Sections ~\ref{sec:experiment} and \ref{sec:experiment_hd} show two types of failures of conventional cGANs in CGM with regression labels: (1) cGAN ($\bm{K}$ classes) cannot generate visually realistic and diverse images; (2) cGAN (concat) fails to generate images with respect to conditioning regression labels.
 
	In machine learning, \textit{vicinal risk minimization} (VRM) \cite{vapnik2000nature, chapelle2001vicinal} is an alternative rule to ERM. VRM assumes that a sample point shares the same label with other samples in its vicinity. Motivated by VRM, in generative modeling conditional on regression labels where we estimate a conditional distribution $p(\bm{x}|y)$ ($\bm{x}$ is an image and $y$ is a regression label), it is natural to assume that a small perturbation to $y$ results in a negligible change to $p(\bm{x}|y)$. This assumption is consistent with our perception of the world. For example, the image distribution of facial features for a population of 15-year-old teenagers should be close to that of 16-year olds. 
	
	We therefore introduce the \textit{continuous conditional GAN} (CcGAN) to tackle \textbf{(P1)} and \textbf{(P2)}. To our best knowledge, this is the first generative model for image data conditional on regression labels. It is noted that \cite{semiGAN} and \cite{2018reg} train GANs in an unsupervised manner and synthesize unlabeled fake images for a subsequent image regression task. \cite{olmschenk2019dense} proposes a semi-supervised GAN for dense crowd counting. \cite{dosovitskiy2016learning} uses a \textit{convolutional neural network} (CNN) to generate images of objects in terms of some high-level parameters such as object style, viewpoint, color, brightness, etc. \cite{infogan2016} proposes InfoGAN, which can control some continuous or discrete attributes of generated images. Some text-to-image generation methods \cite{yan2016attribute2image, NEURIPS2019_1d72310e, tao2020df} train generative models conditional on high-dimensional attribute vectors with continuous or discrete elements. The objectives of these works are entirely different from ours since they do not aim to estimate the image distribution conditional on regression labels. Moreover, some recent works \cite{zhao2020differentiable, karras2020training, tran2020data, zhao2020image} propose several novel schemes to train GANs when training data are limited, which seems to be relevant to \textbf{(P1)}. However, they are also fundamentally different from CcGAN, since they are designed for unconditional and class-conditional scenarios rather than continuous ones. Our contributions can be summarized as follows:
	
	\begin{itemize}[leftmargin=0.4cm]
		
		\item We propose in Section \ref{sec:solution_1} a solution to address \textbf{(P1)}, which consists of two novel empirical discriminator losses, termed the \textit{hard vicinal discriminator loss} (HVDL) and the \textit{soft vicinal discriminator loss} (SVDL), and a novel empirical generator loss. We take the vanilla cGAN loss as an example to show how to derive HVDL, SVDL, and the novel empirical generator loss by reformulating existing empirical cGAN losses.
		
		\item In Section \ref{sec:solution_2}, we propose two novel label input mechanisms, consisting of a \textit{naive label input} (NLI) mechanism and an \textit{improved label input} (ILI) mechanism, as solutions to address \textbf{(P2)}.
		
		\item We derive in Section \ref{sec:convergence_rate} the error bounds of a discriminator trained with HVDL and SVDL. These error bounds not only help us understand how HVDL and SVDL influence the discriminator training but also guide our implementation in practice (especially the selection of hyper-parameters).
		
		\item We propose in Section \ref{sec:sfid} a novel evaluation metric, termed \textit{Sliding Fr\'echet Inception Distance} (SFID), to evaluate the generative image modeling conditional on regression labels when there are insufficient real images to compute Intra-FID \cite{miyato2018cgans}. 
		
		\item In Sections \ref{sec:experiment} and \ref{sec:experiment_hd}, we propose two new benchmark datasets, RC-49 and Cell-200, for generative image modeling conditional on regression labels, since very few benchmark datasets are suitable for the studied continuous scenario. We conduct extensive experiments on four benchmark datasets with various resolutions (from $64\times 64$ to $256\times 256$) to demonstrate that CcGAN not only generates diverse, high-quality, and label consistent images, but also substantially outperforms cGAN both visually and quantitatively. The effectiveness of SFID is also studied on the RC-49 dataset at the end of Section \ref{sec:experiment}.
	\end{itemize}
	
	A preliminary version of this paper has been presented at the International Conference on Learning Representations \cite{ding2021ccgan}. The current paper strengthens the initial version in several ways. (1) We propose an improved label input (ILI) mechanism to better incorporate regression labels into CcGAN. Our experiments in this paper demonstrate the superiority of ILI to the naive label input method in \cite{ding2021ccgan}. (2) We introduce Lemmas~\ref{lem:conditional_mean_bound_HVDL} and \ref{lem:conditional_mean_bound_SVDL}, which are used to derive the error bounds of HVDL and SVDL but are omitted in \cite{ding2021ccgan}. The motivation for deriving these error bounds is also better illustrated, and an improved proof for the derivation is provided in Appendix. (3) We propose SFID to replace Intra-FID when Intra-FID is not applicable. (4) We create a new benchmark dataset called Cell-200 for generative modeling conditional on regression labels. (5) We conduct a more extensive empirical study to demonstrate the effectiveness of CcGAN. This extensive study includes more datasets (e.g., Cell-200 and Steering Angle) and more complicated settings (e.g., various image resolutions). We also add a new baseline, cGAN (concat), to the comparison to better demonstrate that conventional cGANs are inapplicable to our task.

	\section{From cGAN to CcGAN}\label{sec:cGAN2CcGAN}
\begin{figure*}[!htbp]
		\centering
		\includegraphics[width=0.65\linewidth]{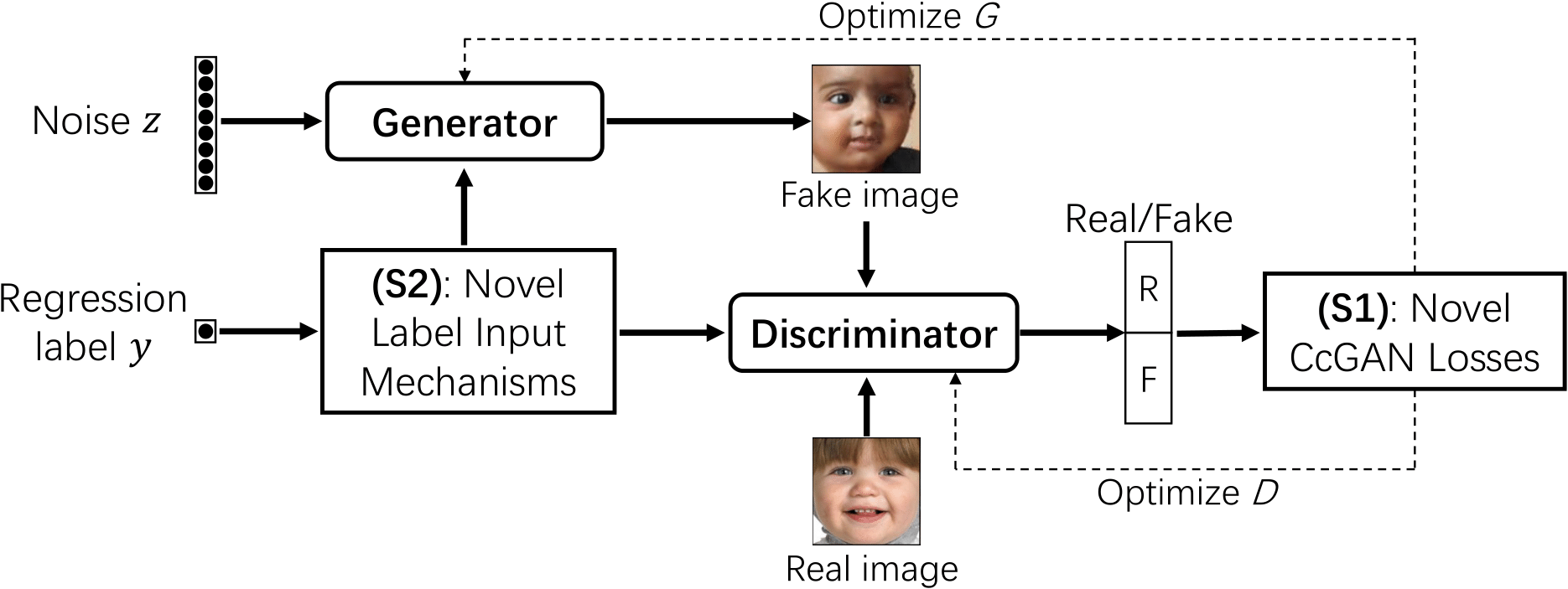}
		\caption{{\color{black} \textbf{A typical workflow of the proposed CcGAN framework.} A regression label $y$ is input into the generator ($G$) and the discriminator ($D$) by novel label input mechanisms proposed in Section \ref{sec:solution_2}. Novel empirical losses proposed in Section \ref{sec:solution_1} are adopted to optimize $G$ and $D$, respectively. The CcGAN framework is compatible with modern GAN architectures (e.g., SNGAN \cite{miyato2018spectral}, SAGAN \cite{SAGAN-zhang19d}) and advanced GAN training techniques (e.g., DiffAugment \cite{zhao2020differentiable}).}}
		\label{fig:workflow_CcGAN}
	\end{figure*}

	In this section, we introduce the \textit{continuous conditional GAN} (CcGAN), consisting of solutions to \textbf{(P1)} and \textbf{(P2)}. The combinations of two vicinal discriminator losses (HVDL and SVDL) proposed in Section \ref{sec:solution_1} and two novel label input mechanisms (NLI and ILI) proposed in Section \ref{sec:solution_2} result in four CcGAN methods denoted by HVDL+NLI, SVDL+NLI, HVDL+ILI, and SVDL+ILI, respectively. The overall workflow of CcGAN is visualized in Fig.\ \ref{fig:workflow_CcGAN}.
	
	\subsection{Solution to (P1): Reformulated Empirical Losses }\label{sec:solution_1}
	Theoretically, cGAN losses (e.g., the vanilla cGAN loss \cite{mirza2014conditional}, the Wasserstein loss \cite{pmlr-v70-arjovsky17a, gulrajani2017improved}, and the hinge loss \cite{miyato2018spectral}) are suitable for both class labels and regression labels; however, their empirical versions fail in the continuous scenario (i.e., \textbf{(P1)}). Our first solution \textbf{(S1)} focuses on reformulating these empirical cGAN losses for continuous labels. Without loss of generality, we only take the vanilla cGAN loss as an example to show such reformulation (the empirical versions of the Wasserstein loss and the hinge loss can be reformulated similarly). 
	
	The vanilla discriminator loss and generator loss \cite{mirza2014conditional} are defined as:
	\begin{align}
		\mathcal{L}(D) = & -\mathbb{E}_{y\sim p_r(y)}\left[\mathbb{E}_{\bm{x}\sim p_r(\bm{x}|y)}\left[ \log \left( D(\bm{x},y) \right) \right] \right] \nonumber \\
		 &  - \mathbb{E}_{y\sim p_g(y)}\left[\mathbb{E}_{\bm{x}\sim p_g(\bm{x}|y)}\left[ \log \left( 1 - D(\bm{x},y) \right) \right] \right] \nonumber \\
		= & -\int \log(D(\bm{x},y))p_r(\bm{x},y) d\bm{x}dy \nonumber \\
		& - \int \log(1-D(\bm{x},y))p_g(\bm{x},y)d\bm{x}dy, \label{eq:loss_cD} \\
		\mathcal{L}(G) = &-\mathbb{E}_{y\sim p_g(y)}\left[\mathbb{E}_{\bm{z}\sim q(\bm{z})}\left[ \log \left( D(G(\bm{z},y),y) \right) \right] \right] \nonumber \\
		= & -\int \log(D(G(\bm{z},y),y)) q(\bm{z})p_g(y) d\bm{z}dy, \label{eq:loss_cG}
	\end{align}
	where $\bm{x}\in\mathcal{X}$ is an image, $y\in\mathcal{Y}$ is a label, $p_r(y)$ and $p_g(y)$ are respectively the actual and fake label marginal distributions, $p_r(\bm{x}|y)$ and $p_g(\bm{x}|y)$ are respectively the actual and fake image distributions conditional on $y$, $p_r(\bm{x}, y)$ and $p_g(\bm{x}, y)$ are respectively the actual and fake joint distributions of $\bm{x}$ and $y$, and $q(\bm{z})$ is the probability density function of $\mathcal{N}(\bm{0},\bm{I})$.
	
	Since the distributions in the losses of Eqs.\ \eqref{eq:loss_cD} and \eqref{eq:loss_cG} are unknown, for class-conditional generative modeling, \cite{mirza2014conditional} follows ERM and minimizes the empirical losses:
	\begin{equation}
		\label{eq:loss_cD_emp}
		\begin{aligned}
			\widehat{\mathcal{L}}^{\delta}(D)  = &-\frac{1}{N^r}\sum_{c=1}^C \sum_{j=1}^{N_c^r}\log(D(\bm{x}_{c,j}^r, c)) \\ 
			&  - \frac{1}{N^g}\sum_{c=1}^C\sum_{j=1}^{N_c^g}\log(1-D(\bm{x}_{c,j}^g, c)),
		\end{aligned}
	\end{equation}

	\begin{equation}
		\label{eq:loss_cG_emp}
		\widehat{\mathcal{L}}^{\delta}(G) = - \frac{1}{N^g}\sum_{c=1}^C\sum_{j=1}^{N_c^g}\log (D(G(\bm{z}_{c,j}, c), c)),
	\end{equation}
	where $C$ is the number of classes, $N^r$ and $N^g$ are respectively the number of real and fake images, $N_c^r$ and $N_c^g$ are respectively the number of real and fake images with label $c$, $\bm{x}_{c,j}^r$ and $\bm{x}_{c,j}^g$ are respectively the $j$-th real image and the $j$-th fake image with label $c$,  and the $\bm{z}_{c,j}$ are independently and identically sampled from $q(\bm{z})$. Eq.\ \eqref{eq:loss_cD_emp} implies we estimate $p_r(\bm{x},y)$ and $p_g(\bm{x}, y)$ by their empirical probability density functions as follows:
	\begin{equation}
		\label{eq:joint_emp_esti}
		\begin{aligned}
			\hat{p}_r^{\delta}(\bm{x},y) & = \frac{1}{N^r}\sum_{c=1}^C\sum_{j=1}^{N_c^r}\delta(\bm{x}-\bm{x}_{c,j}^r)\delta(y-c),\\
			\hat{p}_g^{\delta}(\bm{x},y) & = \frac{1}{N^g}\sum_{c=1}^C\sum_{j=1}^{N_c^g}\delta(\bm{x}-\bm{x}_{c,j}^g)\delta(y-c),
		\end{aligned}
	\end{equation}
	where $\delta(\cdot)$ is a Dirac delta function (Appendix A of \cite{salasnich2014quantum}) centered at 0. However, $\hat{p}_r^{\delta}(\bm{x},y)$ and $\hat{p}_g^{\delta}(\bm{x},y)$ are not good estimates in the continuous scenario because of \textbf{(P1)}.
	
\begin{figure*}[!htbp]
		\captionsetup[subfloat]{farskip=2pt,captionskip=1pt}
		\centering
		\subfloat[Hard Vicinity]{\includegraphics[width=0.45\textwidth]{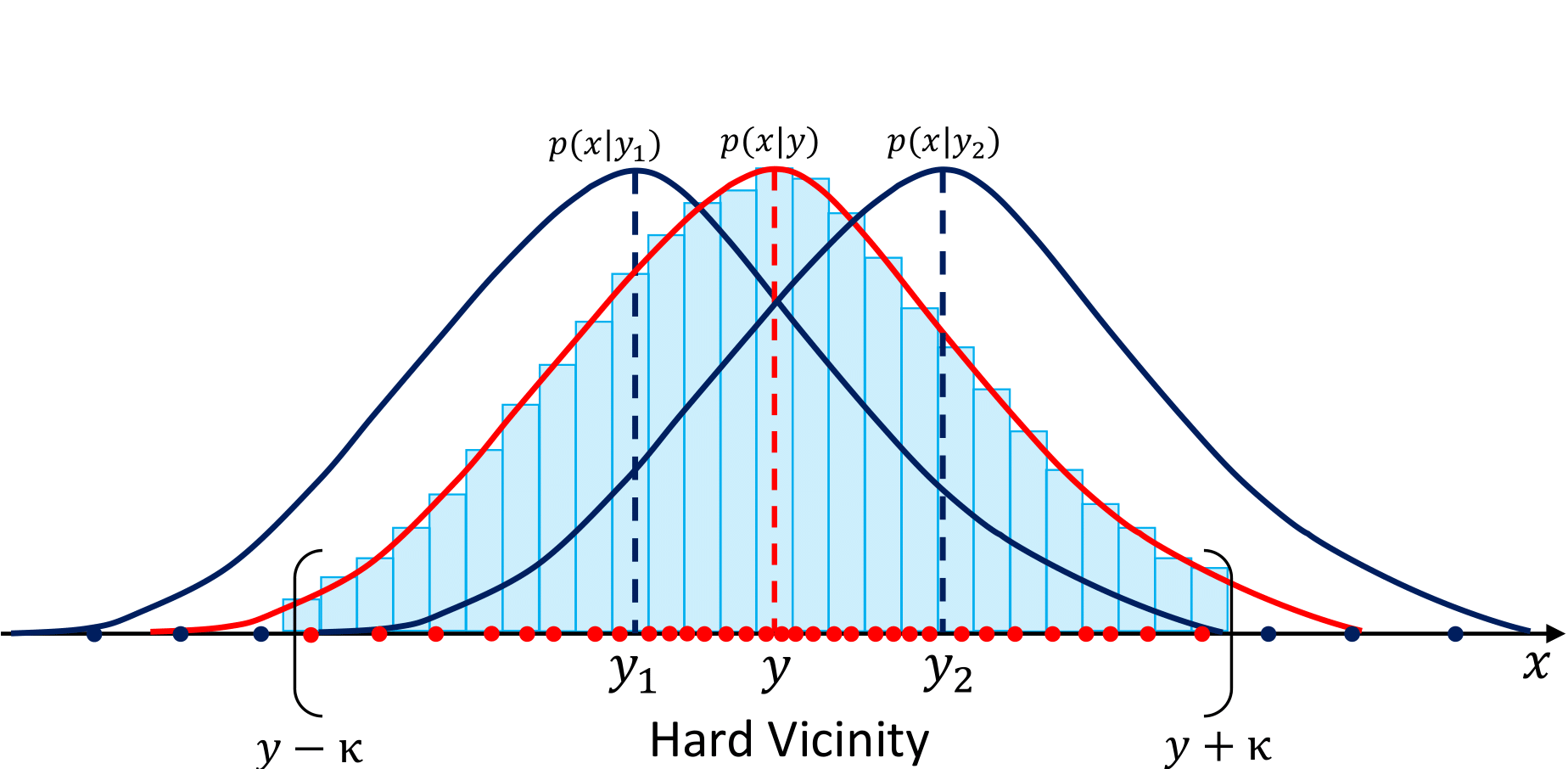}\label{fig:visualization_HVE}}  
		\hspace{2mm}
		\subfloat[Soft Vicinity]{\includegraphics[width=0.45\textwidth]{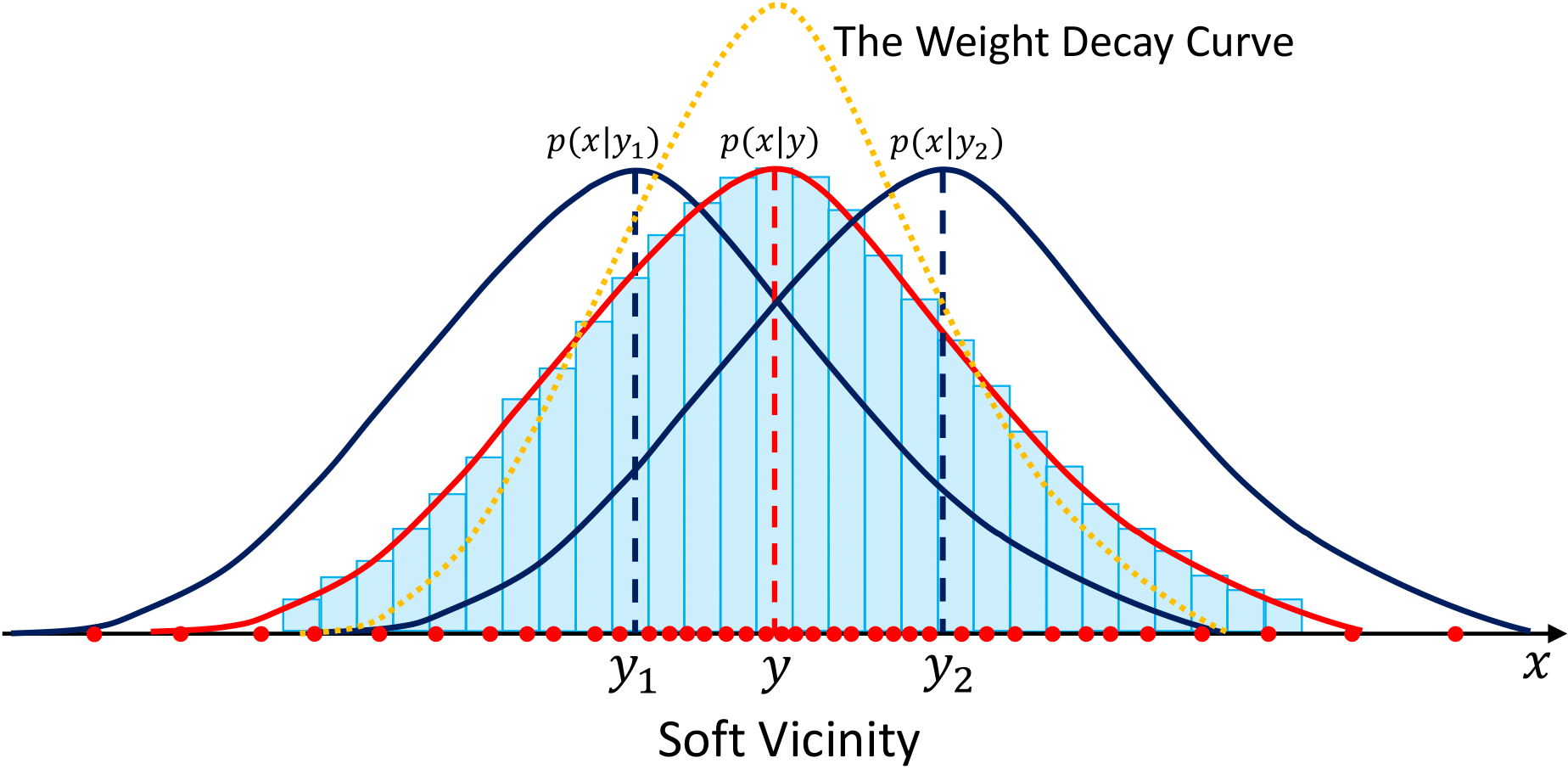}\label{fig:visualization_SVE}}   
		\caption{ HVE (Eqs.\ \eqref{eq:HVE_joint_estiamte_dist_real} and \eqref{eq:HVE_joint_estiamte_dist_fake}) and SVE (Eqs.\ \eqref{eq:SVE_joint_estiamte_dist_real} and \eqref{eq:SVE_joint_estiamte_dist_fake}) estimate $p(\bm{x}|y)$ (a univariate Gaussian conditional on $y$) by using two samples in hard and soft vicinities, respectively, of $y$. To estimate $p(\bm{x}|y)$ (the red Gaussian curve) only from samples drawn from $p(\bm{x}|y_1)$ and $p(\bm{x}|y_2)$ (the blue Gaussian curves), estimation is based on the samples (red dots) in a hard vicinity (defined by $y\pm \kappa$) or a soft vicinity (defined by the weight decay curve) around $y$. The histograms in blue are samples in the hard or soft vicinity. The labels $y_1$, $y$, and $y_2$ on the $x$-axis denote the means of $\bm{x}$ conditional on $y_1$, $y$, and $y_2$, respectively. }
		\label{fig:visualization_HVE_and_SVE}
	\end{figure*}

	To overcome \textbf{(P1)}, we propose a novel estimate for each of $p_r(\bm{x},y)$ and $p_g(\bm{x},y)$, termed the \textit{hard vicinal estimate\/} (HVE). We also provide an intuitive alternative to HVE, named the \textit{soft vicinal estimate\/} (SVE). The HVEs of $p_r(\bm{x},y)$ and $p_g(\bm{x},y)$ are:
	\begin{align}
		\hat{p}_r^{\text{HVE}}(\bm{x},y) = & C_1\cdot\left[\frac{1}{N^r}\sum_{j=1}^{N^r}\exp{\left(-\frac{(y-y_j^r)^2}{2\sigma^2}\right)} \right] \nonumber \\
		& \cdot \left[ \frac{1}{N^r_{y,\kappa}}\sum_{i=1}^{N^r}\mathbbm{1}_{\{|y-y_i^r|\leq\kappa\}}\delta(\bm{x}-\bm{x}_i^r) \right], \label{eq:HVE_joint_estiamte_dist_real} \\
		\hat{p}_g^{\text{HVE}}(\bm{x},y) = & C_2\cdot\left[\frac{1}{N^g}\sum_{j=1}^{N^g}\exp{\left(-\frac{(y-y_j^g)^2}{2\sigma^2}\right)} \right] \nonumber\\
		& \cdot \left[ \frac{1}{N^g_{y,\kappa}}\sum_{i=1}^{N^g}\mathbbm{1}_{\{|y-y_i^g|\leq\kappa\}}\delta(\bm{x}-\bm{x}_i^g) \right], \label{eq:HVE_joint_estiamte_dist_fake}
	\end{align}
    where $\bm{x}_i^r$ and $\bm{x}_i^g$ are respectively real image $i$ and fake image $i$, $y_i^r$ and $y_i^g$are respectively the labels of $\bm{x}_i^r$ and $\bm{x}_i^g$, $\kappa$ and $\sigma$ are two positive hyper-parameters, $C_1$ and $C_2$ are two constants making these two estimates valid probability density functions, $N^r_{y,\kappa}$ is the number of the $y_i^r$ satisfying $|y-y_i^r|\leq\kappa$, $N^g_{y,\kappa}$ is the number of the $y_i^g$ satisfying $|y-y_i^g|\leq\kappa$, and $\mathbbm{1}$ is an indicator function with support in the subscript. The terms in the first square brackets of $\hat{p}_r^{\text{HVE}}$ and $\hat{p}_g^{\text{HVE}}$ imply we estimate the marginal label distributions $p_r(y)$ and $p_g(y)$ by \textit{kernel density estimates} (KDEs) \cite{davis2011remarks, parzen1962estimation, silverman1986density, hastie2009elements}. The terms in the second square brackets are designed based on the assumption that a small perturbation to $y$ results in negligible changes to $p_r(\bm{x}|y)$ and $p_g(\bm{x}|y)$. If this assumption holds, we can use images with labels in a small vicinity of $y$ to estimate $p_r(\bm{x}|y)$ and $p_g(\bm{x}|y)$. The SVEs of $p_r(\bm{x},y)$ and $p_g(\bm{x},y)$ are:
	\begin{align}
		\hat{p}_r^{\text{SVE}}(\bm{x},y) = & C_3\cdot\left[\frac{1}{N^r}\sum_{j=1}^{N^r}\exp{\left(-\frac{(y-y_j^r)^2}{2\sigma^2}\right)} \right] \nonumber\\
		& \cdot \left[\frac{\sum_{i=1}^{N^r}w^r(y_i^r,y)\delta(\bm{x}-\bm{x}_i^r)}{\sum_{i=1}^{N^r}w^r(y_i^r,y)} \right], \label{eq:SVE_joint_estiamte_dist_real}\\
		\hat{p}_g^{\text{SVE}}(\bm{x},y) = & C_4\cdot\left[\frac{1}{N^g}\sum_{j=1}^{N^g}\exp{\left(-\frac{(y-y_j^g)^2}{2\sigma^2}\right)} \right] \nonumber\\
		& \cdot \left[ \frac{\sum_{i=1}^{N^g}w^g(y_i^g,y)\delta(\bm{x}-\bm{x}_i^g)}{\sum_{i=1}^{N^g}w^g(y_i^g,y)} \right], \label{eq:SVE_joint_estiamte_dist_fake} 
	\end{align}
	where $C_3$ and $C_4$ are two constants making these two estimates valid probability density functions, 
	\begin{equation}
		\label{eq:SVE_weights}
		w^r(y_i^r, y) =  e^{-\nu(y_i^r-y)^2} \quad \text{and} \quad  w^g(y_i^g, y)  =  e^{-\nu(y_i^g-y)^2},
	\end{equation}
	and the hyper-parameter $\nu>0$. In Eqs.\ \eqref{eq:SVE_joint_estiamte_dist_real} and \eqref{eq:SVE_joint_estiamte_dist_fake}, similar to the HVEs, we estimate $p_r(y)$ and $p_g(y)$ by KDEs. Instead of using samples in a hard vicinity, the SVEs use all respective samples to estimate $p_r(\bm{x}|y)$ and $p_g(\bm{x}|y)$ but each sample is assigned a weight based on the distance of its label from $y$. Two diagrams in Fig.\ \ref{fig:visualization_HVE_and_SVE} visualize the process of using hard/soft vicinal samples to estimate the Gaussian distribution $p(\bm{x}|y)$ conditional on $y$, for univariate $\bm{x}$.

	By plugging Eq.\ \eqref{eq:HVE_joint_estiamte_dist_real}, \eqref{eq:HVE_joint_estiamte_dist_fake}, \eqref{eq:SVE_joint_estiamte_dist_real}, and \eqref{eq:SVE_joint_estiamte_dist_fake} into Eq.\ \eqref{eq:loss_cD}, we derive the \textit{hard vicinal discriminator loss} (HVDL) and the \textit{soft vicinal discriminator loss} (SVDL) as follows:
	\begin{equation}
		\label{eq:HVDL}
		\begin{aligned}
			& \widehat{\mathcal{L}}^{\text{HVDL}}(D) \\
			= & - \frac{C_5}{N^r}\sum_{j=1}^{N^r}\sum_{i=1}^{N^r}\mathbb{E}_{\epsilon^r\sim\mathcal{N}(0,\sigma^2)}\left[ W_1\log(D(\bm{x}_i^r, y_j^r+\epsilon^r)) \right]\\
			&- \frac{C_6}{N^g}\sum_{j=1}^{N^g}\sum_{i=1}^{N^g}\mathbb{E}_{\epsilon^g\sim\mathcal{N}(0,\sigma^2)}\left[ W_2\log(1-D(\bm{x}_i^g, y_j^g+\epsilon^g)) \right],
		\end{aligned}
	\end{equation}
	\begin{equation}
		\label{eq:SVDL}
		\begin{aligned}
			& \widehat{\mathcal{L}}^{\text{SVDL}}(D) \\
			= & - \frac{C_7}{N^r}\sum_{j=1}^{N^r}\sum_{i=1}^{N^r}\mathbb{E}_{\epsilon^r\sim\mathcal{N}(0,\sigma^2)}\left[W_3\log(D(\bm{x}_i^r, y_j^r+\epsilon^r)) \right]\\
			&- \frac{C_8}{N^g}\sum_{j=1}^{N^g}\sum_{i=1}^{N^g}\mathbb{E}_{\epsilon^g\sim\mathcal{N}(0,\sigma^2)}\left[ W_4\log(1-D(\bm{x}_i^g, y_j^g+\epsilon^g)) \right],
		\end{aligned}
	\end{equation}
	where $\epsilon^r\triangleq y-y_j^r$, $\epsilon^g\triangleq y-y_j^g$,
	\begin{align}
		& W_1=\frac{\mathbbm{1}_{\{|y_j^r+\epsilon^r-y_i^r|\leq\kappa\}}}{N_{y^r_j+\epsilon^r,\kappa}^r},\quad W_2=\frac{\mathbbm{1}_{\{|y_j^g+\epsilon^g-y_i^g|\leq\kappa\}}}{N_{y^g_j+\epsilon^g,\kappa}^g} \nonumber \\ 
		& 
		W_3=\frac{w^r(y_i^r,y_j^r+\epsilon^r)}{\sum_{i=1}^{N^r}w^r(y_i^r,y_j^r+\epsilon^r)},\quad
		W_4=\frac{w^g(y_i^g,y_j^g+\epsilon^g)}{\sum_{i=1}^{N^g}w^g(y_i^g,y_j^g+\epsilon^g)}, \nonumber
	\end{align}
	and $C_5$, $C_6$, $C_7$, and $C_8$ are some constants. 

	\textbf{Generator training:} The generator of CcGAN is trained by minimizing Eq. \eqref{eq:loss_G_CcGAN},
	\begin{equation}
		\label{eq:loss_G_CcGAN}
		\widehat{\mathcal{L}}^{\epsilon}(G) = - \frac{1}{N^g}\sum_{i=1}^{N^g}\mathbb{E}_{\epsilon^g\sim\mathcal{N}(0,\sigma^2)}\log (D(G(\bm{z}_{i}, y_i^g+\epsilon^g), y_i^g+\epsilon^g)).
	\end{equation}
	
	\textbf{How do HVDL, SVDL, and Eq.\ \eqref{eq:loss_G_CcGAN} overcome \textbf{(P1)}?} 
	
	{\setlength{\parindent}{0cm}\textbf{(i)}} Given a label $y$ as the condition, we use images in a hard/soft vicinity of $y$ to train the discriminator instead of just using images with label $y$. It enables us to estimate $p_r(\bm{x}|y)$ when there are not enough real images with label $y$.
	
	{\setlength{\parindent}{0cm}\textbf{(ii)}} From Eqs.\ \eqref{eq:HVDL} and \eqref{eq:SVDL}, we can see that the KDEs in Eqs.\ \eqref{eq:HVE_joint_estiamte_dist_real}, \eqref{eq:HVE_joint_estiamte_dist_fake}, \eqref{eq:SVE_joint_estiamte_dist_real}, and \eqref{eq:SVE_joint_estiamte_dist_fake} are adjusted by adding Gaussian noise to the labels. Moreover, in Eq.\ \eqref{eq:loss_G_CcGAN}, we add Gaussian noise to seen labels (assume the $y_i^g$ are seen) to train the generator to generate images at unseen labels. This enables estimation of $p_r(\bm{x}|y^\prime)$ when $y^\prime$ is not in the training set.

	\begin{remark}
		Based on the kernel density estimation \cite{davis2011remarks, parzen1962estimation, silverman1986density, hastie2009elements} and the property of the Dirac delta function (Appendix A of \cite{salasnich2014quantum}), $\int\delta(\bm{x}-\bm{x}^r_i)d\bm{x}=\int\delta(\bm{x}-\bm{x}^g_i)d\bm{x}=1$ and $C_1=C_2=C_3=C_4=1/\sigma$. Therefore, $C_5=C_6$ and $C_7=C_8$, which implies these constants $C_1,\ldots, C_8$ can be ignored when minimizing $\widehat{\mathcal{L}}^{\text{HVDL}}(D)$ and $\widehat{\mathcal{L}}^{\text{SVDL}}(D)$.
	\end{remark}
	
	\begin{remark}
		An algorithm is proposed in Supp. \ref{supp:algorithm_train} for training CcGAN in practice. Moreover, CcGAN does not require any specific network architecture, so it can use modern GAN architectures such as SNGAN \cite{miyato2018spectral}, SAGAN \cite{SAGAN-zhang19d} and BigGAN \cite{brock2018large}. CcGAN is also compatible with modern GAN training techniques such as DiffAugment \cite{zhao2020differentiable}. 
	\end{remark}

	\begin{remark}[A rule of thumb for hyper-parameter selection]
		\label{rmk:rule_of_thumb}
		In our experiments, we normalize labels to real numbers in $[0,1]$ and the hyper-parameter selection is conducted based on the normalized labels. To be more specific, the hyper-parameter $\sigma$ is computed based on a rule of thumb formula for the bandwidth selection of KDE \cite{silverman1986density}, i.e., $\sigma=\left({4\hat{\sigma}^5_{y^r}}/{3N^r}\right)^{1/5}$, where $\hat{\sigma}_{y^r}$ is the sample standard deviation of normalized labels in the training set. Let $\kappa_{\text{base}}=\max\left(y^r_{[2]}-y^r_{[1]}, y^r_{[3]}-y^r_{[2]}, \dots, y^r_{[N^r_{\text{uy}}]}-y^r_{[N^r_{\text{uy}}-1]} \right)$, where $y^r_{[l]}$ is the $l$-th smallest normalized distinct real label and $N^r_{\text{uy}}$ is the number of normalized distinct labels in the training set. Then $\kappa$ is set as a multiple of $\kappa_{\text{base}}$ (i.e., $\kappa=m_{\kappa}\kappa_{\text{base}}$) where the multiplier $m_{\kappa}$ stands for 50\% of the minimum number of neighboring labels used for estimating $p_r(\bm{x}|y)$ given a label $y$. For example, $m_{\kappa}=1$ implies using 2 neighboring labels (one on the left while the other one on the right). In our experiments, $m_{\kappa}$ is generally set as 1 or 2. In some extreme case when many distinct labels have too few real samples, we may consider increasing $m_{\kappa}$. We also found $\nu=1/\kappa^2$ works well in  (\ref{eq:SVE_weights}) in our experiments.
	\end{remark}

	\subsection{Solutions to (P2): Novel Label Input Mechanisms}\label{sec:solution_2} 
	In this section, we propose two solutions, consisting of a naive and an improved label input mechanism, to solve \textbf{(P2)}. 
	
\begin{figure}[!htbp]
		\centering
		\includegraphics[width=0.8\linewidth]{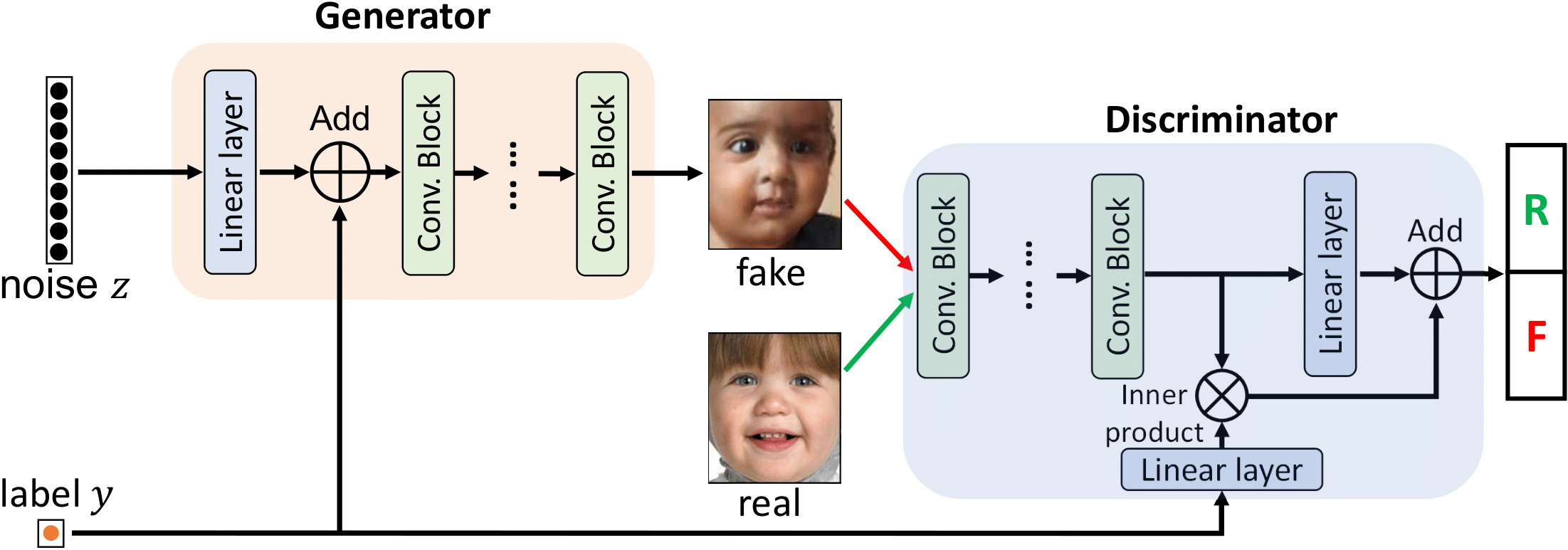}
		\caption{{\color{black} \textbf{The workflow of the naive label input (NLI) mechanism.} NLI inputs a regression label $y$ into $G$ by adding $y$ element-wise to the output of the first linear layer. NLI inputs $y$ into $D$ by label projection \cite{miyato2018cgans}.}}
		\label{fig:naive_label_input}
	\end{figure}

	{\setlength{\parindent}{0cm}\textbf{A naive label input (NLI) mechanism:}} We first propose a naive approach to incorporate the regression labels into the cGANs. For $G$, we add the label $y$ element-wise to the output of its first linear layer. For $D$, the label $y$ is first projected to the latent space learned by an extra linear layer. Then, we incorporate the embedded label into the discriminator by label projection \cite{miyato2018cgans}. Fig.\ \ref{fig:naive_label_input} visualizes the naive label input mechanism.   
	
\begin{figure}[!htbp]
		\centering
		\includegraphics[width=0.7\linewidth]{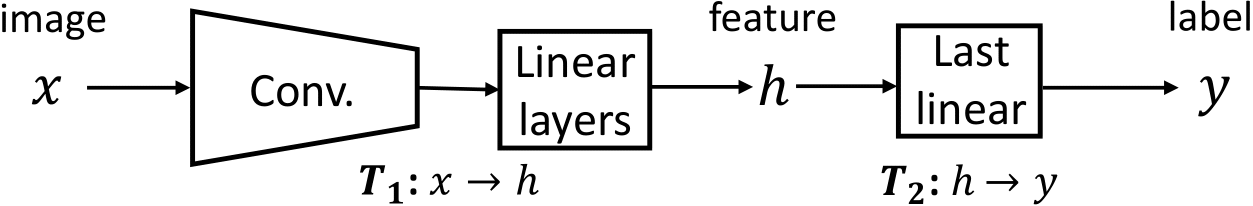}
		\caption{\textbf{The pre-trained CNN $T_1+T_2$ for label embedding.} The first subnetwork $T_1$ consists of some convolutional layers (Conv.) and some linear layers. The second subnetwork $T_2$ includes one linear layer.}
		\label{fig:pre_trained_CNN_for_label_embedding}
	\end{figure}

\begin{figure}[!htbp]
		\centering
		\includegraphics[width=0.5\linewidth]{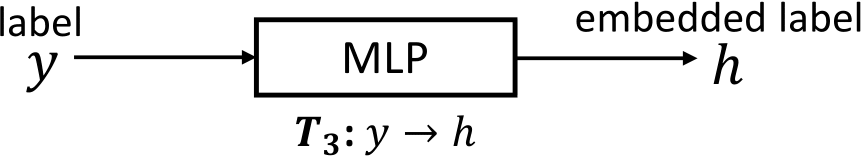}
		\caption{\textbf{The label embedding network is a multilayer perceptron (MLP).}}
		\label{fig:label_embedding_network}
	\end{figure}

\begin{figure}[!htbp]
		\centering
		\includegraphics[width=0.8\linewidth]{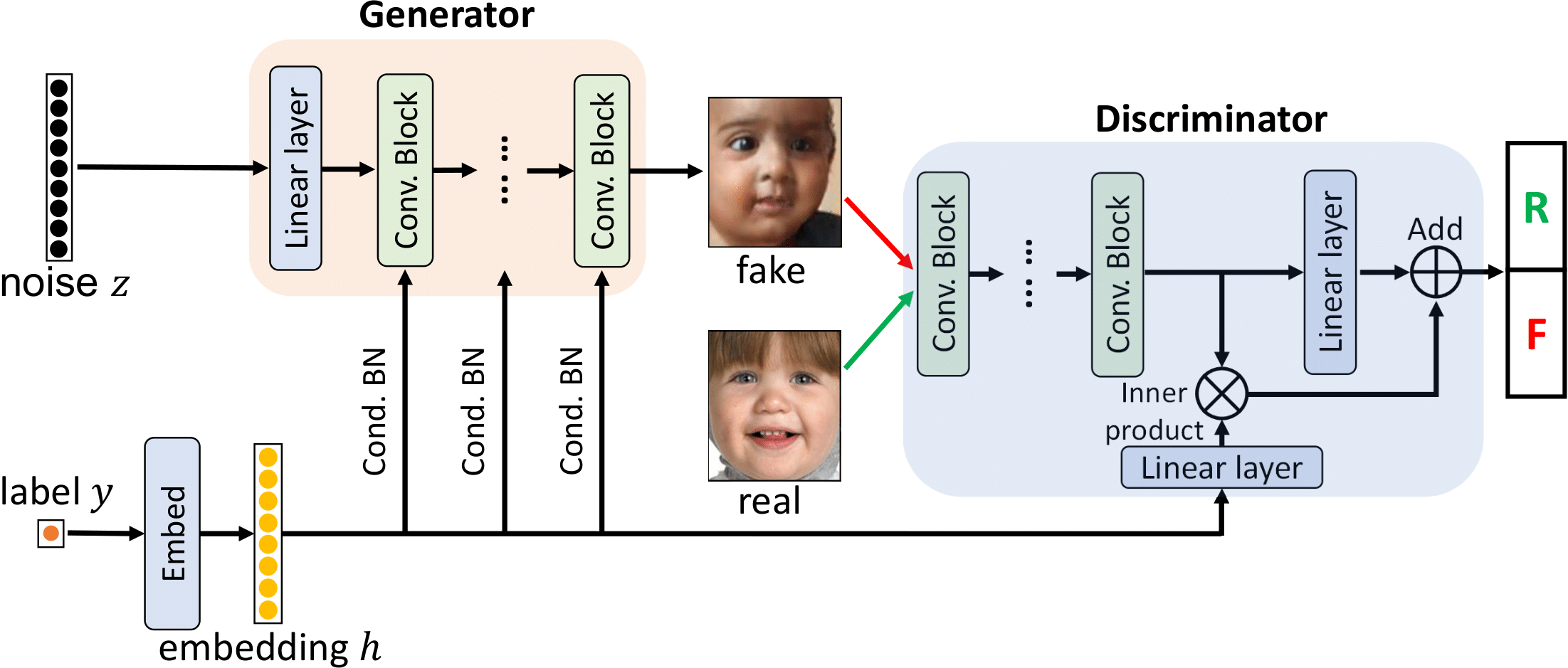}
		\caption{{\color{black} \textbf{The workflow of the improved label input (ILI) mechanism.} ILI first uses an embedding network to convert $y$ into its high-dimensional representation $h$. Then, $h$ is input into $G$ and $D$ by conditional batch normalization \cite{de2017modulating} and label projection \cite{miyato2018cgans}, respectively.}}
		\label{fig:improved_label_input}
	\end{figure}

	{\setlength{\parindent}{0cm}\textbf{An improved label input (ILI) mechanism:}} Empirical studies in Section \ref{sec:experiment} show that CcGAN with the naive label input mechanism already substantially outperforms cGAN. Nevertheless, it still suffers from severe label inconsistency on some datasets (e.g., Cell-200 and Steering Angle). To improve the label consistency of CcGAN, we propose an improved label input (ILI) mechanism. The ILI approach consists of a pre-trained CNN and a label embedding network. The pre-trained CNN, as shown in Fig. \ref{fig:pre_trained_CNN_for_label_embedding}, includes two subnetworks, $T_1$ and $T_2$, where $T_1$ maps an image $\bm{x}$ to a feature space and $T_2$ maps the extracted feature $\bm{h}$ to a regression label $y$. The dimension of the feature space is set to 128 in our experiments. The label embedding network $T_3$, as shown in Fig. \ref{fig:label_embedding_network}, is a multilayer perceptron (MLP) \cite{hastie2009elements} mapping a regression label $y$ back to its hidden representation $\bm{h}$ in the feature space defined by $T_1$. Assume that there are $m$ distinct regression labels in the training set, i.e., $y_1^u, y_2^u, \dots, y_m^u$, then the label embedding network $T_3$ is trained by:
	\begin{equation}
		\label{eq:label_embedding_loss}
        \min_{T_3}\frac{1}{m}\sum_{i=1}^{m}\mathbb{E}_{\gamma\sim\mathcal{N}(0,\sigma_{\gamma}^2)}\left[ (T_2(T_3(y_i^u+\gamma)) - (y_i^u+\gamma))^2 \right],
	\end{equation}
	where $\sigma_{\gamma}$ is often a small value and is set at 0.2 in this paper. Then, given a regression label $y$, we can evaluate $T_3(y)$ to get a unique hidden representation of $y$ which will be incorporated into CcGAN as the condition (visualized in Fig. \ref{fig:improved_label_input}). Specifically, for $G$, we input the embedded label by using conditional batch normalization \cite{de2017modulating}. For $D$, similar to the naive approach, we input the embedded label into $D$ by label projection \cite{miyato2018cgans}. 
	
	\section{Error Bounds of $D$ Trained With HVDL and SVDL} \label{sec:convergence_rate}  %
	
	In this section, we derive the error bounds of a discriminator $D$ trained with $\widehat{\mathcal{L}}^{\text{HVDL}}(D)$ and $\widehat{\mathcal{L}}^{\text{SVDL}}(D)$ under the theoretical discriminator loss $\mathcal{L}(D)$. Denote by $D^*$ the optimal discriminator \cite{goodfellow2014generative} which minimizes $\mathcal{L}(D)$. Let $\widehat{D}^{\text{HVDL}}\triangleq{\arg\min}_{D\in\mathcal{D}}\widehat{\mathcal{L}}^{\text{HVDL}}(D)$; similarly, we define $\widehat{D}^{\text{SVDL}}$. We are interested in a reasonable bound (i.e, error bound) of the distance of $\widehat{D}^{\text{HVDL}}$ and $\widehat{D}^{\text{SVDL}}$ from $D^*$ under $\mathcal{L}(D)$, i.e., $\mathcal{L}(\widehat{D}^{\text{HVDL}})-\mathcal{L}(D^*)$ and $\mathcal{L}(\widehat{D}^{\text{SVDL}})-\mathcal{L}(D^*)$. These error bounds theoretically illustrate how HVDL and SVDL influence the discriminator training, which can guide our implementation of HVDL and SVDL in practice such as the selection of $\kappa$ and $\nu$.
	
	Before we move to the derivation, without loss of generality, we first assume $y \in [0,1]$. Then, we introduce some notations. Let $\mathcal{D}$ stand for the \textit{Hypothesis Space} of $D$. Please note that $\mathcal{D}$ may not cover $D^*$. Let $\hat{p}_r^{\text{KDE}}(y)$ and $\hat{p}_g^{\text{KDE}}(y)$ stand for the KDEs of $p_r(y)$ and $p_g(y)$ respectively. Let $p^r_w(y^\prime|y)\triangleq \frac{w^r(y^\prime,y)p^r(y^\prime)}{W^r(y)}$, $p^g_w(y^\prime|y)\triangleq \frac{w^g(y^\prime,y)p^g(y^\prime)}{W^g(y)}$, $W^r(y)\triangleq\int w^r(y^\prime,y)p_r(y^\prime)dy^\prime$ and $W^g(y)\triangleq\int w^g(y^\prime,y)p_g(y^\prime)dy^\prime$.

	
	\begin{definition} 
		\label{def:holder_class}
		(H\"older Class) Define the H\"older class of functions as:
		\begin{equation}
			\scalemath{0.85}{\Sigma(L)\triangleq\left\{p: \forall t_1, t_2\in\mathcal{Y}, \exists L >0, s.t. \frac{|p^\prime(t_1)-p^\prime(t_2)|}{|t_1-t_2|}\leq L \right\}. }
		\end{equation}
	\end{definition}

	Please see Supp. \ref{supp:CcGAN_convergence_rate} for more details of these notations. Moreover, we will also work with the following assumptions:
	
	{\setlength{\parindent}{0cm}\textbf{(A1)}} All $D$'s in $\mathcal{D}$ are measurable and uniformly bounded. Let $U \triangleq  \max \{  \sup_{D\in \mathcal{D}} \left[-\log D\right], \allowbreak \sup_{D\in \mathcal{D}} \left[ -\log (1-D) \right] \}$ and $U < \infty$;\\ {\setlength{\parindent}{0cm}\textbf{(A2)}} For $\forall \bm{x}\in\mathcal{X}$ and $y, y^\prime\in\mathcal{Y}$, $\exists g^r(\bm{x}) >0 $ and $ M^r>0$, s.t. $|p_r(\bm{x}|y^\prime)-p_r(\bm{x}|y)|\leq g^r(\bm{x})|y^\prime-y|$ with $\int g^r(\bm{x}) d\bm{x} = M^r$; \\
	{\setlength{\parindent}{0cm}\textbf{(A3)}} For $\forall \bm{x}\in\mathcal{X}$ and $y, y^\prime\in\mathcal{Y}$, $\exists g^g(\bm{x}) >0 $ and $ M^g>0$, s.t. $|p_g(\bm{x}|y^\prime)-p_g(\bm{x}|y)|\leq g^g(\bm{x})|y^\prime-y|$ with $\int g^g(\bm{x}) d\bm{x} = M^g$; \\
	{\setlength{\parindent}{0cm}\textbf{(A4)}} $p_r(y)\in\Sigma(L^r)$ and $p_g(y)\in\Sigma(L^g)$.

	With these definitions and assumptions, we derive two lemmas based on which we derive the error bounds of a discriminator trained by using HVDL and SVDL in Theorems \ref{thm:convergence_rate_HVDL} and \ref{thm:convergence_rate_SVDL}.
	
    \begin{lemma}[Lemma for HVDL]
    	\label{lem:conditional_mean_bound_HVDL}
    	Suppose that (A1)-(A2) and (A4) hold, then $\forall \delta\in(0,1)$, with probability at least $1-\delta$, 
    		\begin{align}
    			&\sup_{D\in\mathcal{D}}\left| \frac{1}{N_{y,\kappa}}\sum_{i=1}^{N}\mathbbm{1}_{\{ |y-y_i|\leq \kappa \}} \left[ -\log D(\bm{x}_i, y) \right] \right. \nonumber\\
    			& \quad \left. \vphantom{\sum_{i=1}^{N}} - \mathbb{E}_{\bm{x}\sim p(\bm{x}|y)} \left[ -\log D(\bm{x}, y) \right] \right| \nonumber\\
    			& \leq  U\sqrt{\frac{1}{2 N_{y,\kappa}}\log\left( \frac{2}{\delta} \right)} + \kappa U M,
    			\label{eq:conditional_mean_bound_HVDL}
    		\end{align}
    		for a fixed $y$. If image-label pairs are real, then  $N=N^r$, $N_{y,\kappa}=N^r_{y,\kappa}$, $p=p_r$, and $M=M^r$. Similarly, we have $N=N^g$, $N_{y,\kappa}=N^g_{y,\kappa}$, $p=p_g$, and $M=M^g$ for fake image-label pairs.
    \end{lemma}
	
	\begin{lemma}[Lemma for SVDL]
    	\label{lem:conditional_mean_bound_SVDL}
    	Suppose that (A1), (A2) and (A4) hold, then $\forall\delta\in (0,1)$, with probability at least $1-\delta$,
    		\begin{align}
    			&\sup_{D\in\mathcal{D}}\left| \frac{\frac{1}{N}\sum_{i=1}^{N}w(y_i,y)\left[ -\log D(\bm{x}_i, y) \right]}{\frac{1}{N}\sum_{i=1}^{N}w(y_i,y)} \right. \nonumber\\
    			& \quad \left. \vphantom{\sum_{i=1}^{N}} - \mathbb{E}_{\bm{x}\sim p(\bm{x}|y)} \left[ -\log D(\bm{x}, y) \right] \right| \nonumber\\
    			& \leq  \frac{2U}{W(y)}\sqrt{\frac{1}{2N} \log\left( \frac{4}{\delta}\right) } + UM \mathbb{E}_{y^\prime \sim p_w(y^\prime|y)} \left[|y^\prime-y| \right] ,
    			\label{eq:conditional_mean_bound_SVDL}
    		\end{align}
    		for a fixed $y$. If image-label pairs are real, then  $N=N^r$, $N_{y,\kappa}=N^r_{y,\kappa}$, $p=p_r$, $p_w=p_w^r$, $w=w^r$, $W=W^r$, and $M=M^r$. Similarly, we have $N=N^g$, $N_{y,\kappa}=N^g_{y,\kappa}$, $p=p_g$, $p_w=p_w^g$, $w=w^g$, $W=W^g$, and $M=M^g$ for fake image-label pairs.
    \end{lemma}
	
	\begin{theorem}[Error Bound of $D$ trained with HVDL]
		\label{thm:convergence_rate_HVDL} 
		Assume that (A1)-(A4) hold, then $\forall \delta\in(0,1)$, with probability at least $1-\delta$,
		\begin{align}
			&\mathcal{L}(\widehat{D}^{\text{HVDL}})-\mathcal{L}(D^*) \leq   2U\left(\sqrt{\frac{C_{1,\delta}^{\text{KDE}}\log N^r}{N^r\sigma}} +  L^r\sigma^2 \right)\nonumber\\
			& + 2U\left(\sqrt{\frac{C_{2,\delta}^{\text{KDE}}\log N^g}{N^g\sigma}} + L^g\sigma^2 \right) +  2\kappa U (M^r+M^g) \nonumber\\
			& +  2U\sqrt{\frac{1}{2} \log \left( \frac{8}{\delta}\right)} \left(  \mathbb{E}_{y\sim \hat{p}_r^{\text{KDE}}(y)} \left[ \sqrt{\frac{1}{N_{y, \kappa}^r}} \, \right] \right. \nonumber\\
			&\quad\left. + \mathbb{E}_{y\sim \hat{p}_g^{\text{KDE}}(y)} \left[ \sqrt{\frac{1}{N_{y, \kappa}^g}}\, \right] \right) \nonumber\\
			&+ \mathcal{L}(\widetilde{D})-\mathcal{L}(D^*), \label{eq:convergence_rate_HVDL}
		\end{align}
		for some constants $C_{1,\delta}^{\text{KDE}}, C_{2,\delta}^{\text{KDE}}$ depending on $\delta$. 
	\end{theorem}

	\begin{theorem}[Error Bound of $D$ trained with SVDL]
		\label{thm:convergence_rate_SVDL}
		Assume that (A1)-(A4) hold, then $\forall \delta\in(0,1)$, with probability at least $1-\delta$,
		\begin{align}
			&\mathcal{L}(\widehat{D}^{\text{SVDL}})-\mathcal{L}(D^*) \leq 2U\left(\sqrt{\frac{C_{1,\delta}^{\text{KDE}}\log N^r}{N^r\sigma}} + L^r\sigma^2 \right) \nonumber\\
			& + 2U\left(\sqrt{\frac{C_{2,\delta}^{\text{KDE}}\log N^g}{N^g\sigma}} + L^g\sigma^2 \right) \nonumber\\
			& + 4U\sqrt{\frac{1}{2}\log \left( \frac{16}{\delta}\right)} \left(  \frac{1}{\sqrt{N^r}}\mathbb{E}_{y\sim \hat{p}_r^{\text{KDE}}(y)}\left[  \frac{1}{W^r(y)}\right] \right.\nonumber\\
			& \quad \left. + \frac{1}{\sqrt{N^g}}\mathbb{E}_{y\sim \hat{p}_g^{\text{KDE}}(y)} \left[  \frac{1}{W^g(y)}\right] \right) \nonumber\\
			& + 2U \left( M^r\mathbb{E}_{y\sim \hat{p}_r^{\text{KDE}}(y)}\left[ \mathbb{E}_{y^\prime \sim \hat{p}^r_{w}(y^\prime|y)}\left|y^\prime-y\right|  \right] \right. \nonumber\\
			& \quad \left. + M^g\mathbb{E}_{y\sim \hat{p}_g^{\text{KDE}}(y)}\left[  \mathbb{E}_{y^\prime \sim \hat{p}^g_{w}(y^\prime|y)}\left|y^\prime-y\right| \right] \right) \nonumber\\
			& + \mathcal{L}(\widetilde{D})-\mathcal{L}(D^*), 
			\label{eq:convergence_rate_SVDL}
		\end{align}
		for some constant $C_{1,\delta}^{\text{KDE}}, \; C_{2,\delta}^{\text{KDE}}$ depending on $\delta$. 
	\end{theorem}

	\begin{remark}
		Please see Supp.\ \ref{supp:CcGAN_convergence_rate} for the proofs to these lemmas and theorems.
	\end{remark}

	
	\begin{remark}[Illustration of Theorems \ref{thm:convergence_rate_HVDL} and \ref{thm:convergence_rate_SVDL}]
		\label{rmk:illustration_error_bounds}
		Both theorems imply HVDL and SVDL perform well if the output of $D$ is not too close to 0 or 1 (i.e., favor small $U$). The first two terms in both upper bounds control the quality of KDE, which implies KDE works better if we have a large $N^r$ and a large $N^g$ but a small $\sigma$. The remaining terms of the two bounds are different. In the HVDL case, we favor small $\kappa$, $M^r$, and $M^g$. However, we should avoid setting $\kappa$ too small, because we prefer large $N^r_{y,\kappa}$ and $N^g_{y,\kappa}$. In the SVDL case, we prefer small $M^r$ and $M^g$ but large $W^r(y)$ and $W^g(y)$. Large $W^r(y)$ and $W^g(y)$ imply that the weight function decays slowly (i.e., small $\nu$). However, we should avoid setting $\nu$ too small because a small $\nu$ leads to large $\mathbb{E}_{y^\prime \sim \hat{p}^r_{w}(y^\prime|y)}\left|y^\prime-y\right|$ and $\mathbb{E}_{y^\prime \sim \hat{p}^g_{w}(y^\prime|y)}\left|y^\prime-y\right|$ (i.e., large weights for $y^\prime$'s which are far away from $y$). The rule of thumb formulae to select $\kappa$ and $\nu$ in Remark \ref{rmk:rule_of_thumb} are consistent with our analysis here. Besides the rules of thumb, future work  should propose a more refined hyper-parameter selection method.
	\end{remark}

	\section{Sliding Fr\'echet Inception Distance} \label{sec:sfid} 
	
\begin{figure}[!htbp]
 		\centering
 		\includegraphics[width=0.6\linewidth]{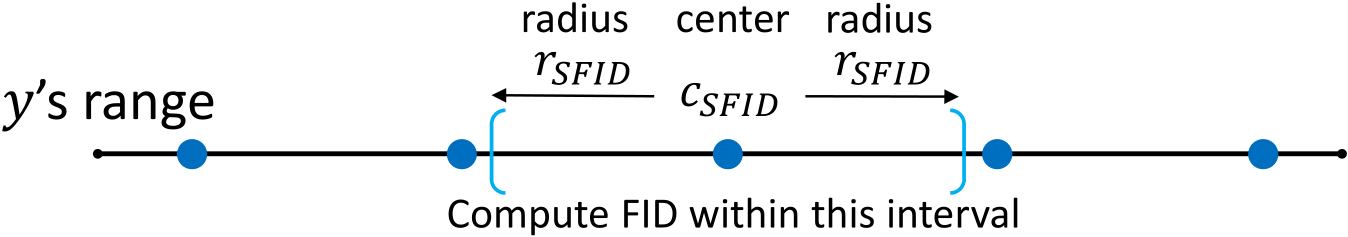}
 		\caption{\textbf{Sliding Fr\'echet Inception Distance (SFID).} We preset finite centers (blue dots) on $y$'s range evenly and a radius $r_{\text{SFID}}$. Given an interval $[c_{\text{SFID}}-r_{\text{SFID}}, c_{\text{SFID}}+r_{\text{SFID}}]$, we compute FID between fake and real images with labels in this interval. SFID is the average of these FIDs.}
 		\label{fig:SFID}
 	\end{figure}

	A conditional GAN (no matter the type of the condition) needs to be evaluated from three perspectives \cite{miyato2018cgans,SAGAN-zhang19d,devries2019evaluation}: (1) the visual quality, (2) the intra-label diversity (the diversity of fake images with the same label), and (3) the label consistency (whether assigned labels of fake images are consistent with their actual labels). Measuring the performance of cGANs from these three perspectives is often conducted by using a popular overall metric, termed the Intra-FID \cite{miyato2018cgans,SAGAN-zhang19d,devries2019evaluation}. Intra-FID computes the \textit{Fr\'echet inception distance} (FID) \cite{heusel2017gans} separately at each of the distinct labels and reports the average FID score. Intra-FID is also used in our experiments on RC-49, UTKFace, and Cell-200 in Section \ref{sec:experiment}; however, Intra-FID is not reliable or is even inapplicable when we have very few (even zero) real images for some distinct regression labels, e.g., the experiment on the Steering Angle dataset in Section \ref{sec:steeringangle}. We therefore propose a novel metric, termed the \textit{Sliding Fr\'echet Inception Distance} (SFID), to replace Intra-FID in this scenario. SFID computes FID within an interval sliding on the range of the regression label $y$, and then reports the average of these FIDs. Specifically, we first prespecify a finite set of \textit{SFID centers}  $c_{\text{SFID}}$  evenly over the range of $y$ and a constant \textit{SFID radius} $r_{\text{SFID}}$. Then, based on the $c_{\text{SFID}}$ and $r_{\text{SFID}}$, we can define many joint \textit{SFID intervals} of the form  $[c_{\text{SFID}}-r_{\text{SFID}}, c_{\text{SFID}}+r_{\text{SFID}}]$. For each SFID interval, we compute FID between real and fake images with labels within this interval. Finally, SFID reports the average of these FIDs. Usually, we also report the standard deviation of these FIDs. We visualize the procedure for computing SFID in Fig. \ref{fig:SFID}. A pseudo code for computing SFID is shown in Alg. \ref{alg:compute_SFID}. Similar to Intra-FID, a small SFID is preferred.
	

    \begin{algorithm}[!htbp]
    	\footnotesize
    	\SetAlgoLined
    	\KwData{Real image-label pairs $D^r=\{ (\bm{x}^r_1, y_{1}^r), \cdots, (\bm{x}^r_{N^r}, y_{N^r}^r) \}$; fake image-label pairs $D^g=\{ (\bm{x}^g_1, y_{1}^g), \cdots, (\bm{x}^g_{N^g}, y_{N^g}^g)\}$; preset SFID centers $\{ y_{1}^c, \cdots, y_{N^c}^c \}$; preset window radius $r_{SFID}$.}
    	\KwResult{$SFID$.} 
    	Initialize real and fake image sets in $l$-th sliding-window, i.e., $D_{l}^r=\phi$ and $D_{l}^g=\phi$, $l=1,2,\cdots, N^c$\;
    	Initialize a FID array $FID(l)=\infty, l=1,2,\cdots, N^c$\; 
    	
    	\For{$l=1$ \KwTo $N^c$}{
    		Update real image set at $y_{l}^c$: $D_l^r = \bigcup_{|y_{i}^r - y_{l}^c|\leq r_{{SFID}}} \{\bm{x}_i^r\}, i=1,2,\cdots, N^r$\;
    		Update fake image set at $y_{l}^c$: $D_l^g = \bigcup_{|y_{i}^g - y_{l}^c|\leq r_{{SFID}}} \{\bm{x}_i^g\}, i=1,2,\cdots, N^g$\;
    		Compute current FID: $FID(l)=FID(D_l^r, D_l^g)$ \; 
    	}
    	Compute $SFID=\frac{1}{N^c} \sum_{l=1}^{N^c} FID(l)$.
    	\caption{An algorithm to compute the Sliding Fr\'echet Inception Distance (SFID).}
    	\label{alg:compute_SFID}
    \end{algorithm}

	\section{Low-resolution Experiments}\label{sec:experiment}
	
	In this section, we study the effectiveness of CcGAN on four image datasets with resolution $64\times 64$. Please note that our CGM task has never been studied in the literature, so there is no direct baseline. We modify conventional cGANs to create cGAN ($K$ classes) and cGAN (concat) as baselines. For a fair comparison, from Sections \ref{sec:rc49} to \ref{sec:steeringangle}, \textbf{all candidate methods use the same network architecture} (a customized DCGAN \cite{radford2015unsupervised} architecture for Cell-200, and the SNGAN \cite{miyato2018spectral} architecture for the remaining datasets) except for the label input modules. The four CcGAN methods (i.e., HVDL+NLI, SVDL+NLI, HVDL+ILI, and SVDL+ILI) are tested in our experiments below. For stability, regression labels in all datasets are normalized to $[0,1]$ during training.
	
	Since GANs \cite{goodfellow2014generative} do not explicitly estimate density functions, to measure the CGM quality, we evaluate the quality of fake images sampled from GANs. Following other cGAN methods \cite{miyato2018cgans,SAGAN-zhang19d,devries2019evaluation}, we use Intra-FID as the overall metric in our RC-49 (Section\ \ref{sec:rc49}), UTKFace (Section\ \ref{sec:utkface}), and Cell-200 (Section\ \ref{sec:cell200}) experiments. The proposed SFID (see Section\ \ref{sec:sfid}) is only used in the experiment conducted on the Steering Angle dataset in Section\ \ref{sec:steeringangle}, where there are not enough real images to compute Intra-FID. The effectiveness of SFID is studied on the RC-49 dataset in Section\ \ref{sec:effectiveness_SFID}, where we can control the sample size of real images. Besides Intra-FID and SFID, in each experiment (except Cell-200), we also compute three separate scores, i.e., Naturalness Image Quality Evaluator (NIQE) \cite{mittal2012making}, Diversity, and Label Score, which evaluate fake images from three different perspectives. Furthermore, following \cite{mirza2014conditional, gulrajani2017improved, miyato2018spectral, brock2018large}, we also report \textit{Inception Score} (IS) \cite{salimans2016improved} and \textit{Fr\'echet Inception Distance} (FID) \cite{heusel2017gans} of each cGAN for completeness; however, as illustrated in Section \ref{sec:evaluation_FID_IS}, IS and FID are not appropriate overall metrics for our experiment. The quantitative performances of the candidate cGANs on the four datasets are summarized in Table~\ref{tab:low_resolution_results}.
	
	In the final experiment of this section, we demonstrate in Section \ref{sec:compare_against_sota} that CcGAN (SVDL+ILI) can significantly outperform state-of-the-art class-conditional GANs such as SNGAN \cite{miyato2018spectral}, SAGAN \cite{SAGAN-zhang19d}, BigGAN \cite{brock2018large}, CR-BigGAN \cite{Zhang2020Consistency}, BigGAN+DiffAug \cite{zhao2020differentiable}, and ReACGAN \cite{reacgan2021} in the CGM task.

    \begin{table*}[!htbp]
		\centering
		\caption{\textbf{Average quality of $64\times 64$ fake images from cGANs and CcGANs with standard deviation after the ``$\pm$" symbol.} We generate 179800, 60000, 200000, and 100000 fake images via each candidate method in the RC-49, UTKFace, Cell-200 and Steering Angle experiments, respectively. These fake images are evaluated through four metrics: Intra-FID, NIQE, Diversity, and Label Score. Note that IS and FID scores are also reported for completeness, but they are not suitable overall metrics for our task (see Section~\ref{sec:compare_against_sota} for details). ``$\downarrow$" (``$\uparrow$") indicates lower (higher) values are preferred. The best and second best results are marked in gray and green respectively.}
		\begin{adjustbox}{width=0.9\textwidth}
			\begin{tabular}{c|c|cccc|cc}
				\toprule
				\textbf{Dataset} & \textbf{Model} & \textbf{Intra-FID} $\downarrow$ & \textbf{NIQE} $\downarrow$  & \textbf{Diversity} $\uparrow$ & \textbf{Label Score} $\downarrow$ & \textbf{IS} $\uparrow$ & \textbf{FID} $\downarrow$ \\
				\midrule
				\multicolumn{1}{c|}{\multirow{6}[0]{*}{\rotatebox{0}{\textbf{RC-49}}}} & cGAN (150 classes)  & $1.720 \pm 0.384$ & $2.731 \pm 0.162$ & $0.779 \pm 0.199$ & $4.815 \pm 5.152$ & $2.382$ & 1.066 \\
				& cGAN (concat) & $1.141 \pm 0.108$ & $1.819 \pm 0.111$ & $2.459 \pm 0.049$ & $30.212 \pm 21.391$ & $11.440$ & 0.295 \\
				& CcGAN (HVDL+NLI) & $0.612 \pm 0.145$ & $1.869 \pm 0.181$ & $2.353 \pm 0.121$ & $5.617 \pm 4.452$ & $14.730$ & 0.285 \\
				& CcGAN (SVDL+NLI) & $0.515 \pm 0.181$ & $1.853 \pm 0.159$ & $2.610 \pm 0.113$ & $4.982 \pm 4.439$ & \cellcolor{green!25}{$19.425$} & \cellcolor{green!25}{0.207} \\
				& CcGAN (HVDL+ILI) & \cellcolor{green!25}{${0.424 \pm 0.081}$} & \cellcolor{green!25}{${1.805 \pm 0.179}$} & \cellcolor{green!25}{${2.814 \pm 0.052}$} & \cellcolor{green!25}{${1.816 \pm 1.481}$} & $17.992$ & 0.213 \\
				& CcGAN (SVDL+ILI) & \cellcolor{gray!25}{${0.389 \pm 0.095}$} & \cellcolor{gray!25}{${1.783 \pm 0.173}$} & \cellcolor{gray!25}{${2.949 \pm 0.069}$} & \cellcolor{gray!25}{${1.940 \pm 1.489}$} & \cellcolor{gray!25}{$20.173$} & \cellcolor{gray!25}{0.197} \\
				\midrule
				\multicolumn{1}{c|}{\multirow{6}[0]{*}{\rotatebox{0}{\textbf{UTKFace}}}} & cGAN (60 classes)  & $4.516 \pm 0.965$ & $2.315 \pm 0.306$ & $0.254 \pm 0.353$ & $11.087 \pm 8.119$ & $2.636$ & 0.963 \\
				& cGAN (concat)  & $0.834 \pm 0.199$ & $2.051 \pm 0.227$ & \cellcolor{gray!25}{${1.394 \pm 0.026}$} & $17.291 \pm 11.717$ & $3.103$ & 0.465 \\
				& CcGAN (HVDL+NLI) & ${0.572 \pm 0.167}$ & ${1.739 \pm 0.145}$ & \cellcolor{green!25}{$1.338 \pm 0.178$} & ${9.782 \pm 7.166}$ & \cellcolor{green!25}{$3.328$} & 0.114 \\
				& CcGAN (SVDL+NLI) & ${0.547 \pm 0.181}$ & ${1.753 \pm 0.196}$ & $1.326 \pm 0.198$ & ${10.739 \pm 8.340}$ & $3.307$ & \cellcolor{green!25}{0.087} \\
				& CcGAN (HVDL+ILI) & \cellcolor{green!25}{${0.480 \pm 0.145}$} & \cellcolor{gray!25}{${1.709 \pm 0.169}$} & ${1.280 \pm 0.203}$ & \cellcolor{green!25}{${7.505 \pm 5.857}$} & $3.256$ & \cellcolor{gray!25}{0.056} \\
				& CcGAN (SVDL+ILI) & \cellcolor{gray!25}{${0.425 \pm 0.157}$} & \cellcolor{green!25}{${1.725 \pm 0.171}$} & ${1.298 \pm 0.176}$ & \cellcolor{gray!25}{${7.452 \pm 6.022}$} & \cellcolor{gray!25}{$3.382$} & 0.142 \\
				\midrule
				\multicolumn{1}{c|}{\multirow{6}[0]{*}{\rotatebox{0}{\textbf{Cell-200}}}} & cGAN (100 classes)  & $90.255 \pm 64.595$ & $2.130 \pm 2.440$ & \longdash & $66.748 \pm 51.711$ & \longdash & 30.086 \\ 
				& cGAN (concat)  & $41.599 \pm 21.430$ & $3.250 \pm 0.646$ & \longdash & $73.187 \pm 51.133$ & \longdash & 37.689 \\
				& CcGAN (HVDL+NLI) & ${50.052 \pm 20.584}$ & ${1.488 \pm 0.153}$ & \longdash & ${72.599 \pm 37.425}$ & \longdash & 40.279 \\
				& CcGAN (SVDL+NLI) & ${56.078 \pm 19.334}$ & ${1.829 \pm 0.386}$ & \longdash & ${83.367 \pm 49.577}$ & \longdash & 51.318 \\
				& CcGAN (HVDL+ILI) & \cellcolor{green!25}{${8.759 \pm 6.652}$} & \cellcolor{green!25}{${1.283 \pm 0.534}$} & \longdash & \cellcolor{gray!25}{${5.861 \pm 4.900}$} & \longdash & \cellcolor{green!25}{3.263} \\
				& CcGAN (SVDL+ILI) & \cellcolor{gray!25}{${7.266 \pm 2.305}$} & \cellcolor{gray!25}{${1.220 \pm 0.515}$} & \longdash & \cellcolor{green!25}{${5.905 \pm 5.020}$} & \longdash & \cellcolor{gray!25}{1.684} \\
				\midrule
				\multicolumn{1}{c|}{\multirow{6}[0]{*}{\rotatebox{0}{\textbf{Steering Angle}}}} & cGAN (210 classes)  & $3.285 \pm 0.647$ & $1.296 \pm 0.095$ & $0.603 \pm 0.396$ & $14.596 \pm 15.402$ & $2.572$ & 0.976 \\
				& cGAN (concat)  & $2.446 \pm 1.122$ & $1.717 \pm 0.003$ & \cellcolor{gray!25}{${1.255 \pm 0.015}$} & $41.686 \pm 25.864$ & $3.251$ & \cellcolor{green!25}{0.255} \\
				& CcGAN (HVDL+NLI) & ${1.969 \pm 0.676}$ & \cellcolor{gray!25}{${1.093 \pm 0.024}$} & ${0.991 \pm 0.361}$ & ${22.322 \pm 18.758}$ & $3.587$ & 0.316 \\
				& CcGAN (SVDL+NLI) & ${1.866 \pm 0.649}$ & \cellcolor{green!25}{${1.098 \pm 0.038}$} & ${1.007 \pm 0.248}$ & ${19.678 \pm 18.281}$ & $3.968$ & \cellcolor{gray!25}{0.212} \\
				& CcGAN (HVDL+ILI) & \cellcolor{green!25}{${1.635 \pm 0.699}$} & ${1.152 \pm 0.047}$ & ${1.153 \pm 0.153}$ & \cellcolor{gray!25}{${10.868 \pm 9.644}$} & \cellcolor{gray!25}{$4.592$} & {0.327} \\
				& CcGAN (SVDL+ILI) & \cellcolor{gray!25}{${1.546 \pm 0.626}$} & ${1.130 \pm 0.078}$ & \cellcolor{green!25}{${1.156\pm 0.189}$} & \cellcolor{green!25}{${10.933 \pm 8.978}$ } & \cellcolor{green!25}{$4.439$} & {0.331} \\
				\bottomrule
			\end{tabular}%
		\end{adjustbox}
		\label{tab:low_resolution_results}%
	\end{table*}%

	\subsection{RC-49}\label{sec:rc49}
	
	Since most benchmark datasets in the GAN literature do not have continuous, scalar regression labels, we propose a new benchmark dataset---RC-49, a synthetic dataset created by rendering 49 3-D chair models at different yaw angles. Each of 49 chair models is rendered at 899 yaw angles ranging from 0.1$^{\circ}$ to 89.9$^{\circ}$ with step size 0.1$^{\circ}$. Therefore, RC-49 consists of 44,051 $64\times 64$ rendered RGB images and 899 distinct angles. Please see Supp. \ref{supp:details_of_rc49} for more details of the data generation. Example images are shown in Fig.\  \ref{fig:rc49_visual_results} in Appendix. 
	
	{\setlength{\parindent}{0cm}\textbf{Experimental setup:}} Not all images are used for the GAN training. A yaw angle is selected for training if its last digit is odd. Moreover, at each selected angle, only 25 images are randomly chosen for training. Thus, the training set includes 11250 images and 450 distinct angles. The remaining images are held out for evaluation. 
	
	When training cGAN ($K$ classes), we divide $[0.1^{\circ}, 89.9^{\circ}]$ into 150 equal intervals where each interval is treated as a class. When training CcGAN, we use the rule of thumb formulae in Remark \ref{rmk:rule_of_thumb} to select the three hyper-parameters of HVDL and SVDL, i.e., $\sigma\approx 0.047$, $\kappa\approx 0.004$ and $\nu=50625$. The two novel label input mechanisms for CcGAN (NLI and ILI) are implemented in this experiment. For ILI, we pre-train a modified ResNet-34 \cite{he2016deep} with 3 linear layers after the average pooling layer and we only keep the last linear layer for label embedding (i.e., the $T_1+T_2$ in Fig.\ \ref{fig:pre_trained_CNN_for_label_embedding}). We use a five-layer MLP with 128 nodes in each layer to convert an angle into its hidden representation (i.e., the $T_3$ in Fig.\ \ref{fig:label_embedding_network}). All candidates are trained for 30,000 iterations with batch size 256. Afterwards, we evaluate the trained GANs on all 899 angles by generating 200 fake images for each angle.  Please see Supp.\ \ref{supp:details_of_rc49} for the network architectures and more details about the training/testing setup. 
	
	{\setlength{\parindent}{0cm}\textbf{Quantitative and visual results:}} To evaluate (1) the visual quality, (2) the intra-label diversity, and (3) the label consistency of fake images, we study an overall metric and three separate metrics here. (i) \textbf{Intra-FID} \cite{miyato2018cgans} is utilized as the overall metric. It computes FID \cite{heusel2017gans} separately at each of the 899 evaluation angles and reports the average FID score along with the standard deviation of these 899 FIDs. (ii) \textbf{NIQE} \cite{mittal2012making} measures the visual quality only.  (iii) \textbf{Diversity} is the average entropy of predicted chair types of fake images over evaluation angles.
	(iv) \textbf{Label Score} is the average absolute error between assigned angles and predicted angles. Please see Supp. \ref{supp:rc49_performance} for details of these metrics.
	
	We report in Table \ref{tab:low_resolution_results} the performances of each GAN. The example fake images in Fig.\ \ref{fig:rc49_visual_results} in Appendix and line graphs in Fig.\ \ref{fig:rc49_line_graphs} support the quantitative results. cGAN (150 classes) often generates unrealistic, identical images for a target angle (i.e., low visual quality and low intra-label diversity). ``Binning'' $[0.1^{\circ}, 89.9^{\circ}]$ into other number of classes (e.g., 90 classes and 210 classes) is also tried but does not improve cGAN's performance. cGAN (concat) has good visual quality and high intra-label diversity but terrible label consistency. In contrast, the four CcGAN methods perform well from all three perspectives, i.e., good visual quality, high intra-label diversity, and high label consistency. Moreover, both ILI-based CcGANs outperform the two NLI-based CcGANs in terms of all four metrics, especially the label score. 
	

	\begin{figure}[!htbp]
		\centering
		\includegraphics[width=0.8\linewidth]{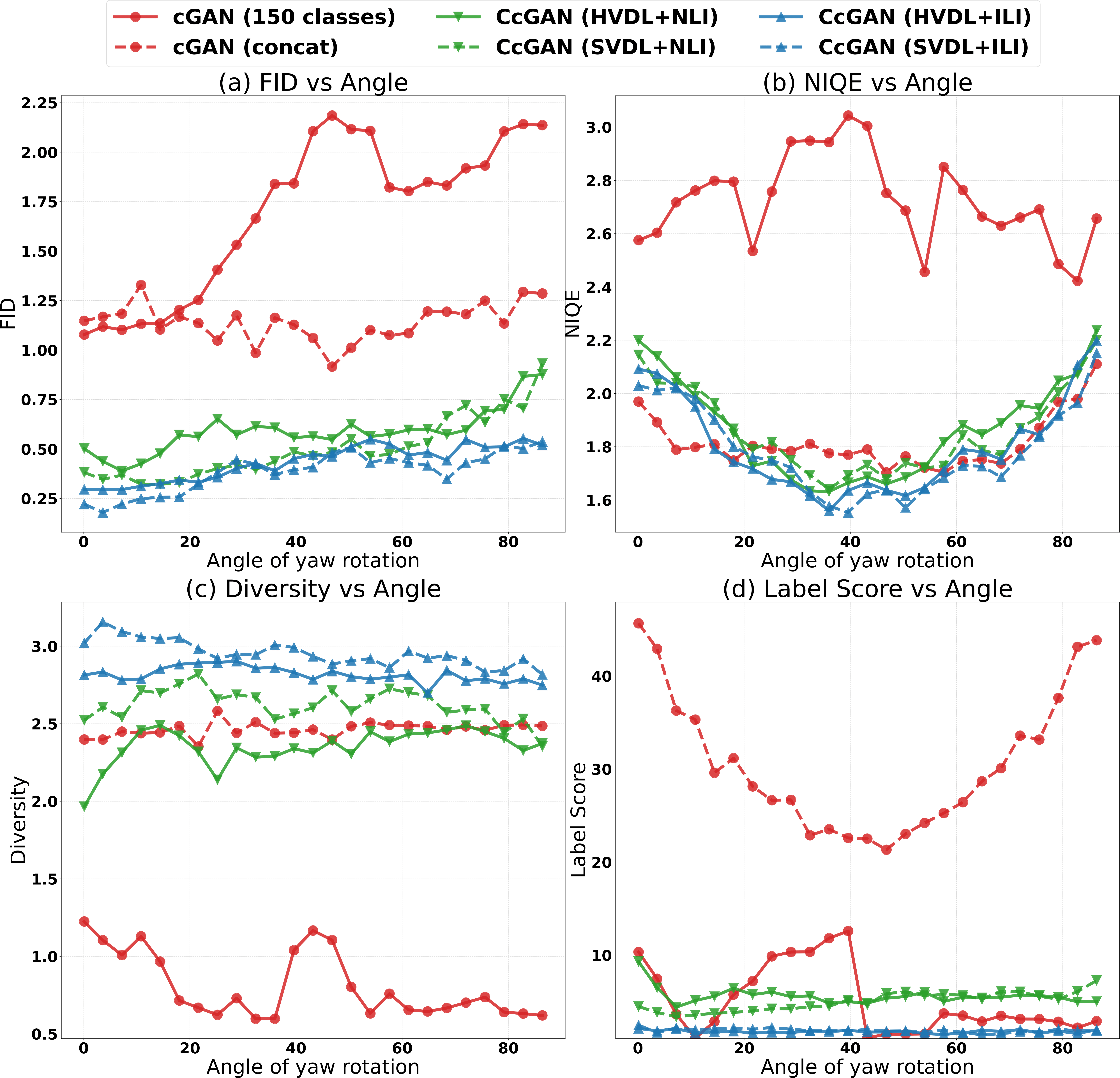}
		\caption{ \textbf{Line graphs of FID/NIQE/Diversity/Lable Score versus yaw angle for RC-49.} Figs. (a) to (c) show that four CcGAN methods consistently outperform cGAN (150 classes) across all angles. All graphs of CcGANs appear much smoother than those of cGAN (150 classes) because of HVDL and SVDL. Figs. (a) and (d) show that four CcGAN methods consistently outperform cGAN (concat) across all angles. Moreover, in most graphs, we can clearly see ILI-based CcGANs perform better than NLI-based CcGANs.}
		\label{fig:rc49_line_graphs}
	\end{figure}
	
	\begin{figure}[!htbp]  
		\centering
		\includegraphics[width=0.5\linewidth]{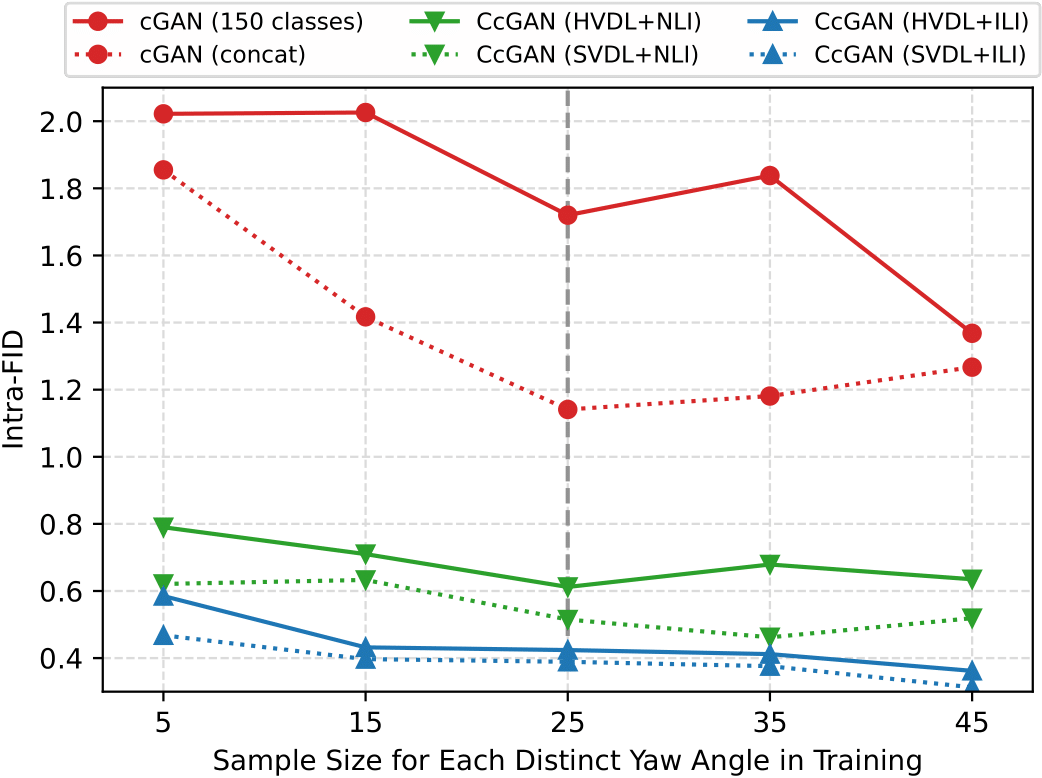}
		\caption{ \textbf{Line graphs of Intra-FID versus the sample size for each distinct training angle of RC-49.} The grey vertical dashed line stands for the sample size used in the main study of the RC-49 experiment. Four CcGAN methods substantially outperform two cGANs and ILI performs better than NLI no matter what the sample size for each distinct angle in the training set. The overall trend in this figure shows that a smaller sample size deteriorates the performance of both cGAN and CcGAN.}
		\label{fig:Extra_Exp2_RC49}
	\end{figure}
	
	{\setlength{\parindent}{0cm}\textbf{Extra experimental results:}} To test cGAN and CcGAN under more challenging scenarios, we vary the sample size for each distinct angle in the training set from 45 to 5. We visualize the line graphs of Intra-FID versus the sample size for each distinct training angle in Fig.\ \ref{fig:Extra_Exp2_RC49}. From this figure, we can see the four CcGAN methods substantially outperform two cGANs and ILI performs better than NLI no matter what is the sample size for each distinct angle in the training set. The overall trend in this figure also shows that smaller sample size reduces the performance of both cGAN and CcGAN.

	\subsection{UTKFace}\label{sec:utkface}
	
	In this section, we compare CcGAN and cGAN on UTKFace \cite{utkface}, a dataset consisting of RGB images of human faces which are labeled by age.
	
	{\setlength{\parindent}{0cm}\textbf{Experimental setup:}} In this experiment, we only use images with age in $[1, 60]$. Some images with bad visual quality and watermarks are also discarded. After the preprocessing, 14,760 images are left. The number of images for each age ranges from 50 to 1051. We resize all selected images to $64\times 64$. Some example UTKFace images are shown in the first image array in Fig.\ \ref{fig:UTKFace_visual_results}. 
		
	When implementing cGAN ($K$ classes), each age is treated as a class. For CcGAN we use the rule of thumb formulae in Remark\ \ref{rmk:rule_of_thumb} to select the three hyper-parameters of HVDL and SVDL, i.e., $\sigma\approx 0.041$, $\kappa\approx 0.017$ and $\nu=3600$. Similar to the RC-49 experiment, we use NLI and ILI to incorporate ages into CcGAN. All GANs are trained for 40,000 iterations with batch size 512. In testing, we generate 1,000 fake images from each trained GAN for each age.  Please see Supp. \ref{supp:details_of_utkface} for more details of the data preprocessing, network architectures and training/testing setup.
	
\begin{figure}[!htbp]
		\centering
		\includegraphics[width=0.8\linewidth]{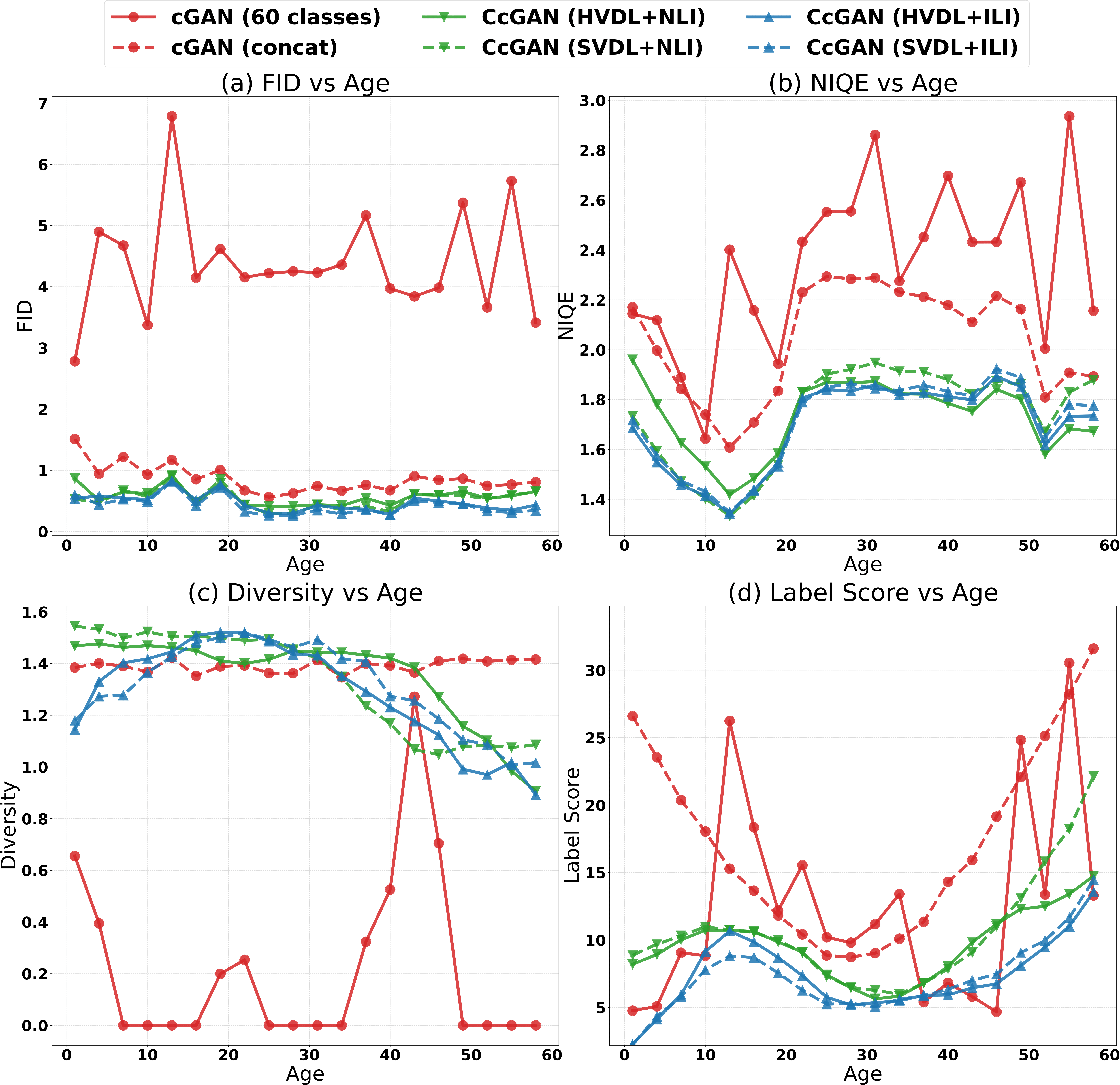}
		\caption{\textbf{Line graphs of FID/NIQE/Diversity/Lable Score versus Age for UTKFace.}  The four CcGAN methods significantly outperform cGAN (60 classes) in Figs. (a) to (c). All graphs of CcGANs appear much smoother than those of cGAN (60 classes) because of HVDL and SVDL. Figs. (a) and (d) show that four CcGAN methods consistently outperform cGAN (concat) across all ages. Fig. (d) also shows that the ILI-based CcGANs have higher label consistency than the NLI-based CcGANs.}
		\label{fig:utkface_line_graphs}
	\end{figure}

	{\setlength{\parindent}{0cm}\textbf{Quantitative and visual results:}} Similar to the RC-49 experiment, we evaluate the quality of fake images by Intra-FID, NIQE, Diversity (entropy of predicted races), and Label Score. We report in Table \ref{tab:low_resolution_results} the average quality of 60,000 fake images. From this table, we can see the four CcGAN methods substantially outperform both cGANs and ILI performs better than NLI. Notably, although cGAN (concat) has the highest Diversity score, the huge label score reveals that cGAN (concat) cannot control the image generation with respective to ages. Thus, cGAN (concat) fails in this experiment. We also show in Fig.\ \ref{fig:UTKFace_visual_results} some example fake images from candidate models and line graphs of FID/NIQE/Diversity/Lable Score versus Age in Fig.\ \ref{fig:utkface_line_graphs}. Analogous to the quantitative comparisons, we can see that CcGAN performs much better than cGAN.
	

	\begin{figure}[!htbp]
		\centering
		\includegraphics[width=0.5\linewidth]{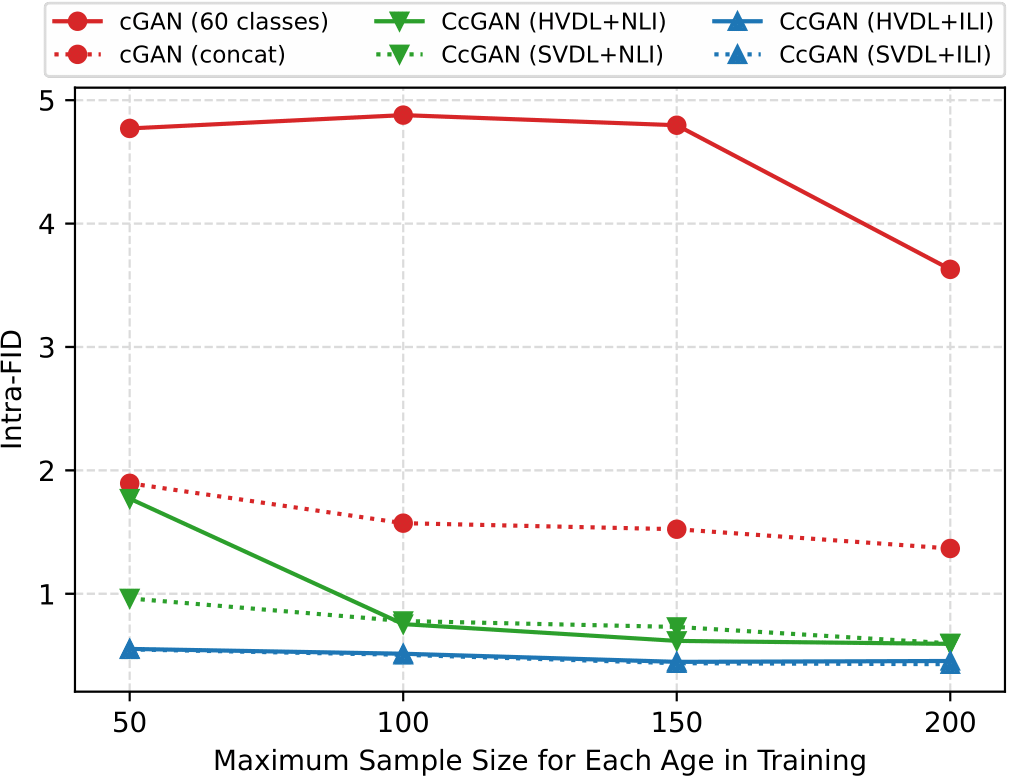}
		\caption{ \textbf{Line graphs of Intra-FID versus the maximum sample size for each distinct age in the training set of UTKFace.} Fig.\ \ref{fig:Extra_Exp7_UTKFace} shows that four CcGAN methods perform much better than cGAN and ILI is better than NLI. Moreover, a smaller sample size deteriorates the performance of both cGAN and CcGAN. }
		\label{fig:Extra_Exp7_UTKFace}
    \end{figure}

	{\setlength{\parindent}{0cm}\textbf{Extra experimental results:}} The histogram in Fig.\ \ref{fig:utkface_histogram} shows that the UTKFace dataset is highly imbalanced. To balance the training data and also test the performance of cGAN and CcGAN under smaller sample sizes, we vary the maximum sample size for each distinct age in the training from 200 to 50. Note that, we do not restrict the maximum sample size in the main study. Since we have a much smaller sample size, we reduce the number of iterations for the GAN training from 40,000 to 20,000 and slightly increase $m_{\kappa}$ in Remark\ \ref{rmk:rule_of_thumb} from 1 to 2 (we therefore use a wider hard/soft vicinity). We visualize the line graphs of Intra-FID versus the maximum sample size for each age of cGAN and CcGAN in Fig.\ \ref{fig:Extra_Exp7_UTKFace}. From the figure, we can clearly see that a smaller sample size worsens the performance of both cGAN and CcGAN. Moreover, the Intra-FID scores of two cGANs often stay at a high level and are larger than those of the four CcGAN methods. The ILI-based CcGANs are also better than the NLI-based CcGANs.

	\subsection{Cell-200}\label{sec:cell200}
	
	In addition to RC-49, we propose another benchmark dataset--Cell-200, a dataset of synthetic fluorescence microscopy images with cell populations generated by SIMCEP \cite{lehmussola2007computational}. Please see Supp. \ref{supp:cell200_data} for more details about the data generation. Some example images are shown in Fig.\  \ref{fig:cell200_visual_results}. 
	
	{\setlength{\parindent}{0cm}\textbf{Experimental setup:}} The Cell-200 dataset consists of 200,000 $64\times 64$ grayscale images. The number of cells per image ranges from 1 to 200 and there are 1,000 images for each cell count. However, only a subset of Cell-200 with only odd cell counts and 10 images per count (1,000 training images in total) is used for the GAN training.
	
	When training cGAN ($K$ classes), we divide $[1, 200]$ into 100 equal intervals where each interval is treated as a class (i.e., $K=100$). We use the rule of thumb formulae in Remark \ref{rmk:rule_of_thumb} to select the three hyperparameters of HVDL and SVDL, i.e., $\sigma\approx 0.077$, $\kappa\approx 0.020$ and $\nu=2500$.  Both cGAN and CcGAN are trained for 5,000 iterations. Afterwards, we evaluate the trained GANs on all 200 cell counts by generating 1,000 fake images for each count.  Please see Supp.\ \ref{supp:details_of_cell200} for the network architectures and more details about the training/testing setup. 
	
\begin{figure*}[!htbp]
		\centering
		\includegraphics[width=0.98\linewidth]{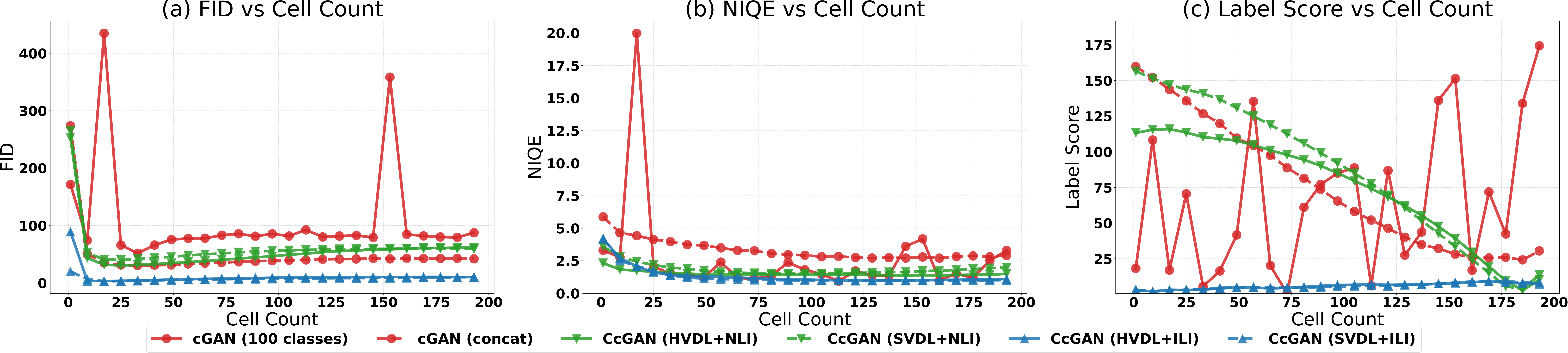} 
		\caption{\textbf{Line graphs of FID/NIQE/Lable Score versus Cell Count for Cell-200.} Figs. (a) to (c) show that, although the NLI-based CcGANs do not perform well, the ILI-based CcGANs outperform both cGANs across most cell counts. All graphs of CcGANs also appear much smoother than those of cGAN (100 classes) because of HVDL and SVDL. Moreover, in all figures, we can see ILI-based CcGANs perform better than NLI-based CcGANs especially in Fig. (c).}
		\label{fig:cell200_line_graphs}
	\end{figure*}

	{\setlength{\parindent}{0cm}\textbf{Quantitative and visual results:}} We evaluate the quality of fake images by Intra-FID, NIQE, and Label Score. Please note that the Diversity score is not available in this experiment because there is no class label in Cell-200. We report in Table \ref{tab:low_resolution_results} the average quality of 200,000 fake images from cGAN and CcGAN. We also show in Fig.\ \ref{fig:cell200_visual_results} some example fake images from cGAN and CcGAN and line graphs of FID/NIQE/Label Score versus Cell Count in Fig.\ \ref{fig:cell200_line_graphs}. Unlike the experimental results on RC-49 and UTKFace, although NLI-based CcGANs outperform cGAN (100 classes) in terms of Intra-FID and NIQE, their label scores are very high, which implies low label consistency. Fortunately, two ILI-based CcGANs still perform very well and substantially outperform two cGANs and two NLI-based CcGANs.


	\subsection{Steering Angle}\label{sec:steeringangle}

	In this section, we demonstrate the effectiveness of the proposed CcGAN on the Steering Angle dataset, a subset of an autonomous driving dataset \cite{steeringangle}. The complete dataset consists of 109,231 RGB images. Each image is taken by using a dash camera mounted on a car and, at the same moment, the angle of the steering wheel rotation of the same car (i.e., steering angle) is recorded by a device attached to the steering wheel. Thus, each image in this autonomous driving dataset is paired with a steering angle ranging from $-338.82^\circ$ to $501.78^\circ$. 
	
	{\setlength{\parindent}{0cm}\textbf{Experimental setup:}} To make the training and evaluation easier, we remove many images in this autonomous driving dataset where an image is removed due to at least one of the following reasons:
	\begin{itemize}
		\item The image is incorrectly labeled (e.g., some images show that the car was turning left/right but the corresponding steering angles are zero).
		\item The image has very bad visual quality due to overexposure or underexposure.
		\item There is no reference object (e.g., double yellow lines or side roads) in the image to let a human visually determine whether the car was turning left/right.
		\item The corresponding steering angle is outside $[-80^\circ, 80^\circ]$.
	\end{itemize}
	Eventually, there are 12,271 images left with 1,904 distinct steering angles in $[-80^\circ, 80^\circ]$. These images are then resized to $64\times 64$ and they form a subset of the autonomous driving dataset \cite{steeringangle}, termed \textit{Steering Angle} in this paper. Please note that the Steering Angle dataset is highly imbalanced and a histogram of steering angles is shown in Fig.\ \ref{fig:steeringangle_histogram}.
	
	When training cGAN ($K$ classes), we divide $[-80^\circ, 80^\circ]$ into 210 equal intervals where each interval is taken as a class (i.e., $K=210$). When implementing CcGAN, the three hyper-parameters of HVDL and SVDL are selected by the rule of thumb formulae in Remark \ref{rmk:rule_of_thumb}, i.e., $\sigma\approx 0.029$, $\kappa\approx 0.032$ and $\nu\approx1000.438$. All GANs are trained for 20,000 iterations. To evaluate the candidate models, we choose 2,000 evenly spaced angles in $[-80^\circ, 80^\circ]$ and generate 50 images from each candidate GAN model for each of these angles. Please see Supp. \ref{supp:details_of_steeringangle} for the network architectures and more details about the training/testing setup.

	{\setlength{\parindent}{0cm}\textbf{Quantitative and visual results:}} To evaluate the quality of fake images, we use the proposed \textit{Sliding Fr\'echet Inception Distance} (SFID) as the overall metric instead of Intra-FID, since we have very few real images for many angles (e.g., angles close to the two end points of $[-80^\circ, 80^\circ]$). We preset 1,000 SFID centers in $[-80^\circ, 80^\circ]$ and let the SFID radius be $2^\circ$. NIQE, Diversity (entropy of predicted types of scenes), and Label Score are also reported. Please see Supp. \ref{supp:steeringangle_performance} for more details of these performance measures. 
	
\begin{figure}[!htbp]
		\centering
		\includegraphics[width=0.8\linewidth]{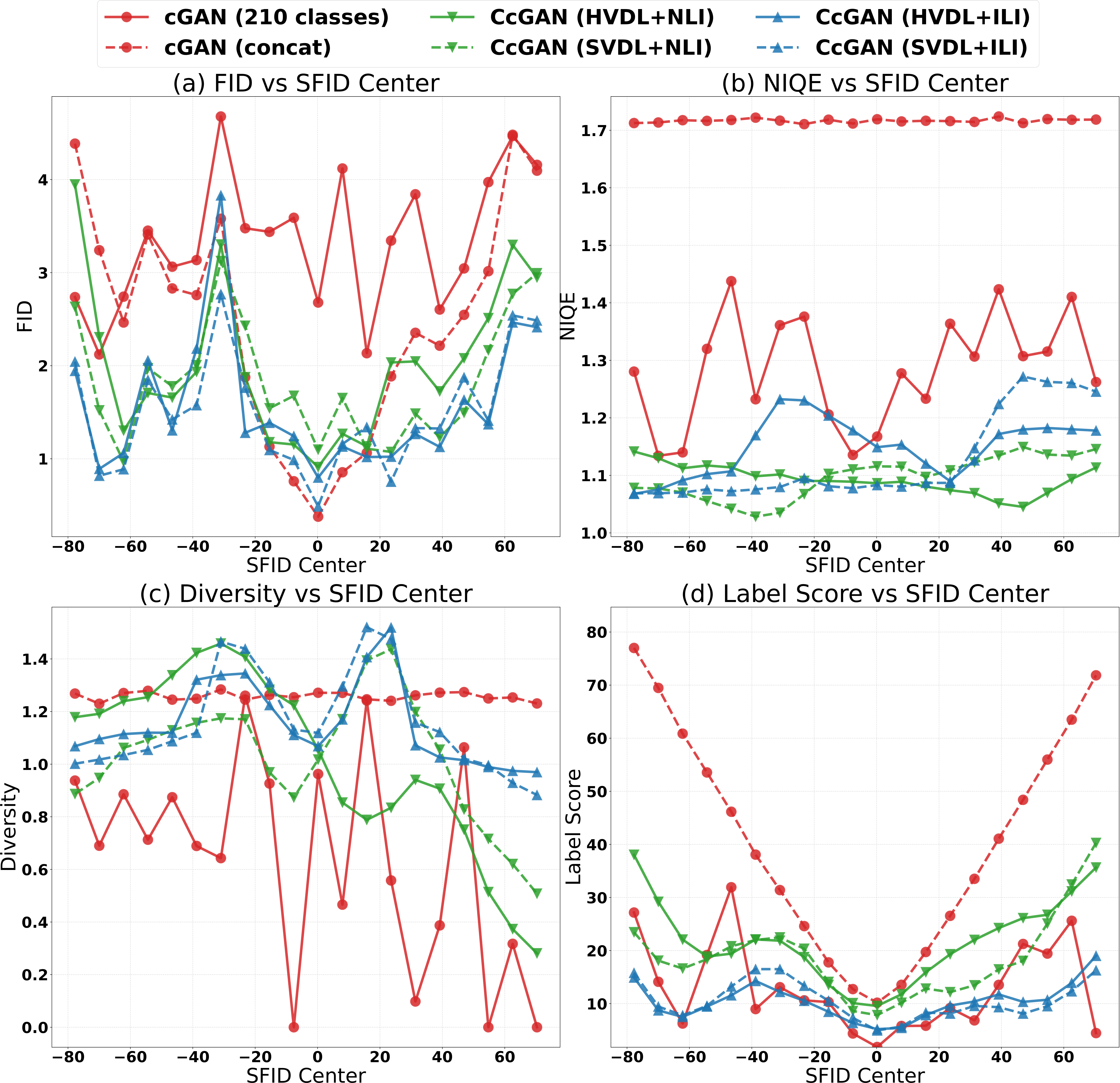} 
		\caption{\textbf{Line graphs of FID/NIQE/Diversity/Lable Score versus SFID Center for the Steering Angle dataset.} To plot these line graphs, we compute these metrics within each SFID interval defined by the corresponding SFID center. Figs. (a) to (d) show that, although the NLI-based CcGANs do not have good label consistency, the ILI-based CcGANs substantially outperform cGAN (210 classes) in most SFID intervals. All graphs of CcGANs also appear much smoother than those of cGAN (210 classes) because of HVDL and SVDL. Figs. (b) to (c) show that athough cGAN (concat) has the highest Diversity score, it also has the worst NIQE score and Label Score.}
		\label{fig:SteeringAngle_line_graphs}
	\end{figure}

	We report in Table\ \ref{tab:low_resolution_results} the average quality of 100,000 fake images from each candidate method. Some example fake images are also shown in Fig.\ \ref{fig:SteeringAngle_visual_results} in Appendix. We also compute FID, NIQE, Diversity, and Label Score in each SFID interval and plot the line graphs of FID/NIQE/Diversity/Label Score versus SFID Center in Fig.\ \ref{fig:SteeringAngle_line_graphs}. Based on these quantitative and visual results, we can conclude:
	\begin{itemize}
		\item The two ILI-based CcGAN methods are better than cGAN (210 classes) in terms of all four metrics; however, the two NLI-based CcGAN methods have lower label consistency than cGAN (210 classes). Although cGAN (concat) has the highest Diversity score, the four CcGAN methods outperform cGAN (concat) in terms of the other three metrics. 
		\item Although the NIQE score and Label Score of cGAN (210 classes) are not grossly uncompetitive, cGAN (210 classes) has a very low Diversity score and Fig.\ \ref{fig:SteeringAngle_line_graphs}(c) shows that the Diversity scores are almost zero at some angles. Example fake images in Fig.\ \ref{fig:SteeringAngle_visual_results} also show that cGAN (210 classes) has the mode collapse problem \cite{srivastava2017veegan, pmlr-v70-arjovsky17a, chang2018escaping} (i.e., it always generates the same image for some angles).
		\item Although cGAN (concat) has the highest Diversity score, its NIQE score and Label Score are terrible, implying bad visual quality and low label consistency. Example fake images in Fig.\ \ref{fig:SteeringAngle_visual_results} support these quantitative results. 
		\item The line graphs in Fig.\ \ref{fig:SteeringAngle_line_graphs} show that the performance of cGAN (210 classes) is very unstable across all SFID intervals. In contrast, cGAN (concat) has very smooth graphs but its graph for Label Score is above all other graphs.
		\item The two ILI-based CcGANs perform better than the two NLI-based CcGANs in terms of all metrics except NIQE. 
	\end{itemize}

	
	\subsection{Effectiveness of SFID}\label{sec:effectiveness_SFID} 
	In this section, we study the effectiveness of SFID on RC-49. Since RC-49 has a large enough sample size of real images, we can get a reliable Intra-FID. At the same time, we can also deliberately reduce the sample size of real images to mimic the scenario where a reliable Intra-FID is not applicable but SFID still works well. The experiment in this section can also be conducted on Cell-200 but is omitted in this paper. We study 11 combinations of the SFID radius ($r_\text{SFID}$) and the number of SFID centers (\# $c_\text{SFID}$) in this experiment where we use SFID to evaluate cGAN and two CcGAN methods (i.e., HVDL+NLI and HVDL+ILI) pre-trained in Section \ref{sec:rc49}. In Setting 1, we let $r_\text{SFID}=0$ so SFID degenerates to Intra-FID. In the same setting, we evaluate the three GANs on all 899 distinct angles and all real images in RC-49 are used to compute Intra-FID, so Setting 1 is taken as the oracle in this experiment. In Setting 2, we also let $r_\text{SFID}=0$ so SFID degenerates to Intra-FID again. In Setting 2, however, we only evaluate GANs on the 450 distinct angles which are seen in the training set of the experiment in Section \ref{sec:rc49}. Moreover, to simulate the scenario where we have very few real images to compute Intra-FID, we deliberately reduce the number of real images at each distinct angle from 49 to 10. Therefore, in Setting 2, there are 10 real images for each angle seen in the training set and 0 real image for each angle unseen in the training set. Setting 2 is treated as the baseline in this experiment. Settings 3 to 11 are designed to show the effectiveness of SFID so we let $r_\text{SFID}>0$. Similar to Setting 2, from Settings 3 to 11, real images are available only for those 450 distinct angles seen in the training set and only 10 real images are available for each angle. We consider three values for $r_\text{SFID}$ (i.e., 0.5, 1, and 2) and three values for the number of $c_\text{SFID}$'s (i.e., 400, 600, and 800). 
	
	For all settings, we compute one FID in each SFID interval (in Settings 1 and 2, the SFID interval degenerates to an angle) and report in Table\ \ref{tab:effectiveness_of_SFID} the mean of these FIDs along with their standard deviation after the ``$\pm$" symbol. Setting 1 is the oracle setting whose evaluation results can be seen as the ground truth, and we hope the evaluation results of SFID are close to Setting 1. In Setting 2, when we have very few real images (even zero) for each angle, Intra-FID overestimates the average FID of each GAN (e.g., from 1.7201 to 1.9664 for cGAN) and underestimates the quantitative difference between cGAN and CcGANs (e.g., from $(1.7201-0.6119)/1.7201\approx 64.4\%$ to $(1.9664-1.2102)/1.9664\approx 38.5\%$ for cGAN and HVDL+NLI). However, the performance of our proposed SFID in Settings 3 to 11 is very close to the oracle setting. If we compare within Settings 3 to 11, we can see $r_\text{SFID}$ is inversely proportional to the SFID score while \# $c_{SFID}$ does not have obvious influence on SFID. From Table\ \ref{tab:effectiveness_of_SFID}, we may conclude that as long as $r_\text{SFID}$ is set at a moderate level, SFID is a valid proxy to the oracle Intra-FID when there are insufficient real images to compute Intra-FID.
	
	\begin{table*}[!h] 
		\centering
		\caption{\textbf{Evaluation results of SFID on RC-49 under different setups of $r_\text{SFID}$ and number of $c_\text{SFID}$'s.} In the first two settings, SFID degenerates to Intra-FID since $r_\text{SFID}=0$. In Setting 1, we evaluate GANs on all 899 distinct angles and all real images are used to compute Intra-FID, so Setting 1 is the oracle setting. In Setting 2, we evaluate GANs on the 450 angles seen in the training set and, for each angle, 10 real images are used to compute Intra-FID, so Setting 2 is treated as the baseline. The performance of our proposed SFID (Settings 3 to 11) is close to the oracle setting while Intra-FID (Setting 2) tends to overestimate the average FID and underestimate the quantitative difference between cGAN and CcGANs.}
		\begin{adjustbox}{width=0.7\textwidth}
			\begin{tabular}{cccccc}
				\toprule
				\textbf{Setting} & $\bm{r_\text{SFID}}$ & \# $\bm{c_\text{SFID}}$ & \textbf{cGAN (150 classes)} & \textbf{HVDL+NLI} & \textbf{HVDL+ILI} \\
				\midrule
				1 {\tiny (Oracle)} & 0     & 899   & $1.720\pm 0.384$ & $0.612\pm 0.145$ & $0.424\pm 0.081$ \\
				2 {\tiny (Baseline)} & 0   & 450   & $1.966\pm 0.497$ & $1.210\pm 0.298$ & $1.032\pm 0.304$ \\ \hline
				3     & 0.5   & 400   & $1.764\pm 0.401$ & $0.675\pm 0.145$ & $0.426\pm 0.087$ \\
				4     & 1     & 400   & $1.726\pm 0.379$ & $0.618\pm 0.123$ & $0.412\pm 0.079$ \\
				5     & 2     & 400   & $1.685\pm 0.356$ & $0.581\pm 0.104$ & $0.401\pm 0.073$ \\
				6     & 0.5   & 600   & $1.764\pm 0.399$ & $0.675\pm 0.144$ & $0.425\pm 0.087$ \\
				7     & 1     & 600   & $1.727\pm 0.378$ & $0.618\pm 0.123$ & $0.412\pm 0.078$ \\
				8     & 2     & 600   & $1.685\pm 0.357$ & $0.582\pm 0.104$ & $0.401\pm 0.073$ \\
				9     & 0.5   & 800   & $1.765\pm 0.399$ & $0.675\pm 0.144$ & $0.425\pm 0.087$ \\
				10    & 1     & 800   & $1.726\pm 0.378$ & $0.618\pm 0.123$ & $0.412\pm 0.078$ \\
				11    & 2     & 800   & $1.686\pm 0.357$ & $0.582\pm 0.104$ & $0.401\pm 0.073$ \\
				\bottomrule
			\end{tabular}%
		\end{adjustbox}
		\label{tab:effectiveness_of_SFID}%
	\end{table*}%

	\subsection{Evaluation in Terms of IS and FID}\label{sec:evaluation_FID_IS}
	For completeness, we also report in Table~\ref{tab:low_resolution_results} the FID and IS scores of fake images generated from candidate methods. However, we emphasize that IS and FID cannot measure intra-label diversity and label consistency since computing IS and FID does not need the actual and assigned labels of fake images. Nonetheless, we see that CcGANs, especially ILI-based CcGANs, outperform the two conventional cGANs in terms of IS and FID too. Please see Supp.\ \ref{supp:exp_results_fid_is} for detailed setups of this evaluation and more illustrations.

	\subsection{Comparison Against State of The Art cGANs}\label{sec:compare_against_sota}
	In Sections~\ref{sec:rc49} to \ref{sec:steeringangle}, to make the comparison fair and focus attention on the effectiveness of the proposed loss functions and label input mechanisms, both conventional cGANs and CcGANs adopt the same network architectures (e.g., SNGAN) and training techniques (e.g., with or without DiffAugment \cite{zhao2020differentiable}). Unlike the above experiments, in this section, we aim to show that CcGAN (SVDL+ILI) can still outperform state of the art cGANs including SNGAN \cite{miyato2018spectral}, SAGAN \cite{SAGAN-zhang19d}, BigGAN \cite{brock2018large}, CR-BigGAN \cite{Zhang2020Consistency}, BigGAN+DiffAugment \cite{zhao2020differentiable}, and ReACGAN \cite{reacgan2021}, which are equipped with advanced network architectures and training techniques.
	
	Among these state of the art cGANs, SNGAN \cite{miyato2018spectral} and SAGAN \cite{SAGAN-zhang19d} propose new network architectures. BigGAN \cite{brock2018large} proposes not only the BigGAN architecture but also a bag of effective techniques for training cGANs. CR-BigGAN \cite{Zhang2020Consistency} proposes consistency regularization for BigGAN's training. BigGAN+DiffAugment \cite{zhao2020differentiable} applies the DiffAugment technique to stabilize cGANs' training. DiffAugment is applicable to CcGAN too. ReACGAN \cite{reacgan2021} introduces novel training techniques for ACGAN \cite{odena2017conditional}, which makes ACGAN state of the art again.
	
	We want to emphasize that, although the above state of the art cGANs work well in the class-conditional CGM tasks, they cannot model the distribution of images conditional on regression labels due to lack of a suitable label input mechanism. To make them fit into the regression scenario, we apply the binning strategy in Section~\ref{sec:rc49} to convert rotation angles into class labels and then train them on RC-49 in a class-conditional manner. For the implementation of CcGAN (SVDL+ILI), unlike the setup in Section~\ref{sec:rc49}, we test with more advanced network architectures including SAGAN, and BigGAN. We also incorporate DiffAugment into CcGAN training, i.e., SAGAN+DiffAugment and BigGAN+DiffAugment. Similar to Section~\ref{sec:rc49}, we evaluate candidate models on 899 distinct angles by generating 200 fake images for each angle. Please see Supp.~\ref{supp:comparison_against_sota} for detailed training and evaluation setups.
	
	Comparison results are summarized in Table~\ref{tab:compare_against_sota}. We can see class-conditional SNGAN, SAGAN, and BigGAN suffer from mode collapse problems due to their very low Diversity. CR-BigGAN has higher Diversity but very low label consistency. The DiffAugment technique can substantially improve BigGAN's performance, but it is still much worse than CcGAN (SVDL+ILI). ReACGAN performs best among class-conditional GANs, but it is outperformed by CcGAN (SVDL+ILI) which is equipped with the SAGAN/BigGAN architecture and DiffAugment. These results demonstrate that, without HVDL/SVDL and NLI/ILI, existing architectures or training techniques cannot solve \textbf{(P1)} and \textbf{(P2)}.

	\begin{table*}[!h]
		\centering
		\caption{ \textbf{Comparison against state of the art cGANs on RC-49.} In this table, we show the average quality of 179,800 fake images from class-conditional GANs and CcGAN (SVDL+ILI) with the standard deviation after the ``$\pm$" symbol. IS and FID scores are also reported for completeness. ``$\downarrow$" (``$\uparrow$") indicates lower (higher) values are preferred. The best and second best results are marked in gray and green respectively. }
		\begin{adjustbox}{width=1\textwidth}
		\begin{tabular}{l|l|cccc|cc}
			\toprule
			\textbf{Model} & \textbf{ \begin{tabular}[l]{@{}l@{}} Network Architecture \\ and Training Technique \end{tabular} } & \textbf{Intra-FID} $\downarrow$ & \textbf{NIQE} $\downarrow$  & \textbf{Diversity} $\uparrow$ & \textbf{Label Score} $\downarrow$ & \textbf{IS} $\uparrow$ & \textbf{FID} $\downarrow$ \\
			\midrule
			\multirow{6}[2]{*}{cGAN (150 classes)} & SNGAN \cite{miyato2018spectral} (2017) & $1.720 \pm 0.384$ & $2.731 \pm 0.162$ & $0.779 \pm 0.199$ & $4.815 \pm 5.152$ & 2.382  & 1.066  \\
			& SAGAN \cite{SAGAN-zhang19d} (2018) & $1.288 \pm 0.223$ & $2.517 \pm 0.257$ & $0.898 \pm 0.372$ & $36.388 \pm 22.619$ & 1.950  & 1.061  \\
			& BigGAN \cite{brock2018large} (2019) & $2.193 \pm 0.372$ & $2.399 \pm 0.201$ & $0.543 \pm 0.315$ & $4.467 \pm 3.008$ & 1.615  & 1.375  \\
			& CR-BigGAN \cite{Zhang2020Consistency} (2020) & $0.981 \pm 0.239$ & $2.570 \pm 0.142$ & $2.000 \pm 0.119$ & $21.530 \pm 17.861$ & 3.565  & 0.479  \\
			& BigGAN+DiffAugment \cite{zhao2020differentiable} (2020) & $0.545 \pm 0.206$ & $2.138 \pm 0.120$ & $3.250 \pm 0.051$ & $9.595 \pm 3.896$ & 25.400  & 0.119  \\
			& ReACGAN \cite{reacgan2021} (2021) & $0.178 \pm 0.036$ & $1.938 \pm 0.141$ & $3.145 \pm 0.028$ & \cellcolor{gray!25}{$1.422 \pm 1.337$} & 21.945  & 0.056  \\
			\midrule
			\multirow{3}[2]{*}{CcGAN (SVDL+ILI)} & SNGAN & $0.389 \pm 0.095$ & \cellcolor{gray!25}{$1.783 \pm 0.173$} & $2.949 \pm 0.069$ & \cellcolor{green!25}{$1.940 \pm 1.489$} & 20.173  & 0.197  \\
			& SAGAN+DiffAugment & \cellcolor{green!25}{$0.106 \pm 0.029$} & \cellcolor{green!25}{$1.784 \pm 0.173$} & \cellcolor{green!25}{$3.590 \pm 0.041$} & $2.431 \pm 0.449$ & \cellcolor{gray!25}{36.546}  & \cellcolor{green!25}{0.026}  \\
			& BigGAN+DiffAugment & \cellcolor{gray!25}{$0.086 \pm 0.028$} & $1.950 \pm 0.166$ & \cellcolor{gray!25}{$3.620 \pm 0.090$} & $2.468 \pm 1.920$ & \cellcolor{green!25}{30.686}  & \cellcolor{gray!25}{0.013}  \\
			\bottomrule
		\end{tabular}%
		\end{adjustbox}
		\label{tab:compare_against_sota}%
	\end{table*}%

	\section{High-resolution Experiments}\label{sec:experiment_hd}
	
	Besides the low-resolution experiments in Section \ref{sec:experiment}, we also compare CcGAN (SVDL+ILI) with cGAN ($K$ classes) and cGAN (concat) on high-resolution images. We consider three datasets with different resolutions, i.e., RC-49 ($128\times 128$ and $256\times 256$), UTKFace ($128\times 128$ and $192\times 192$), and Steering Angle ($128\times 128$). Since high-resolution experiments are more challenging than low-resolution ones, we make some changes to the setups in Section \ref{sec:experiment} to improve GANs' performance. First, all candidates use the more advanced SAGAN \cite{SAGAN-zhang19d} architecture. Second, cGAN ($K$ classes) and cGAN (concat) are trained with the hinge loss \cite{lim2017geometric}. CcGAN (SVDL+ILI) is trained with the reformulated hinge loss (see Eq.\ \eqref{eq:SVDL_hinge} in Supp.\ \ref{supp:reformulated_hinge_loss}). Third, DiffAugment \cite{zhao2020differentiable}, a recently proposed technique for training GANs with few samples, is also applied to improve the performance of the candidate methods.
		
	Both visual and quantitative results demonstrate that the high-resolution fake images generated from CcGAN are visually realistic, diverse, and label consistent. These results also show that CcGAN is compatible with state-of-the-art GAN architectures and training techniques. Furthermore, the failure patterns of cGAN ($K$ classes) and cGAN (concat) in this experiment are consistent with those in Section \ref{sec:experiment}. cGAN ($K$ classes) tends to have high label consistency but bad visual quality and low diversity. Oppositely, cGAN (concat) often has high diversity but bad/fair visual quality and terrible label consistency.

	\begin{table*}[!h]
		\centering
		\caption{\textbf{Average quality of high-resolution fake images from cGAN and CcGAN with the standard deviation after the ``$\pm$" symbol.} We generate 179800, 60000, and 100000 fake images via each candidate method for the RC-49, UTKFace, and Steering Angle experiments, respectively. These fake images are evaluated under four metrics: Intra-FID, NIQE, Diversity, and Label Score. ``$\downarrow$" (``$\uparrow$") indicates lower (higher) values are preferred. The best result are marked in gray.}
		\begin{adjustbox}{width=0.75\textwidth}
			\begin{tabular}{c|c|cccc}
				\toprule
				\textbf{Dataset} & \textbf{Model} & \textbf{Intra-FID} $\downarrow$ & \textbf{NIQE} $\downarrow$  & \textbf{Diversity} $\uparrow$ & \textbf{Label Score} $\downarrow$ \\
				\midrule
				\multicolumn{1}{c|}{\multirow{3}[0]{*}{\begin{tabular}[c]{@{}c@{}} \textbf{RC-49} \\ ($128\times 128$) \end{tabular}}} & cGAN (150 classes) & $1.250 \pm 0.492$ & $2.293 \pm 0.133$ & $2.341 \pm 0.224$ & \cellcolor{gray!25}{${2.032 \pm 1.653}$} \\
				& cGAN (concat) & $1.128 \pm 0.166$ & $2.104 \pm 0.104$ & $3.431 \pm 0.039$ & $29.414 \pm 7.052$ \\
				& CcGAN (SVDL+ILI) & \cellcolor{gray!25}{${0.111 \pm 0.033}$} & \cellcolor{gray!25}{${1.775 \pm 0.051}$} & \cellcolor{gray!25}{${3.552 \pm 0.047}$} & $2.643 \pm 2.077$ \\
				\midrule
				\multicolumn{1}{c|}{\multirow{3}[0]{*}{\begin{tabular}[c]{@{}c@{}} \textbf{RC-49} \\ ($256\times 256$) \end{tabular}}} & cGAN (150 classes) & $1.224 \pm 0.336$ & $2.147 \pm 0.085$ & $2.462 \pm 0.095$ & \cellcolor{gray!25}{${2.790 \pm 2.852}$} \\
				& cGAN (concat) & $1.325 \pm 0.227$ & $3.153 \pm 0.122$ & \cellcolor{gray!25}{${3.120 \pm 0.043}$} & $28.776 \pm 20.273$ \\
				& CcGAN (SVDL+ILI) & \cellcolor{gray!25}{${0.495 \pm 0.139}$} & \cellcolor{gray!25}{${1.655 \pm 0.070}$} & $2.844 \pm 0.101$ & $3.260 \pm 2.641$ \\
			    \midrule
				\multicolumn{1}{c|}{\multirow{3}[0]{*}{\begin{tabular}[c]{@{}c@{}} \textbf{UTKFace} \\ ($128\times 128$) \end{tabular}}} & cGAN (60 classes) & $1.195 \pm 0.356$ & $1.381 \pm 0.208$ & $0.788 \pm 0.425$ & \cellcolor{gray!25}{${6.150 \pm 5.268}$} \\
				 & cGAN (concat) & $0.408 \pm 0.144$ & $1.377 \pm 0.079$ & \cellcolor{gray!25}{${1.332 \pm 0.026}$} & $18.064 \pm 12.550$ \\
				 & CcGAN (SVDL+ILI) & \cellcolor{gray!25}{${0.367 \pm 0.123}$} & \cellcolor{gray!25}{${1.113 \pm 0.033}$} & $1.199 \pm 0.232$ & $7.747 \pm 6.580$ \\
				\midrule
				\multicolumn{1}{c|}{\multirow{3}[0]{*}{\begin{tabular}[c]{@{}c@{}} \textbf{UTKFace} \\ ($192\times 192$) \end{tabular}}} & cGAN (60 classes) & $0.908 \pm 0.327$ & $1.755 \pm 0.215$ & $1.047 \pm 0.381$ & \cellcolor{gray!25}{${6.639 \pm 5.686}$} \\
				& cGAN (concat) & $0.591 \pm 0.214$ & $2.352 \pm 0.126$ & \cellcolor{gray!25}{${1.358 \pm 0.019}$} & $17.116 \pm 11.652$ \\
				& CcGAN (SVDL+ILI) & \cellcolor{gray!25}{${0.499 \pm 0.186}$} & \cellcolor{gray!25}{${1.661 \pm 0.047}$} & $1.207 \pm 0.260$ & $7.885 \pm 6.272$ \\
				\midrule
				\multicolumn{1}{c|}{\multirow{3}[0]{*}{\begin{tabular}[c]{@{}c@{}} \textbf{Steering Angle} \\ ($128\times 128$) \end{tabular}}} & cGAN (210 classes) & $4.963 \pm 0.916$ & $2.520 \pm 0.282$ & $0.564 \pm 0.401$ & $31.756 \pm 23.005$ \\
				& cGAN (concat) & $2.140 \pm 0.821$ & $2.542 \pm 0.006$ & \cellcolor{gray!25}{${1.292 \pm 0.014}$} & $42.757 \pm 27.341$ \\
				& CcGAN (SVDL+ILI) & \cellcolor{gray!25}{${1.689 \pm 0.443}$} & \cellcolor{gray!25}{${2.411 \pm 0.100}$} & $1.088 \pm 0.243$ & \cellcolor{gray!25}{${18.438 \pm 16.072}$} \\
				\bottomrule
			\end{tabular}%
		\end{adjustbox}
		\label{tab:high_resolution_results}%
	\end{table*}%
	
	\begin{figure*}[!h]
		\centering
		\includegraphics[width=0.65\textwidth]{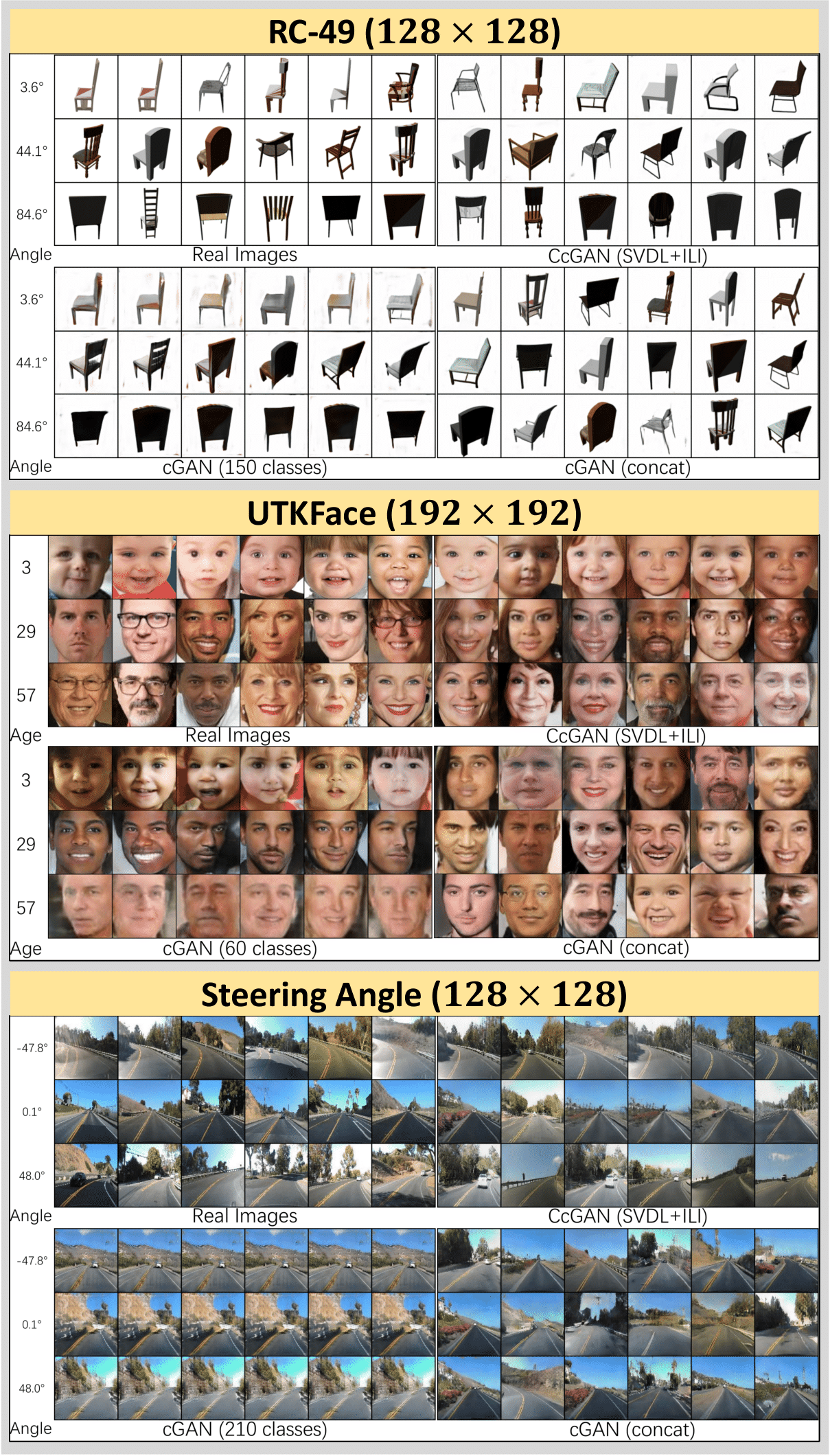}
		\caption{\textbf{Some example high-resolution images for the RC-49, UTKFace, and Steering Angle experiments, respectively.} In the RC-49 experiment, the fake images from CcGAN are almost indistinguishable from real images. Conversely, cGAN (150 classes) has bad visual quality and low diversity. cGAN (concat) has fair visual quality and poor label consistency. In the UTKFace experiment, CcGAN can generate visually realistic, diverse, and label consistent images. Fake images from cGAN (60 classes) are visually poor and lack diversity (e.g., the last row only has white male). cGAN (concat) fails to condition the image generation on age (e.g., the first row has many adults). In the Steering Angle experiment, CcGAN substantially outperforms both cGANs. Notably, cGAN (210 classes) has the mode collapse problem \cite{srivastava2017veegan, pmlr-v70-arjovsky17a, chang2018escaping} on this dataset.}
		\label{fig:hd_visual_results}
	\end{figure*}

	
	\subsection{High-resolution RC-49}\label{sec:rc49_hd}
	In this experiment, we test three candidate methods on RC-49 with two resolutions, i.e., $128\times 128$ and $256\times 256$. Most training setups are consistent with Section \ref{sec:rc49} and please see Supp.\ \ref{supp:rc-49_hd_setups} for details. The quantitative and visual results are shown in Table~\ref{tab:high_resolution_results} and Fig.~\ref{fig:hd_visual_results}. We can see CcGAN can generate high-quality images and the example fake images in Fig.~\ref{fig:hd_visual_results} are indistinguishable from real images. However, cGAN (150 classes) and cGAN (concat) fail again. Fake images generated from cGAN (150 classes) are visually unrealistic and less diverse. Conversely, cGAN (concat) can generate images with fair visual quality and high diversity, but it cannot control the image generation via conditioning angles.

	\subsection{High-resolution UTKFace}\label{sec:utkface_hd}
	In this experiment, we test three candidate methods on UTKFace with two resolutions, i.e., $128\times 128$ and $192\times 192$. Most training setups are consistent with Section \ref{sec:utkface} except that we let $\nu=900$ when implementing CcGAN (SVDL+ILI). Please see Supp.\ \ref{supp:utkface_hd_setups} for details. The quantitative and visual results are shown in Table~\ref{tab:high_resolution_results} and Fig.~\ref{fig:hd_visual_results}. We can see CcGAN substantially outperforms two cGANs. Fake images generated from cGAN (150 classes) are visually unrealistic and less diverse. cGAN (concat) cannot condition image generation on age.

	\subsection{High-resolution Steering Angle}\label{sec:steering_angle_hd}
	Although the low-resolution Steering Angle experiment is already challenging enough due to high imbalance, we further increase the image resolution to $128\times 128$, making the generative modeling more difficult. Most training setups are consistent with Section \ref{sec:steeringangle} and please see Supp.\ \ref{supp:steering_angle_hd_setups} for details. The quantitative and visual results are shown in Table~\ref{tab:high_resolution_results} and Fig.~\ref{fig:hd_visual_results}. Both visual and quantitative results of cGAN (210 classes) imply severe mode collapse problem~\cite{srivastava2017veegan, pmlr-v70-arjovsky17a, chang2018escaping}. Similar to previous experiments, cGAN (concat) has a very high Diversity score, but its label consistency is terrible. On the contrary, the proposed CcGAN performs well in all three evaluation perspectives.

	\section{Conclusion}\label{sec:discussion}
	
	We propose CcGAN in this paper for generative image modeling conditional on regression labels. In CcGAN, two novel empirical discriminator losses (HVDL and SVDL), a novel empirical generator loss and two novel label input mechanisms (NLI and ILI) are proposed to overcome the two problems of existing cGANs. The error bounds of a discriminator trained under HVDL and SVDL are studied in this work. Two new benchmark datasets, RC-49 and Cell-200, are created for the continuous scenario. A new evaluation metric, termed SFID, is also proposed to replace Intra-FID when there are insufficient real images. Finally we demonstrate the superiority of the proposed CcGAN to representative conventional cGANs on RC-49, UTKFace, Cell-200, and Steering Angle datasets with both low and high image resolutions.

	\section*{Acknowledgments}
	This work was supported by UBC ARC Sockeye, Compute Canada, and the Natural Sciences and Engineering Research Council of Canada (NSERC) under Grants CRDPJ 476594-14, RGPIN-2019-05019, and RGPAS2017-507965.

	\bibliographystyle{IEEEtran}
	\bibliography{./reference_CcGAN}

	\newpage
	
	\section*{Supplementary Material}
	\addcontentsline{toc}{section}{Supplementary Material}

	\renewcommand{\thesection}{S.\arabic{section}} 
	\renewcommand{\thesubsection}{\thesection.\arabic{subsection}}
	\renewcommand\thefigure{\thesection.\arabic{figure}}
	\renewcommand\thetable{\thesection.\arabic{table}}
	\renewcommand{\theequation}{S.\arabic{equation}}
	\renewcommand{\thetheorem}{S.\arabic{theorem}} 
	\renewcommand{\thedefinition}{S.\arabic{definition}} 
	\renewcommand{\thelemma}{S.\arabic{lemma}} 
	\renewcommand{\theremark}{S.\arabic{remark}}

	\section{GitHub repository}\label{supp:codes}
	Please find the codes for this paper at
	\begin{center}
		\url{https://github.com/UBCDingXin/improved_CcGAN}
	\end{center}

	\section{Algorithms for CcGAN training}\label{supp:algorithm_train}

	\begin{algorithm}[!ht]
		\footnotesize
		\SetAlgoLined
		\KwData{$N^r$ real image-label pairs $\Omega^r=\{(\bm{x}^r_1, y^r_1),\dots,(\bm{x}^r_{N^r}, y^r_{N^r})\}$, $N_{\text{uy}}^r$ ordered distinct labels $\Upsilon=\{y_{[1]}^r, \dots, y_{[N_{\text{uy}}^r]}^r \}$ in the dataset, preset $\sigma$ and $\kappa$, number of iterations $K$, the discriminator batch size $m^d$, and the generator batch size $m^g$.}
		\KwResult{Trained generator $G$.} 
		\For{$k=1$ \KwTo $K$}{
			\textbf{Train D}\;
			Draw $m^d$ labels $Y^d$ with replacement from $\Upsilon$\;
			Create a set of target labels $Y^{d,\epsilon}=\{ y_i+\epsilon| y_i\in Y^d, \epsilon\in \mathcal{N}(0,\sigma^2), i=1,\dots,m^d \}$ ($D$ is conditional on these labels)  \;
			Initialize $\Omega_d^r=\phi, \Omega_d^f=\phi$\;
			\For{$i=1$ \KwTo $m^d$}{
				Randomly choose an image-label pair $(\bm{x},y) \in \Omega^r$ satisfying $|y-y_i-\epsilon|\leq\kappa$ where $y_i+\epsilon\in Y^{d,\epsilon}$ and let $\Omega_d^r=\Omega_d^r \cup (\bm{x}, y_i+\epsilon)$. \;
				Randomly draw a label $y^\prime$ from $U(y_i+\epsilon-\kappa, y_i+\epsilon+\kappa)$ and generate a fake image $\bm{x}^\prime$ by evaluating $G(\bm{z}, y^\prime)$, where $\bm{z}\sim \mathcal{N}(\bm{0},\bm{I})$. Let $\Omega_d^f=\Omega_d^f \cup (\bm{x}^\prime, y_i+\epsilon)$. \;
			}
			Update $D$ with samples in set $\Omega_d^r$ and $\Omega_d^f$ via gradient-based optimizers based on Eq.~\eqref{eq:HVDL}\; 
			
			\textbf{Train G}\;
			Draw $m^g$ labels $Y^g$ with replacement from $\Upsilon$\;
			Create another set of target labels $Y^{g,\epsilon}=\{ y_i+\epsilon| y_i\in Y^g, \epsilon\in \mathcal{N}(0,\sigma^2), i=1,\dots,m^g \}$ ($G$ is conditional on these labels) \;
			Generate $m^g$ fake images conditional on $Y^{g,\epsilon}$ and put these image-label pairs in $\Omega_g^f$ \;
			Update $G$ with samples in $\Omega_g^f$ via gradient-based optimizers based on Eq.\eqref{eq:loss_G_CcGAN} \;
			
		}
		\caption{An algorithm for CcGAN training with the proposed HVDL.}
		\label{alg:train_ccgan_HVDL}
	\end{algorithm}

	\begin{algorithm}[!ht]
		\footnotesize
		\SetAlgoLined
		\KwData{$N^r$ real image-label pairs $\Omega^r=\{(\bm{x}^r_1, y^r_1),\dots,(\bm{x}^r_{N^r}, y^r_{N^r})\}$, $N_{\text{uy}}^r$ ordered distinct labels $\Upsilon=\{y_{[1]}^r, \dots, y_{[N_{\text{uy}}^r]}^r \}$ in the dataset, preset $\sigma$ and $\nu$, number of iterations $K$, the discriminator batch size $m^d$, and the generator batch size $m^g$.}
		\KwResult{Trained generator $G$.} 
		\For{$k=1$ \KwTo $K$}{
			\textbf{Train D}\;
			Draw $m^d$ labels $Y^d$ with replacement from $\Upsilon$\;
			Create a set of target labels $Y^{d,\epsilon}=\{ y_i+\epsilon| y_i\in Y^d, \epsilon\in \mathcal{N}(0,\sigma^2), i=1,\dots,m^d \}$ ($D$ is conditional on these labels)  \;
			Initialize $\Omega_d^r=\phi, \Omega_d^f=\phi$\;
			\For{$i=1$ \KwTo $m^d$}{
				Randomly choose an image-label pair $(\bm{x},y) \in \Omega^r$ satisfying $e^{-\nu (y-y_i-\epsilon)^2}>10^{-3}$ where $y_i+\epsilon\in Y^{d,\epsilon}$ and let $\Omega_d^r=\Omega_d^r \cup (\bm{x}, y_i+\epsilon)$. This step is used to exclude real images with too small weights. \;
				Compute $w_i^r(y, y_i + \epsilon)=e^{-\nu (y_i + \epsilon - y)^2 }$\;
				Randomly draw a label $y^\prime$ from $U(y_i+\epsilon-\sqrt{-\frac{\log 10^{-3}}{\nu}}, y_i+\epsilon+\sqrt{-\frac{\log 10^{-3}}{\nu}})$ and generate a fake image $\bm{x}^\prime$ by evaluating $G(\bm{z}, y^\prime)$, where $\bm{z}\sim \mathcal{N}(\bm{0},\bm{I})$. Let $\Omega_d^f=\Omega_d^f \cup (\bm{x}^\prime, y_i+\epsilon)$. \;
				Compute $w_i^g(y^\prime, y_i + \epsilon)=e^{-\nu (y_i + \epsilon - y^\prime)^2 }$\;
			}
			Update $D$ with samples in set $\Omega_d^r$ and $\Omega_d^f$ via gradient-based optimizers based on Eq.~\eqref{eq:SVDL}\; 
			
			\textbf{Train G}\;
			Draw $m^g$ labels $Y^g$ with replacement from $\Upsilon$\;
			Create another set of target labels $Y^{g,\epsilon}=\{ y_i+\epsilon| y_i\in Y^g, \epsilon\in \mathcal{N}(0,\sigma^2), i=1,\dots,m^g \}$ ($G$ is conditional on these labels) \;
			Generate $m^g$ fake images conditional on $Y^{g,\epsilon}$ and put these image-label pairs in $\Omega_g^f$ \;
			Update $G$ with samples in $\Omega_g^f$ via gradient-based optimizers based on Eq.\eqref{eq:loss_G_CcGAN} \;
			
		}
		\caption{An algorithm for CcGAN training with the proposed SVDL.}
		\label{alg:train_ccgan_SVDL}
	\end{algorithm}

	\begin{remark}
		If should be noted that, for computational efficiency, the normalizing constants $N_{y^r_j+\epsilon^r,\kappa}^r$, $N_{y^g_j+\epsilon^g,\kappa}^g$, $\sum_{i=1}^{N^r}w^r(y_i^r,y_j^r+\epsilon^r)$, and $\sum_{i=1}^{N^g}w^g(y_i^g,y_j^g+\epsilon^g)$ in Eq. \eqref{eq:HVDL} and \eqref{eq:SVDL} are excluded from the training and only used for theoretical analysis.
	\end{remark}

	\section{Theoretical Analysis for HVDL and SVDL}\label{supp:CcGAN_convergence_rate}
	
	In this section, we provide a self-contained theoretical analysis of HVDL and SVDL. To make the derivation clearer, we use some notations and definitions slightly different from those in the main content of the paper.  Some necessary assumptions, lemmas, and theorems are also introduced or derived in Sections \ref{supp:CcGAN_assumptions_bound} and \ref{supp:CcGAN_lemmas_theorems_bound}. The main theorems on the error bounds of $D$ are derived in Section \ref{supp:CcGAN_main_theorems_bound}.

	\subsection{Some Necessary Definitions and Notations}\label{supp:CcGAN_def_notation_bound}
	This section summarizes some necessary definitions and notations used in the derivation. Please note that \textbf{all these definitions and notations are valid in Supp.\ \ref{supp:CcGAN_convergence_rate} only}. 
	\begin{itemize}
		\item Unlike other contents of this paper, we use different symbols to denote random variables/vectors and the fixed values that random variables/vectors may take. Specifically, a random image and a random label are represented respectively by a bold capital $\bm{X}$ and a capital $Y$. A sequence of $N$ random image-label pairs are represented by $(\bm{X}_1, Y_1), \dots, (\bm{X}_N, Y_N)$. Please note that some subscripts or superscripts may apply to $\bm{X}$ and $Y$ to provide some extra information. An observed (fixed) image and an observed (fixed) label are denoted respectively by a bold lowercase $\bm{x}$ and a lowercase $y$. \textbf{Moreover, without loss of generality, we assume $Y,y\in[0,1]$, i.e., $\mathcal{Y}=[0,1]$}.
		
		\item Let $p(\bm{x}|Y=y)$ denote the conditional probability density function (PDF) of $\bm{X}$ given the occurrence of the value $y$ of $Y$. $p$ may have superscripts or subscripts to provide some extra information.
		
		\item Let $p(y^\prime|Y=y)$ denote the conditional PDF of $Y^\prime$ given the occurrence of the value $y$ of $Y$. $p$ may have superscripts or subscripts to provide some extra information.
		
		\item Let $\mathcal{D}$ stand for the \textit{Hypothesis Space} of $D$. $\mathcal{D}$ is a set of functions that can be represented by $D$ (a neural network with determined architecture but undetermined weights).
		
		\item Let $f(\bm{x},y)=-\log D(\bm{x},y)$ and $\mathcal{F}=-\log\mathcal{D}$. 
		
		\item Let $\hat{p}_r^{\text{KDE}}(y)$ and $\hat{p}_g^{\text{KDE}}(y)$ stand for the KDEs of $p_r(y)$ and $p_g(y)$ respectively.
		
		\item For HVDL, denote respectively by 
		$$p_r^{y,\kappa}(\bm{x}) \triangleq \int p_r(\bm{x} | Y=y^\prime) \frac{\mathbbm{1}_{\{ |y^\prime-y|\leq\kappa \}}p_r(y^\prime)}{\int\mathbbm{1}_{\{ |y^\prime-y|\leq\kappa \}}p_r(y^\prime)dy^\prime} dy^\prime $$ 
		and
		$$p_g^{y,\kappa}(\bm{x}) \triangleq \int p_g(\bm{x} | Y=y^\prime) \frac{\mathbbm{1}_{\{ |y^\prime-y|\leq\kappa \}}p_g(y^\prime)}{\int\mathbbm{1}_{\{ |y^\prime-y|\leq\kappa \}}p_g(y^\prime)dy^\prime} dy^\prime $$ 
		the PDFs of the marginal distributions for real and fake images with labels in $[y-\kappa, y+\kappa]$.

		\item For SVDL, given $y$ and the weight functions, if the number of real and fake images are infinite, the real and fake empirical densities converges to $$p_r^{y,w^r}(\bm{x})\triangleq\int p_r(\bm{x}|Y=y^\prime)\frac{w^r(y^\prime,y)p_r(y^\prime)}{W^r(y)}dy^\prime$$ and $$p_g^{y,w^g}(\bm{x})\triangleq\int p_g(\bm{x}|Y=y^\prime) \frac{w^g(y^\prime,y)p_g(y^\prime)}{W^g(y)}dy^\prime$$ respectively, where $$W^r(y)\triangleq\int w^r(y^\prime,y)p_r(y^\prime)dy^\prime,$$ $$W^g(y)\triangleq\int w^g(y^\prime,y)p_g(y^\prime)dy^\prime,$$ and $w^r$ and $w^g$ are the weight functions defined as follows
		$$w^r(y^\prime, y) =  e^{-\nu(y^\prime-y)^2} \quad \text{and} \quad  w^g(y^\prime, y)  =  e^{-\nu(y^\prime-y)^2}.$$
		We also let $$p^r_w(y^\prime|Y=y)\triangleq \frac{w^r(y^\prime,y)p^r(y^\prime)}{W^r(y)}$$ and $$p^g_w(y^\prime|Y=y)\triangleq \frac{w^g(y^\prime,y)p^g(y^\prime)}{W^g(y)}.$$
		
		\item The H\"older Class defined below is a set of functions with bouned second derivatives, which controls the variation of the function when parameter changes. 
		\begin{definition} 
			\label{supp_def:holder_class}
			(H\"older Class) Define the H\"older class of functions as:
			\begin{equation}
				\Sigma(L)\triangleq\left\{p: \forall t_1, t_2\in\mathcal{Y}, \exists L >0, s.t. \frac{|p^\prime(t_1)-p^\prime(t_2)|}{|t_1-t_2|}\leq L \right\}.
			\end{equation}
		\end{definition}

		\item With some new notations above, we restate the theoretical discriminator losses $\mathcal{L}(D)$ as follows:
		\begin{align}
			\mathcal{L}(D) = & -\mathbb{E}_{Y\sim p_r(y)}\left[\mathbb{E}_{\bm{X}\sim p_r(\bm{x}|Y)}\left[ \log \left( D(\bm{X},Y) \right) \right] \right] \nonumber \\
			&  - \mathbb{E}_{Y\sim p_g(y)}\left[\mathbb{E}_{\bm{X}\sim p_g(\bm{x}|Y)}\left[ \log \left( 1 - D(\bm{X},Y) \right) \right] \right], 
			\label{eq:CcGAN_loss_cD_restate}
		\end{align}
		
		\item Recall that, given a $G$, the optimal discriminator which minimizes $\mathcal{L}(D)$ is in the form of 
		\begin{equation*}
			D^*(\bm{x},y)=\frac{p_r(\bm{x},y)}{p_r(\bm{x},y)+p_g(\bm{x},y)}.
		\end{equation*}
		However, $D^*$ may not be covered by the hypothesis space $\mathcal{D}$. Define $\widetilde{D}$, $\widehat{D}^{\text{HVDL}}$, and $\widehat{D}^{\text{SVDL}}$ as follows
		\begin{align}
			& \widetilde{D}\triangleq{\arg\min}_{D\in\mathcal{D}}\mathcal{L}(D), \nonumber\\
			& \widehat{D}^{\text{HVDL}}\triangleq{\arg\min}_{D\in\mathcal{D}}\widehat{\mathcal{L}}^{\text{HVDL}}(D), \nonumber\\
			& \widehat{D}^{\text{SVDL}}\triangleq{\arg\min}_{D\in\mathcal{D}}\widehat{\mathcal{L}}^{\text{SVDL}}(D). \nonumber
		\end{align}
		Note that $\mathcal{L}(\widetilde{D})-\mathcal{L}(D^*)$ should be a non-negative constant. In CcGAN, we minimize $\widehat{\mathcal{L}}^{\text{HVDL}}(D)$ or $\widehat{\mathcal{L}}^{\text{SVDL}}(D)$ with respect to $D\in\mathcal{D}$, so we are interested in the distance of $\widehat{D}^{\text{HVDL}}$ and $\widehat{D}^{\text{SVDL}}$ from $D^*$, i.e., $\mathcal{L}(\widehat{D}^{\text{HVDL}})-\mathcal{L}(D^*)$ and $\mathcal{L}(\widehat{D}^{\text{SVDL}})-\mathcal{L}(D^*)$. 
	\end{itemize}
	
	\subsection{Some Necessary Assumptions}\label{supp:CcGAN_assumptions_bound}
	In this theoretical analysis, we work with the following assumptions:
	
	{\setlength{\parindent}{0cm}\textbf{(A1)}} All $D$'s in $\mathcal{D}$ are measurable and uniformly bounded. Let 
	$$U \triangleq  \max \{  \sup_{D\in \mathcal{D}} \left[-\log D\right], \allowbreak \sup_{D\in \mathcal{D}} \left[ -\log (1-D) \right] \}$$ 
	and $U < \infty$;
	
	{\setlength{\parindent}{0cm}\textbf{(A2)}} For $\forall \bm{x}\in\mathcal{X}$ and $y, y^\prime\in\mathcal{Y}$, $\exists g^r(\bm{x}) >0 $ and $ M^r>0$, s.t. $|p_r(\bm{x}|Y=y^\prime)-p_r(\bm{x}|Y=y)|\leq g^r(\bm{x})|y^\prime-y|$ with $\int g^r(\bm{x}) d\bm{x} = M^r$;
	
	{\setlength{\parindent}{0cm}\textbf{(A3)}} For $\forall \bm{x}\in\mathcal{X}$ and $y, y^\prime\in\mathcal{Y}$, $\exists g^g(\bm{x}) >0 $ and $ M^g>0$, s.t. $|p_g(\bm{x}|Y=y^\prime)-p_g(\bm{x}|Y=y)|\leq g^g(\bm{x})|y^\prime-y|$ with $\int g^g(\bm{x}) d\bm{x} = M^g$;
	
	{\setlength{\parindent}{0cm}\textbf{(A4)}} $p_r(y)\in\Sigma(L^r)$ and $p_g(y)\in\Sigma(L^g)$.
	
	\subsection{Some Necessary Lemmas and Theorems}\label{supp:CcGAN_lemmas_theorems_bound}
	In this section, we first introduce the Hoeffding's inequality that are widely used later to derive some lemmas.
	\begin{theorem}[Hoeffding's inequality \cite{shalevshwartz2014understanding}]
		\label{supp_thm:CcGAN_hoeffding_inequality}
		Let $Z_1,\dots,Z_m$ be a sequence of i.i.d. random variables and let $\bar{Z}=\frac{1}{m}\sum_{i=1}^mZ_i$. Assume that $\mathbb{E}[\bar{Z}]=\mu$ and $Pr(a\leq Z_i\leq b)=1$ for every i. Then, for any $\epsilon>0$
		\begin{equation*}
			Pr\left[ \left| \frac{1}{m}\sum_{i=1}^mZ_i-\mu \right|>\epsilon \right]\leq 2\exp\left( -\frac{2m\epsilon^2}{(b-a)^2} \right).
		\end{equation*}
	\end{theorem}
	
	\begin{proof}
		Please see \cite[Lemma B.6]{shalevshwartz2014understanding} for the proof.
	\end{proof}
	
	\begin{remark}
		Let $\delta=2\exp\left(-\frac{2m\epsilon^2}{(b-a)^2}\right)$, then $\epsilon=\sqrt{\frac{1}{2m}\log\left( \frac{2}{\delta} \right) } $. Thus, we can get another form of the Hoeffding's inequality. For $\forall \delta \in (0,1)$, with probability at least $1-\delta$, we have
		\begin{equation*}
			\left| \frac{1}{m}\sum_{i=1}^mZ_i-\mu \right|\leq \sqrt{\frac{1}{2m}\log\left( \frac{2}{\delta} \right)}.
		\end{equation*}
	\end{remark}	
	
	\begin{lemma}[Lemma for HVDL]
		\label{supp_lem:CcGAN_conditional_mean_bound_HVDL}
		Suppose that (A1)-(A2) and (A4) hold and let $(\bm{X}_1, Y_1), \dots, (\bm{X}_N, Y_N)$ be a sequence of i.i.d. random image-label pairs, then $\forall \delta\in(0,1)$, with probability at least $1-\delta$, 
		\begin{align}
			&\sup_{D\in\mathcal{D}}\left| \frac{1}{N_{y,\kappa}}\sum_{i=1}^{N}\mathbbm{1}_{\{ |y-Y_i|\leq \kappa \}} \left[ -\log D(\bm{X}_i, y) \right] - \mathbb{E}_{\bm{X}\sim p(\bm{x}|Y=y)} \left[ -\log D(\bm{X}, y) \right] \right| \nonumber\\
			& \leq  U\sqrt{\frac{1}{2 N_{y,\kappa}}\log\left( \frac{2}{\delta} \right)} + \kappa U M,
			\label{supp_eq:CcGAN_conditional_mean_bound_HVDL}
		\end{align}
		for a fixed $y$. If image-label pairs are real, then  $N=N^r$, $N_{y,\kappa}=N^r_{y,\kappa}$, $p=p_r$, and $M=M^r$. Similarly, we have $N=N^g$, $N_{y,\kappa}=N^g_{y,\kappa}$, $p=p_g$, and $M=M^g$ for fake image-label pairs.
	\end{lemma}
	
	\begin{proof}
		Triangle inequality yields 
		\begin{align*}
			&\sup_{D\in\mathcal{D}}\left| \frac{1}{N_{y,\kappa}}\sum_{i=1}^{N}\mathbbm{1}_{\{ |y-Y_i|\leq \kappa \}} \left[ -\log D(\bm{X}_i, y) \right] - \mathbb{E}_{\bm{X}\sim p_r(\bm{x}|Y=y)} \left[ -\log D(\bm{X}, y) \right] \right|\\
			\leq  & \sup_{D\in\mathcal{D}}\left| \frac{1}{N_{y,\kappa}}\sum_{i=1}^{N}\mathbbm{1}_{\{ |y-Y_i|\leq \kappa \}} \left[ -\log D(\bm{X}_i, y) \right] - \mathbb{E}_{\bm{X}\sim p^{y,\kappa}(\bm{x})} \left[ -\log D(\bm{X}, y) \right] \right|\\
			&  + \sup_{D\in\mathcal{D}}\left| \mathbb{E}_{\bm{X}\sim p^{y,\kappa}(\bm{x})} \left[ -\log D(\bm{X}, y) \right] - \mathbb{E}_{\bm{X}\sim p(\bm{x}|Y=y)} \left[ -\log D(\bm{X}, y) \right] \right|
		\end{align*}
		We then bound the two terms of the RHS separately as follows:
		\begin{enumerate}
			\item Real images with labels in $[y-\kappa, y+\kappa]$ can be seen as independent samples from $p^{y,\kappa}(\bm{x})$. Then the first term can be bounded by applying Hoeffding's inequality as follows: $\forall \delta\in (0,1)$, with at least probability $1-\delta$,
			\begin{equation}
				\label{supp_eq:CcGAN_hoeffding_bound_HVDL}
				\begin{aligned}
					&\sup_{D\in\mathcal{D}}\left| \frac{1}{N_{y,\kappa}}\sum_{i=1}^{N}\mathbbm{1}_{\{ |y-Y_i|\leq \kappa \}} \left[ U \frac{-\log D(\bm{X}_i, y)}{U} \right] - \mathbb{E}_{\bm{X}\sim p^{y,\kappa}(\bm{x})} \left[ U \frac{-\log D(\bm{X}, y)}{U} \right] \right|\\
					& \leq   U\sqrt{\frac{1}{2 N_{y,\kappa}}\log\left( \frac{2}{\delta} \right)}.
				\end{aligned}
			\end{equation}
			\item For the second term, we have			
			\begin{align}
				&\sup_{D\in\mathcal{D}}\left| \mathbb{E}_{\bm{X}\sim p^{y,\kappa}(\bm{x})} \left[ -\log D(\bm{X}, y) \right] - \mathbb{E}_{\bm{X}\sim p(\bm{x}|Y=y)} \left[ -\log D(\bm{X}, y) \right] \right| \nonumber\\
				= & \sup_{D\in\mathcal{D}}\left| \int \left[-\log D(\bm{x},y)\right]\cdot\left[ p^{y,\kappa}(\bm{x}) - p(\bm{x}|Y=y) \right] d\bm{x} \right| \nonumber\\
				\leq & \sup_{D\in\mathcal{D}} \int \left|-\log D(\bm{x},y)\right|\cdot\left| p^{y,\kappa}(\bm{x}) - p(\bm{x}|Y=y) \right| d\bm{x} \nonumber\\
				\leq &  U\int \left| p^{y,\kappa}(\bm{x}) - p(\bm{x}|Y=y) \right| d\bm{x}.
				\label{supp_eq:CcGAN_TV_bound_HDVL}
			\end{align}
			Then, we focus on $	\left| p^{y,\kappa}(\bm{x}) - p(\bm{x}|Y=y) \right| $. By the definition of $p^{y,\kappa}(\bm{x})$ and defining $p_\kappa(y^\prime)=\frac{\mathbbm{1}_{\{ |y^\prime-y|\leq\kappa \}}p(y^\prime)}{\int\mathbbm{1}_{\{ |y^\prime-y|\leq\kappa \}}p(y^\prime)dy^\prime}$, we have
			\begin{equation*}
				\begin{aligned}
					& \left| p^{y,\kappa}(\bm{x}) - p(\bm{x}|Y=y) \right| \\
					= & \left| \int p(\bm{x}|Y=y^\prime) p_\kappa(y^\prime)dy^\prime - p(\bm{x}|Y=y) \right|\\
					\leq & \int \left| p(\bm{x}|Y=y^\prime) - p(\bm{x}|Y=y) \right|p_\kappa(y^\prime)dy^\prime\\
					&(\text{by (A2), and let $g=g^r$ for real images and $g=g^g$ for fake images})\\
					\leq & \int g(\bm{x}) |y^\prime-y| p_\kappa(y^\prime) dy^\prime\\
					\leq & \kappa g(\bm{x}).
				\end{aligned}
			\end{equation*}
			Thus, Eq. \eqref{supp_eq:CcGAN_TV_bound_HDVL} is upper bounded as follows,
			\begin{align}
				&\sup_{D\in\mathcal{D}}\left| \mathbb{E}_{\bm{X}\sim p^{y,\kappa}(\bm{x})} \left[ -\log D(\bm{X}, y) \right] - \mathbb{E}_{\bm{X}\sim p(\bm{x}|Y=y)} \left[ -\log D(\bm{X}, y) \right] \right| \nonumber\\
				\leq  & U\int \kappa g(\bm{x}) d\bm{x} \nonumber\\
				& (\text{by (A2)}) \nonumber\\
				= & \kappa UM.
				\label{supp_eq:CcGAN_density_diff_HDVL}
			\end{align}
		\end{enumerate}
		By combining Eq. \eqref{supp_eq:CcGAN_hoeffding_bound_HVDL} and \eqref{supp_eq:CcGAN_density_diff_HDVL}, we can get Eq. \eqref{supp_eq:CcGAN_conditional_mean_bound_HVDL}, which finishes the proof.
	\end{proof}
	
	
	\begin{lemma}[Lemma for SVDL]
		\label{supp_lem:CcGAN_conditional_mean_bound_SVDL}
		Suppose that (A1), (A2) and (A4) hold and let $(\bm{X}_1, Y_1), \dots, (\bm{X}_N, Y_N)$ be a sequence of i.i.d. random image-label pairs, then $\forall\delta\in (0,1)$, with probability at least $1-\delta$,
		\begin{align}
			&\sup_{D\in\mathcal{D}}\left| \frac{\frac{1}{N}\sum_{i=1}^{N}w(Y_i,y)\left[ -\log D(\bm{X}_i, y) \right]}{\frac{1}{N}\sum_{i=1}^{N}w(Y_i,y)} - \mathbb{E}_{\bm{X}\sim p(\bm{x}|Y=y)} \left[ -\log D(\bm{X}, y) \right] \right| \nonumber\\
			& \leq  \frac{2U}{W(y)}\sqrt{\frac{1}{2N} \log\left( \frac{4}{\delta}\right) } + UM \mathbb{E}_{Y^\prime \sim p_w(Y^\prime|Y=y)} \left[|Y^\prime-y| \right] ,
			\label{supp_eq:conditional_mean_bound_SVDL}
		\end{align}
		for a fixed $y$. If image-label pairs are real, then  $N=N^r$, $N_{y,\kappa}=N^r_{y,\kappa}$, $p=p_r$, $p_w=p_w^r$, $w=w^r$, $W=W^r$, and $M=M^r$. Similarly, we have $N=N^g$, $N_{y,\kappa}=N^g_{y,\kappa}$, $p=p_g$, $p_w=p_w^g$, $w=w^g$, $W=W^g$, and $M=M^g$ for fake image-label pairs.
	\end{lemma}
	
	\begin{proof}
		Triangle inequality yields	
		\begin{align}\label{supp_eq:SVDL_tri_cond_mean}
			&\sup_{D\in\mathcal{D}}\left| \frac{\frac{1}{N}\sum_{i=1}^{N}w(Y_i,y)\left[ -\log D(\bm{X}_i, y) \right]}{\frac{1}{N}\sum_{i=1}^{N}w(Y_i,y)} - \mathbb{E}_{\bm{X}\sim p_r(\bm{x}|Y=y)} \left[ -\log D(\bm{X}, y) \right] \right| \nonumber\\
			& \text{(Recall $f(\bm{x}, y) = -\log D(\bm{x}, y) $ and $\mathcal{F} = -\log \mathcal{D}$.)} \nonumber\\
			= & \sup_{f \in \mathcal{F}} \left| \frac{\frac{1}{N}\sum_{i=1}^{N}w(Y_i,y) f(\bm{X}_i, y) }{\frac{1}{N}\sum_{i=1}^{N}w(Y_i,y)} - \mathbb{E}_{\bm{X}\sim p_r(\bm{x}|Y=y)} \left[  f(\bm{X}, y)\right] \right| \nonumber\\
			\leq & \sup_{f \in \mathcal{F}}\left| \frac{\frac{1}{N}\sum_{i=1}^{N}w(Y_i,y)f(\bm{X}_i, y)}{\frac{1}{N}\sum_{i=1}^{N}w(Y_i,y)} - \mathbb{E}_{\bm{X}\sim  p^{y, w}(\bm{x}) } \left[  f(\bm{X}, y)\right]  \right| \nonumber\\
			& + \sup_{f \in \mathcal{F}} \left|\mathbb{E}_{\bm{X}\sim  p^{y, w}(\bm{x}) } \left[  f(\bm{X}, y)\right] - \mathbb{E}_{\bm{X}\sim p_r(\bm{x}|Y=y)} \left[  f(\bm{X}, y)\right]  \right|.  \\
			& \text{($p^{y, w}=p_r^{y, w^r}$ for real images and $p^{y, w}=p_g^{y, w^g}$ for fake images)} \nonumber
		\end{align}
		We then derive bounds for both terms on the RHS of the inequality.
		\begin{enumerate}
			\item For the first term, we can further split it into two parts,	
			\begin{align}\label{eq: 1.a of real cond mean soft}
				& \left| \frac{\frac{1}{N}\sum_{i=1}^{N}w(Y_i,y)f(\bm{X}_i, y)}{\frac{1}{N}\sum_{i=1}^{N}w(Y_i,y)} - \mathbb{E}_{\bm{X}\sim  p^{y, w}(\bm{x}) } \left[  f(\bm{X}, y)\right] \right| \nonumber\\
				\leq  & \left| \frac{\frac{1}{N}\sum_{i=1}^{N}w(Y_i,y)f(\bm{X}_i, y)}{\frac{1}{N}\sum_{i=1}^{N}w(Y_i,y)} - \frac{\frac{1}{N}\sum_{i=1}^{N}w(Y_i,y)f(\bm{X}_i, y)}{W(y)} \right| \nonumber\\
				& + \left| \frac{\frac{1}{N}\sum_{i=1}^{N}w(Y_i,y)f(\bm{X}_i, y)}{W(y)}  - \mathbb{E}_{\bm{X}\sim  p^{y, w}(\bm{x}) } \left[  f(\bm{X}, y)\right]  \right|
			\end{align}
			Focusing on the first part of RHS of Eq.\eqref{eq: 1.a of real cond mean soft}. By (A1), 
			\begin{align}
				& \left| \frac{\frac{1}{N}\sum_{i=1}^{N}w(Y_i,y)f(\bm{X}_i, y)}{\frac{1}{N}\sum_{i=1}^{N}w(Y_i,y)} - \frac{\frac{1}{N}\sum_{i=1}^{N}w(Y_i,y)f(\bm{X}_i, y)}{W(y)} \right| \nonumber\\ 
				\leq & U \frac{\left| \frac{1}{N}\sum_{i=1}^{N}w(Y_i,y) - W(y)\right|}{W(y)} \nonumber
			\end{align}
			Note that $ \forall y, y^\prime,  w(y^\prime,y) = e^{ - \nu |y - y^\prime|^2} \leq 1$ (since $\nu>0$) and hence given $y$, $w(Y^\prime,y)$ is a random variable bounded by $1$. Moreover, given $y$, $W(y)$ is the expectation of $w(Y^\prime,y)$. Then, apply Hoeffding's inequality to the numerator of above, yielding that with probability at least $1 - \delta^\prime$, 
			\begin{equation*}
				\left| \frac{1}{N}\sum_{i=1}^{N}w(Y_i,y) - W(y)\right| 
				\leq  \sqrt{\frac{1}{2N} \log \left(\frac{2}{\delta^\prime} \right) }.
			\end{equation*}
			Thus, by the boundedness of $f$, with probability at least $1 - \delta^\prime$, 
			\begin{align} \label{eq: 1.1.1 cond mean soft}
				&\left| \frac{\frac{1}{N}\sum_{i=1}^{N}w(Y_i,y)f(\bm{X}_i, y)}{\frac{1}{N}\sum_{i=1}^{N}w(Y_i,y)} - \frac{\frac{1}{N}\sum_{i=1}^{N}w(Y_i,y)f(\bm{X}_i, y)}{W(y)} \right|\nonumber\\ 
				\leq & \frac{U}{W(y)}\sqrt{\frac{1}{2N} \log \left(\frac{2}{\delta^\prime} \right) }.
			\end{align}
			
			Then, consider the second part of RHS of Eq.\eqref{eq: 1.a of real cond mean soft}. Recall that  $p^{y, w}(\bm{x})  \triangleq \int p(\bm{x}|Y=y^\prime)\frac{w(y^\prime,y)p(y^\prime)}{W(y)}dy^\prime $. Thus,
			\begin{align*}
				& \left| \frac{\frac{1}{N}\sum_{i=1}^{N}w(Y_i,y)f(\bm{X}_i, y)}{W(y)}  - \mathbb{E}_{\bm{X}\sim  p^{y, w}(\bm{x}) } \left[  f(\bm{X}, y)\right]  \right| \\
				= & \frac{1}{W(y)} \left|\frac{1}{N}\sum_{i=1}^{N}w(Y_i,y)f(\bm{X}_i, y) - \mathbb{E}_{(\bm{X},Y^\prime) \sim p(\bm{x}, y^\prime)}\left[w^r(Y^\prime,y) f(\bm{X}_i, y)  \right]  \right| ,
			\end{align*}
			where $p(\bm{x}, y^\prime) = p(\bm{x}|Y=y^\prime)p(y^\prime)$ denotes PDF of the joint distribution of real image and its label. Again, since $w(Y^\prime,y) f(\bm{X}_i, y) $ is uniformly bounded by $U$ under (A1),  we can apply Hoeffding's inequality. This implies that with probability at least $1- \delta^\prime$, the above can be upper bounded by 
			\begin{equation}
				\label{eq: 1.1.2 cond mean soft}
				\frac{U}{W(y)}\sqrt{\frac{1}{2N} \log\left( \frac{2}{\delta^\prime}\right) }.
			\end{equation}
			Combining Eq. \eqref{eq: 1.1.1 cond mean soft} and \eqref{eq: 1.1.2 cond mean soft} and by setting $\delta' = \frac{\delta}{2}$, we have  with probability at least $1 - \delta$, 
			\begin{equation*}
				\left| \frac{\frac{1}{N}\sum_{i=1}^{N}w(Y_i,y)f(\bm{X}_i, y)}{\frac{1}{N}\sum_{i=1}^{N}w(Y_i,y)} - \mathbb{E}_{\bm{X}\sim  p^{y, w}(\bm{x}) } \left[  f(\bm{X}, y)\right]  \right| \leq  \frac{2U}{W(y)}\sqrt{\frac{1}{2N} \log\left( \frac{4}{\delta}\right) }.
			\end{equation*}
			Since this holds for  $\forall f \in \mathcal{F}$, taking supremum over $f$, we have
			\begin{align}
				\label{eq: 1.1 bound for cond mean soft}
				&\sup_{f\in \mathcal{F}} \left| \frac{\frac{1}{N}\sum_{i=1}^{N}w(Y_i,y)f(\bm{X}_i, y)}{\frac{1}{N}\sum_{i=1}^{N}w(Y_i,y)} - \mathbb{E}_{\bm{X}\sim  p^{y, w}(\bm{x}) } \left[  f(\bm{X}, y)\right] \right| \nonumber\\
				&\leq  \frac{2U}{W(y)}\sqrt{\frac{1}{2N} \log\left( \frac{4}{\delta}\right) }.
			\end{align}
			
			\item For the second term on the RHS of Eq.\ \eqref{supp_eq:SVDL_tri_cond_mean}. By (A1) that $|f|\leq U$,
			\begin{equation*}
				\begin{aligned}
					& \sup_{f \in \mathcal{F}} \left|\mathbb{E}_{\bm{X}\sim  p^{y, w}(\bm{x}) } \left[  f(\bm{X}, y)\right] - \mathbb{E}_{\bm{X}\sim p(\bm{x}|Y=y)}  \left[  f(\bm{X}, y)\right] \right| \\
					& \text{(See Eq.\ \eqref{supp_eq:CcGAN_TV_bound_HDVL})} \\
					\leq & U \int |p^{y, w}(\bm{x}) - p(\bm{x}|Y=y)| d\bm{x}.
				\end{aligned}
			\end{equation*}
			Note that by the definition of $$p^{y,w}(\bm{x})\triangleq \int p(\bm{x}|Y=y^\prime)\frac{w(y^\prime,y)p(y^\prime)}{W(y)}dy^\prime $$ and $$p_{w}\left(y^{\prime} | Y=y\right) \allowbreak\triangleq \frac{w\left(y^{\prime}, y\right) p^{r}\left(y^{\prime}\right)}{W^{r}(y)},$$ we have
			\begin{equation*}
				\begin{aligned}
					|p^{y, w}(\bm{x}) - p(\bm{x}|Y=y)| & =  \left|\int p(\bm{x}|Y=y^\prime) p_{w}\left(y^{\prime} | Y=y\right) dy^\prime - p(\bm{x}|Y=y)   \right|\\
					& \leq \int \left| p(\bm{x}|Y=y^\prime) - p(\bm{x}|Y=y)\right| p_{w}\left(y^{\prime} | Y=y\right) dy^\prime.
				\end{aligned}
			\end{equation*}
			By (A.2) and $y \in [0,1]$, the above is upper bounded by 
			$$g(\bm{x}) \mathbb{E}_{Y^\prime \sim p_{w}\left(y^{\prime} | Y=y\right)} \left[| y - Y^\prime| \right].$$ 
			Thus,
			\begin{align}\label{eq: 2 bound cond mean soft}
				& \sup_{f \in \mathcal{F}} \left|\mathbb{E}_{\bm{X}\sim  p^{y, w}(\bm{x}) } \left[  f(\bm{X}, y)\right] - \mathbb{E}_{\bm{X}\sim p(\bm{x}|Y=y)}  \left[  f(\bm{X}, y)\right] \right| \nonumber\\
				\leq & U \int g(\bm{x}) \mathbb{E}_{Y^\prime \sim p_{w}\left(y^{\prime} | Y=y\right)} \left[| Y^\prime - y| \right] d\bm{x} \nonumber\\
				= & UM\mathbb{E}_{Y^\prime \sim p_{w}\left(y^{\prime} | Y=y\right)} \left[| Y^\prime - y| \right].
			\end{align}
		\end{enumerate}
		
		Therefore, combining both Eq.\eqref{eq: 1.1 bound for cond mean soft} and \eqref{eq: 2 bound cond mean soft}, with probability at least $1- \delta$,
		\[
		\begin{aligned}
			& \sup_{D\in\mathcal{D}}\left| \frac{\frac{1}{N}\sum_{i=1}^{N}w(Y_i,y)\left[ -\log D(\bm{X}_i, y) \right]}{\frac{1}{N}\sum_{i=1}^{N}w(Y_i,y)} - \mathbb{E}_{\bm{X}\sim p(\bm{x}|Y=y)} \left[ -\log D(\bm{X}, y) \right] \right| \\
			\leq & \frac{2U}{W(y)}\sqrt{\frac{1}{2N} \log\left( \frac{4}{\delta}\right) } +  UM\mathbb{E}_{Y^\prime \sim p_{w}\left(y^{\prime} | Y=y\right)} \left[| Y^\prime - y| \right] .
		\end{aligned}
		\]
		This finishes the proof.
	\end{proof}
	
	As introduced in Section \ref{sec:cGAN2CcGAN}, we use KDE for the density of the marginal label distribution with Gaussian kernel. The next theorem  characterizes the difference between a $p_r(y), p_g(y)$ and their KDE using $N$ i.i.d. samples. 
	
	\begin{theorem}
		\label{supp_thm:CcGAN_bound_kde}
		Let $\hat{p}_r^{\text{KDE}}(y)$ and $\hat{p}_g^{\text{KDE}}(y)$ stand for the KDE of $p_r(y)$ and $p_g(y)$ respectively. Under condition (A4), if the KDEs are based on $N$ i.i.d. samples from $p_r/p_g$ and a bandwidth $\sigma$, for all $\delta\in (0,1)$, with probability at least $1-\delta$,
		\begin{align}
			\sup_t\left| \hat{p}_r^{\text{KDE}}(y)-p_r(y) \right|\leq \sqrt{\frac{C_{1,\delta}^{\text{KDE}}\log N}{N\sigma}} + L^r\sigma,\\
			\sup_t\left| \hat{p}_g^{\text{KDE}}(y)-p_g(y) \right|\leq \sqrt{\frac{C_{2,\delta}^{\text{KDE}}\log N}{N\sigma}} + L^g\sigma,
		\end{align}
		for some constants $C_{1,\delta}^{\text{KDE}}, C_{2,\delta}^{\text{KDE}}$ depending on $\delta$.
	\end{theorem}
	
	\begin{proof}
		By (\cite{larry_KDE}; P.12), for any $p(t) \in \Sigma(L)$ (the H\"older Class, see Definition \ref{supp_def:holder_class}), with probability at least $1 - \delta$, 
		\begin{equation*}
			\sup_t\left| \hat{p}^{\text{KDE}}(t)-p(t) \right|\leq \sqrt{\frac{C_{\delta}^{\text{KDE}}\log N}{N\sigma}} + c\sigma,
		\end{equation*}
		for some constants $C_{\delta}^{\text{KDE}}$ and $c$, where $C$ depends on $\delta$ and $c = L\int K(s) |s|^2 ds$. Since in this work, $K$ is chosen as Gaussian kernel, $c = L\int K(s) |s|^2 ds = L$. 
	\end{proof}
	
	Based on above lemmas and theorems, we derive the following two theorems, which will be used in the derivation of the error bounds of $D$ trained with HVDL and SVDL in Section \ref{supp:CcGAN_main_theorems_bound}.
	
	\begin{theorem}
		\label{supp_thm:CcGAN_error_bound_HVDL} 
		Assume that (A1)-(A4) hold, then $\forall \delta\in(0,1)$, with probability at least $1-\delta$,
		\begin{align}\label{supp_eq:CcGAN_error_bound_HVDL}
			&\sup_{D\in\mathcal{D}}\left|\widehat{\mathcal{L}}^{\text{HVDL}}(D)-\mathcal{L}(D)\right| \nonumber\\
			& \leq  U\left(\sqrt{\frac{C_{1,\delta}^{\text{KDE}}\log N^r}{N^r\sigma}} +  L^r\sigma^2 \right)+ U\left(\sqrt{\frac{C_{2,\delta}^{\text{KDE}}\log N^g}{N^g\sigma}} + L^g\sigma^2 \right) +  \kappa U (M^r+M^g) \nonumber\\
			& \quad +  U\sqrt{\frac{1}{2} \log \left( \frac{8}{\delta}\right)} \left(  \mathbb{E}_{Y\sim \hat{p}_r^{\text{KDE}}(y)} \left[ \sqrt{\frac{1}{N_{Y, \kappa}^r}}\right] + \mathbb{E}_{Y\sim \hat{p}_g^{\text{KDE}}(y)} \left[ \sqrt{\frac{1}{N_{Y, \kappa}^g}}\right] \right),
		\end{align}
		for some constants $C_{1,\delta}^{\text{KDE}}, C_{2,\delta}^{\text{KDE}}$ depending on $\delta$. 
	\end{theorem}
	
	\begin{proof}
		
		Let $(\bm{X}^r_1, Y^r_1), \dots, (\bm{X}^r_{N^r}, Y^r_{N^r})$ and $(\bm{X}^g_1,\allowbreak Y^g_1), \dots, (\bm{X}^g_{N^g}, Y^g_{N^g})$ denote respectively real and fake i.i.d. random image-label pairs. 
		
		We first decompose $\sup_{D\in\mathcal{D}}\left|\widehat{\mathcal{L}}^{\text{HVDL}}(D)-\mathcal{L}(D)\right|$ as follows
		\begin{equation*}
			\begin{aligned}
				&\sup_{D\in\mathcal{D}}\left|\widehat{\mathcal{L}}^{\text{HVDL}}(D)-\mathcal{L}(D)\right|\\
				\leq & \sup_{D\in\mathcal{D}}\left| \int \left[\int \left[-\log D(\bm{x},y )\right] p_r(\bm{x}|Y=y)d\bm{x}\right](p_r(y)-\hat{p}_r^{\text{KDE}}(y))dy \right| \\
				+ & \sup_{D\in\mathcal{D}}\left| \int \left[\int \left[-\log(1-D(\bm{x},y))\right] p_g(\bm{x}|Y=y)d\bm{x}\right](p_g(y)-\hat{p}_g^{\text{KDE}}(y))dy \right| \\
				+ & \sup_{D\in\mathcal{D}}\left|\int \left[ \frac{1}{N^r_{y,\kappa}}\sum_{i=1}^{N^r}\mathbbm{1}_{\{ |y-Y_i^r|\leq \kappa \}} \left[ -\log D(\bm{X}_i^r, y) \right] - \mathbb{E}_{\bm{X}\sim p_r(\bm{x}|Y=y)} \left[ -\log D(\bm{X}, y) \right] \right] \hat{p}_r^{\text{KDE}}(y)dy \right|\\
				+ & \sup_{D\in\mathcal{D}}\left|\int \left[ \frac{1}{N^g_{y,\kappa}}\sum_{i=1}^{N^r}\mathbbm{1}_{\{ |y-Y_i^g|\leq \kappa \}} \left[ -\log (1-D(\bm{X}_i^g, y)) \right] - \mathbb{E}_{\bm{X}\sim p_g(\bm{x}|Y=y)} \left[ -\log (1-D(\bm{X}, y)) \right] \right] \hat{p}_g^{\text{KDE}}(y)dy \right|.
			\end{aligned}
		\end{equation*}
		These four terms in the RHS can be bounded separately as follows
		\begin{enumerate}
			\item The first term can be bounded by using Theorem \ref{supp_thm:CcGAN_bound_kde} and the boundness of $D$. For the first term, $\forall \delta_1\in(0,1)$, with at least probability $1-\delta_1$, 
			\begin{equation}
				\label{supp_eq:CcGAN_real_integral_kde_bound}
				\begin{aligned}
					& \sup_{D\in\mathcal{D}}\left| \int \left[\int \left[-\log D(\bm{x},y )\right] p_r(\bm{x}|Y=y)d\bm{x}\right](p_r(y)-\hat{p}_r^{\text{KDE}}(y))dy \right| \\
					\leq & U\left(\sqrt{\frac{C_{1,\delta_1}^{\text{KDE}}\log N^r}{N^r\sigma}} + L^r\sigma^2 \right),
				\end{aligned}
			\end{equation}
			for some constants $C_{1,\delta_1}^{\text{KDE}}$ depending on $\delta_1$. 
			\item Similarly, for the second term, $\forall \delta_2\in(0,1)$, with at least probability $1-\delta_2$, 
			\begin{equation}
				\label{supp_eq:CcGAN_fake_integral_kde_bound}
				\begin{aligned}
					&\sup_{D\in\mathcal{D}}\left| \int \left[\int \left[-\log(1-D(\bm{x},y))\right] p_g(\bm{x}|Y=y)d\bm{x}\right](p_g(y)-\hat{p}_g^{\text{KDE}}(y))dy \right|\\
					\leq  & U\left(\sqrt{\frac{C_{2,\delta_2}^{\text{KDE}}\log N^g}{N^r\sigma}} + L^g\sigma^2 \right),
				\end{aligned}
			\end{equation}
			for some constants $C_{2,\delta_2}^{\text{KDE}}$ depending on $\delta_2$. 
			
			\item The third term can be bounded by using Lemma \ref{supp_lem:CcGAN_conditional_mean_bound_HVDL}. For the third term, $\forall \delta_3\in (0,1)$, with at least probability $1-\delta_3$,
			\begin{equation*}
				\begin{aligned}
					&\sup_{D\in\mathcal{D}}\left|\int \left[ \frac{1}{N^r_{y,\kappa}}\sum_{i=1}^{N^r}\mathbbm{1}_{\{ |y-Y_i^r|\leq \kappa \}} \left[ -\log D(\bm{X}_i^r, y) \right] - \mathbb{E}_{\bm{X}\sim p_r(\bm{x}|Y=y)} \left[ -\log D(\bm{X}, y) \right] \right] \hat{p}_r^{\text{KDE}}(y)dy \right| \\
					\leq & \int\sup_{D\in\mathcal{D}}\left| \frac{1}{N^r_{y,\kappa}}\sum_{i=1}^{N^r}\mathbbm{1}_{\{ |y-Y_i^r|\leq \kappa \}} \left[ -\log D(\bm{X}_i^r, y) \right] - \mathbb{E}_{\bm{X}\sim p_r(\bm{x}|Y=y)} \left[ -\log D(\bm{X}, y) \right] \right|  \hat{p}_r^{\text{KDE}}(y)dy\\
					\leq & \int  \left[ U\sqrt{\frac{1}{2 N^r_{y,\kappa}}\log\left( \frac{2}{\delta_3} \right)} + \kappa U M^r \right] \hat{p}_r^{\text{KDE}}(y)dy
				\end{aligned}
			\end{equation*}
			Note that $ N^r_{y,\kappa} = \sum_{i = 1}^{N^r}\mathbbm{1}_{\{ |y-Y_i^r|\}} $, which is a random variable of $Y_i$'s. The above can be expressed as 
			\begin{equation}
				\label{supp_eq:CcGAN_real_integral_redemacher_bound_HVDL}
				\begin{aligned}
					&\sup_{D\in\mathcal{D}}\left|\int \left[ \frac{1}{N^r_{y,\kappa}}\sum_{i=1}^{N^r}\mathbbm{1}_{\{ |y-Y_i^r|\leq \kappa \}} \left[ -\log D(\bm{X}_i^r, y) \right] - \mathbb{E}_{\bm{X}\sim p_r(\bm{x}|Y=y)} \left[ -\log D(\bm{X}, y) \right] \right] \hat{p}_r^{\text{KDE}}(y)dy \right| \\
					& \leq \kappa U M^r + U\sqrt{\frac{1}{2} \log \left( \frac{2}{\delta_3}\right)} \mathbb{E}_{Y\sim \hat{p}_r^{\text{KDE}}(y)} \left[ \sqrt{\frac{1}{N_{Y, \kappa}^r}}\right].
				\end{aligned}
			\end{equation}
			
			\item Similarly, for the fourth term, $\forall \delta_4 \in (0,1)$, with at least probability $1-\delta_4$, 
			\begin{equation}
				\label{supp_eq:CcGAN_fake_integral_redemacher_bound_HVDL}
				\begin{aligned}
					&\sup_{D\in\mathcal{D}}\left|\int\left\{\int\left[ \frac{1}{N^g_{y,\kappa}}\sum_{i=1}^{N^r}\mathbbm{1}_{\{ |y-Y_i^g|\leq \kappa \}} \left[ -\log (1-D(\bm{X}_i^g, y)) \right] \right.\right.\right.\\
					&\left.\left.\left.\vphantom{\sum_{i=1}^{N^r}} - \mathbb{E}_{\bm{X}\sim p_g(\bm{x}|Y=y)} \left[ -\log (1-D(\bm{X}, y)) \right] \right] d\bm{x} \right\} \hat{p}_g^{\text{KDE}}(y)dy \right|\\
					\leq &  \kappa U M^g + U\sqrt{\frac{1}{2} \log \left( \frac{2}{\delta_4}\right)} \mathbb{E}_{Y\sim \hat{p}_g^{\text{KDE}}(y)} \left[ \sqrt{\frac{1}{N_{Y, \kappa}^g}}\right].
				\end{aligned}
			\end{equation}
		\end{enumerate}
		With $\delta_1=\delta_2=\delta_3=\delta_4=\frac{\delta}{4}$, combining Eq. \eqref{supp_eq:CcGAN_real_integral_kde_bound} - \eqref{supp_eq:CcGAN_fake_integral_redemacher_bound_HVDL} leads to the upper bound in Theorem \ref{supp_thm:CcGAN_error_bound_HVDL}.
	\end{proof}
	
	\begin{theorem}
		\label{supp_thm:CcGAN_error_bound_SVDL}
		Assume that (A1)-(A4) hold, then $\forall \delta\in(0,1)$, with probability at least $1-\delta$,
		\begin{equation}
			\label{supp_eq:CcGAN_error_bound_SVDL}
			\begin{aligned}
				&\sup_{D\in\mathcal{D}}\left|\widehat{\mathcal{L}}^{\text{SVDL}}(D)-\mathcal{L}(D)\right|  \\
				& \leq   U\left(\sqrt{\frac{C_{1,\delta}^{\text{KDE}}\log N^r}{N^r\sigma}} + L^r\sigma^2 \right) + U\left(\sqrt{\frac{C_{2,\delta}^{\text{KDE}}\log N^g}{N^g\sigma}} + L^g\sigma^2 \right)  \\
				& \quad + 4U\sqrt{\frac{1}{2} \log \left( \frac{16}{\delta}\right)} \left(  \frac{1}{\sqrt{N^r}}\mathbb{E}_{Y\sim \hat{p}_r^{\text{KDE}}(y)}\left[  \frac{1}{W^r(Y)}\right]+ \frac{1}{\sqrt{N^g}}\mathbb{E}_{Y\sim \hat{p}_g^{\text{KDE}}(y)} \left[  \frac{1}{W^g(Y)}\right] \right)  \\
				& \quad + 2U  \left( M^r\mathbb{E}_{Y\sim \hat{p}_r^{\text{KDE}}(y)}\left[ \mathbb{E}_{Y^\prime\sim p^r_w(y^\prime|Y)}\left|Y^\prime-Y\right|  \right] +   M^g\mathbb{E}_{Y\sim \hat{p}_g^{\text{KDE}}(y)}\left[  \mathbb{E}_{Y^\prime\sim p^g_w(y^\prime|Y)}\left|Y^\prime-Y\right| \right] \right)
			\end{aligned}
		\end{equation}
		for some constant $C_{1,\delta}^{\text{KDE}},  C_{2,\delta}^{\text{KDE}}$ depending on $\delta$. 
	\end{theorem}
	
	\begin{proof}
		Similar to the decomposition for Theorem \ref{supp_thm:CcGAN_error_bound_HVDL}, we can decompose $\sup_{D\in\mathcal{D}}\allowbreak|\widehat{\mathcal{L}}^{\text{SVDL}}(D)-\mathcal{L}(D)|$ into four terms which can be bounded by using Theorem \ref{supp_thm:CcGAN_bound_kde}, the boundness of $D$, and Lemma \ref{supp_lem:CcGAN_conditional_mean_bound_SVDL}. The detail is omitted because it is almost identical to the one of Theorem \ref{supp_thm:CcGAN_error_bound_HVDL}.
	\end{proof}

	\subsection{Error Bounds of $D$ Trained with HVDL and SVDL}\label{supp:CcGAN_main_theorems_bound}
	
\begin{figure}[!htbp]
		\centering
		\includegraphics[width=0.8\textwidth]{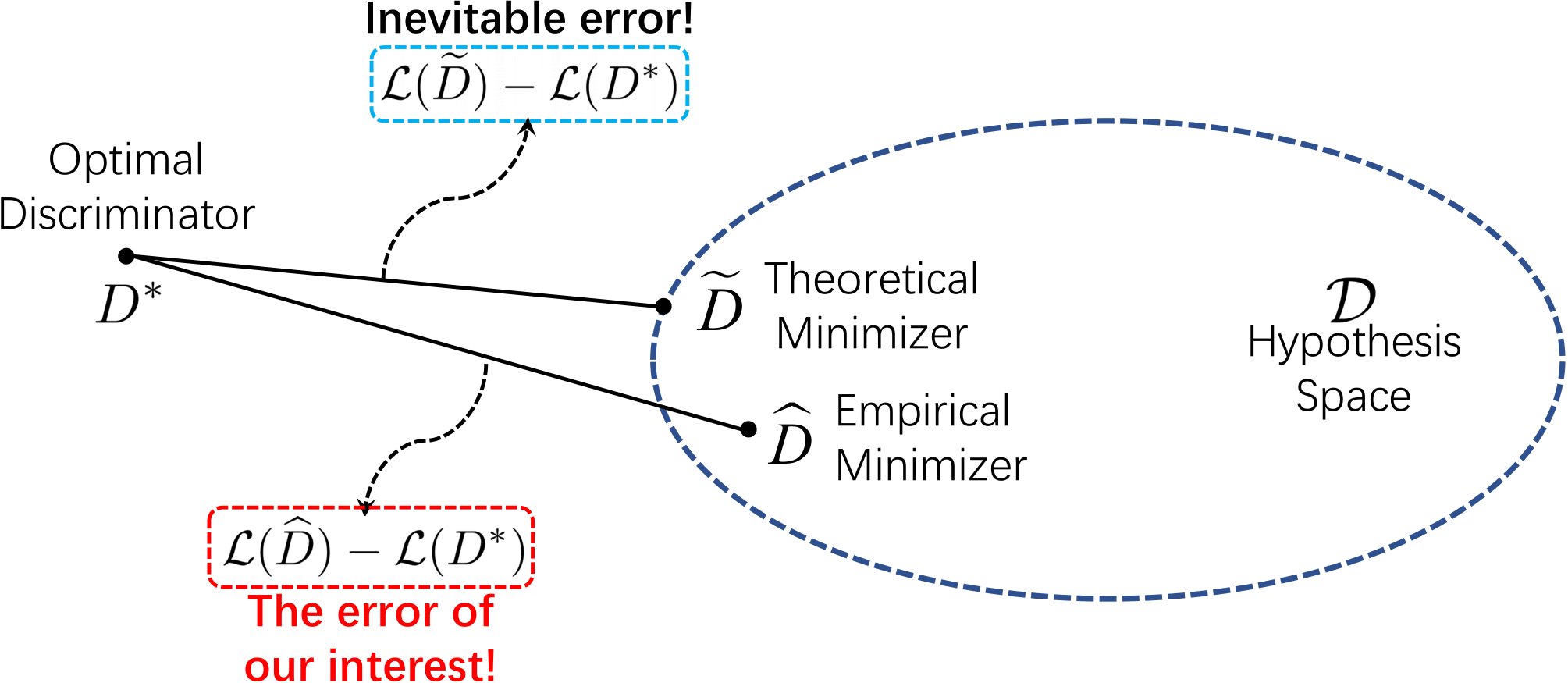}
		\caption{An illustrative diagram for error bounds of HVDL and SVDL.}
		\label{fig:diagram_error_bound}
	\end{figure}

	Based on above theorems and lemmas, we derive the error bounds of $D$ that is trained with HVDL and SVDL respectively. The error bound is characterized by the distance of $\widehat{D}^{\text{HVDL}}$ and $\widehat{D}^{\text{SVDL}}$ from the optimal $D^*$ under the theoretical discriminator loss $\mathcal{L}(D)$, i.e., $\mathcal{L}(\widehat{D}^{\text{HVDL}})-\mathcal{L}(D^*)$ and $\mathcal{L}(\widehat{D}^{\text{SVDL}})-\mathcal{L}(D^*)$ respectively. Please see Theorem \ref{supp_thm:CcGAN_convergence_rate_HVDL} and \ref{supp_thm:CcGAN_convergence_rate_SVDL} for details. An illustrative diagram to visualize the theoretical analysis is shown in Fig.\ \ref{fig:diagram_error_bound}.
	
	\begin{theorem}[Error bound of D trained with HVDL]
		\label{supp_thm:CcGAN_convergence_rate_HVDL} 
		Assume that (A1)-(A4) hold, then $\forall \delta\in(0,1)$, with probability at least $1-\delta$,
		\begin{align}
			&\mathcal{L}(\widehat{D}^{\text{HVDL}})-\mathcal{L}(D^*) \\
			\leq &   2U\left(\sqrt{\frac{C_{1,\delta}^{\text{KDE}}\log N^r}{N^r\sigma}} +  L^r\sigma^2 \right) + 2U\left(\sqrt{\frac{C_{2,\delta}^{\text{KDE}}\log N^g}{N^g\sigma}} + L^g\sigma^2 \right) \nonumber\\
			& +  2\kappa U (M^r+M^g) \nonumber\\
			& +  2U\sqrt{\frac{1}{2} \log \left( \frac{8}{\delta}\right)} \left(  \mathbb{E}_{Y\sim \hat{p}_r^{\text{KDE}}(y)} \left[ \sqrt{\frac{1}{N_{Y, \kappa}^r}} \, \right]   + \mathbb{E}_{Y\sim \hat{p}_g^{\text{KDE}}(y)} \left[ \sqrt{\frac{1}{N_{Y, \kappa}^g}}\, \right] \right) \nonumber\\
			&+ \mathcal{L}(\widetilde{D})-\mathcal{L}(D^*), \label{supp_eq:CcGAN_convergence_rate_HVDL}
		\end{align}
		for some constants $C_{1,\delta}^{\text{KDE}}, C_{2,\delta}^{\text{KDE}}$ depending on $\delta$. 
	\end{theorem}
	
	\begin{proof}
		We first decompose $\mathcal{L}(\widehat{D}^{\text{HVDL}})-\mathcal{L}(D^*)$ as follows
		\begin{equation}
			\begin{aligned}
				& \mathcal{L}(\widehat{D}^{\text{HVDL}})-\mathcal{L}(D^*)\\
				= & \mathcal{L}(\widehat{D}^{\text{HVDL}})-\widehat{\mathcal{L}}(\widehat{D}^{\text{HVDL}})+\widehat{\mathcal{L}}(\widehat{D}^{\text{HVDL}})-  \widehat{\mathcal{L}}(\widetilde{D}) +  \widehat{\mathcal{L}}(\widetilde{D})- \mathcal{L}(\widetilde{D})\\
				&+\mathcal{L}(\widetilde{D})-\mathcal{L}(D^*) \\
				& (\text{by } \widehat{\mathcal{L}}(\widehat{D}^{\text{HVDL}})-  \widehat{\mathcal{L}}(\widetilde{D}) \leq 0)\\
				\leq & 2\sup_{D\in\mathcal{D}}\left|\widehat{\mathcal{L}}^{\text{HVDL}}(D)-\mathcal{L}(D)\right|+\mathcal{L}(\widetilde{D})-\mathcal{L}(D^*)\\
				&\text{(by Theorem \ref{supp_thm:CcGAN_error_bound_HVDL})}\\
				\leq &  2U\left(\sqrt{\frac{C_{1,\delta}^{\text{KDE}}\log N^r}{N^r\sigma}} +  L^r\sigma^2 \right)+ 2U\left(\sqrt{\frac{C_{2,\delta}^{\text{KDE}}\log N^g}{N^g\sigma}} + L^g\sigma^2 \right) \\
				& +  2\kappa U (M^r+M^g) \\
				\hphantom{\leq} & +  2U\sqrt{\frac{1}{2} \log \left( \frac{8}{\delta}\right)} \left(  \mathbb{E}_{Y\sim \hat{p}_r^{\text{KDE}}(y)} \left[ \sqrt{\frac{1}{N_{Y, \kappa}^r}}\right] + \mathbb{E}_{Y\sim \hat{p}_g^{\text{KDE}}(y)} \left[ \sqrt{\frac{1}{N_{Y, \kappa}^g}}\right] \right) \\
				& + \mathcal{L}(\widetilde{D})-\mathcal{L}(D^*).
			\end{aligned}
		\end{equation}
	\end{proof}

	\begin{theorem}[Error bound of D trained with SVDL]
		\label{supp_thm:CcGAN_convergence_rate_SVDL}
		Assume that (A1)-(A4) hold, then $\forall \delta\in(0,1)$, with probability at least $1-\delta$,
		\begin{align}
			&\mathcal{L}(\widehat{D}^{\text{SVDL}})-\mathcal{L}(D^*) \nonumber\\
			\leq & 2U\left(\sqrt{\frac{C_{1,\delta}^{\text{KDE}}\log N^r}{N^r\sigma}} + L^r\sigma^2 \right) + 2U\left(\sqrt{\frac{C_{2,\delta}^{\text{KDE}}\log N^g}{N^g\sigma}} + L^g\sigma^2 \right) \nonumber\\
			& + 4U\sqrt{\frac{1}{2}\log \left( \frac{16}{\delta}\right)} \left(  \frac{1}{\sqrt{N^r}}\mathbb{E}_{Y\sim \hat{p}_r^{\text{KDE}}(y)}\left[  \frac{1}{W^r(Y)}\right] + \frac{1}{\sqrt{N^g}}\mathbb{E}_{Y\sim \hat{p}_g^{\text{KDE}}(y)} \left[  \frac{1}{W^g(Y)}\right] \right) \nonumber\\
			& + 2U \left( M^r\mathbb{E}_{Y\sim \hat{p}_r^{\text{KDE}}(y)}\left[ \mathbb{E}_{Y^\prime \sim \hat{p}^r_{w}(y^\prime|Y)}\left|Y^\prime-Y\right|  \right] + M^g\mathbb{E}_{Y\sim \hat{p}_g^{\text{KDE}}(y)}\left[  \mathbb{E}_{Y^\prime \sim \hat{p}^g_{w}(y^\prime|Y)}\left|Y^\prime-Y\right| \right] \right) \nonumber\\
			& + \mathcal{L}(\widetilde{D})-\mathcal{L}(D^*), 
			\label{supp_eq:CcGAN_convergence_rate_SVDL}
		\end{align}
		for some constant $C_{1,\delta}^{\text{KDE}}, \; C_{2,\delta}^{\text{KDE}}$ depending on $\delta$. 
	\end{theorem}
	
	\begin{proof}
		Smilarly, based on Theorem \ref{supp_thm:CcGAN_error_bound_SVDL}, we can derive Theorem \ref{supp_thm:CcGAN_convergence_rate_SVDL}. The detailed proof is omitted.
	\end{proof}

	\section{More details of the experiment on Low-resolution RC-49 in Section \ref{sec:rc49}}\label{supp:details_of_rc49}
	
	\subsection{Description of RC-49}\label{supp:rc49_data}
	
	To generate RC-49, firstly we randomly select 49 3-D chair object models from the ``Chair'' category provided by ShapeNet \cite{chang2015shapenet}. Then we use Blender v2.79 \footnote{https://www.blender.org/download/releases/2-79/} to render these 3-D models. Specifically, during the rendering, we rotate each chair model along with the yaw axis for a degree between $0.1^{\circ} $ and $89.9^{\circ} $ (angle resolution as $0.1^{\circ} $) where we use the scene image mode to compose our dataset. The rendered images are converted from the RGBA to RGB color model. In total, the RC-49 dataset consists of 44051 images of image size 64$\times$64 in the PNG format.

	\subsection{Network architectures}\label{supp:rc49_nets}
	The RC-49 dataset is a more sophisticated dataset compared with the simulation, thus it requires networks with deeper layers. We employ the SNGAN architecture \cite{miyato2018spectral} in both cGAN and CcGAN consisting of residual blocks for the generator and the discriminator. Moreover, for the generator in cGAN, the regression labels are input into the network by the label embedding \cite{akata2015label} and the conditional batch normalization \cite{de2017modulating}. For the discriminator in cGAN, the regression labels are fed into the network by the label embedding and the label projection \cite{miyato2018cgans}. For CcGAN, the regression labels are fed into networks by the two proposed label input methods (NLI and ILI) in Section \ref{sec:solution_2}. The pre-trained CNN $T_1+T_2$ for ILI is a modified ResNet-34 with two extra linear layers before the final linear layer. The label embedding network $T_3$ is a 5-layer MLP with 128 nodes in each layer. The dimension of the noise $z$ is 128 for NLI-based CcGANs and 256 for ILI-based CcGANs. Please refer to our codes for more details about the network specifications of cGAN and CcGAN.

	\subsection{Training setups}\label{supp:rc49_training_setups}
	
	The cGAN and CcGAN are trained for 30,000 iterations on the training set with the Adam \cite{KingmaB14} optimizer (with $\beta_1=0.5$ and $\beta_2=0.999$), a constant learning rate $10^{-4}$ and batch size 256. 
	
	The rule of thumb formulae in Rmk~\ref{rmk:rule_of_thumb} are used to select the hyper-parameters for HVDL and SVDL, where we let $m_{\kappa}=2$. Thus, the three hyper-parameters in this experiments are set as follows: $\sigma=0.0473$, $\kappa=0.004$, $\nu=50625$. 
	
	The modified ResNet-34 (i.e., the $T_1+T_2$ in Fig.\ \ref{fig:pre_trained_CNN_for_label_embedding}) for ILI is trained for 200 epochs with the SGD optimizer, initial learning rate 0.1 (decay at epoch 60, 120, and 160 with factor 0.2), weight decay $10^{-4}$, and batch size 256. The 5-layer MLP for the label embedding in ILI is trained for 500 epochs with the SGD optimizer, initial learning rate  0.1 (decay at epoch 100, 200, and 400 with factor 0.2), weight decay $10^{-4}$, and batch size 256.
	
	Please see our codes for more details of the training setups.

	\subsection{Testing setups}\label{supp:rc49_testing_setups}
	The RC-49 dataset consists of 899 distinct yaw angles and at each angle there are 49 images (corresponding to 49 types of chairs). At the test stage, we ask the trained cGAN or CcGAN to generate 200 fake images at each of these 899 yaw angles. Please note that, among these 899 yaw angles, only 450 of them are seen at the training stage so real images at the rest 449 angles are not used in the training.
	
	We evaluate the quality of the fake images from three perspectives, i.e., visual quality, intra-label diversity, and label consistency. One overall metric (Intra-FID) and three separate metrics (NIQE, Diversity, and Label Score) are used. Their details are shown in Supp. \ref{supp:rc49_performance}.

	\subsection{Performance measures}\label{supp:rc49_performance}
	Before we conduct the evaluation in terms of the four metrics, we first train an autoencoder (AE) , a regression-oriented ResNet-34 \cite{he2016deep} and a classification-oriented ResNet-34 \cite{he2016deep} on all real images of RC-49. The bottleneck dimension of the AE is 512 and the AE is trained to reconstruct the real images in RC-49 with the MSE loss. The regression-oriented ResNet-34 is trained to predict the yaw angle of a given image. The classification-oriented ResNet-34 is trained to predict the chair type of a given image. The autoencoder and both two ResNets are trained for 200 epochs with a batch size of 256.
	
	\begin{itemize}
		\item \textbf{Intra-FID} \cite{miyato2018cgans}: \emph{We take Intra-FID as the overall score to evaluate the quality of fake images and we prefer the small Intra-FID score.} At each evaluation angle, we compute the FID \cite{heusel2017gans} between 49 real images and 200 fake images in terms of the bottleneck feature of the pre-trained AE. The Intra-FID score is the average FID over all 899 evaluation angles. Please note that we also try to use the classification-oriented ResNet-34 to compute the Intra-FID but the Intra-FID scores vary in a very wide range and sometimes obviously contradict with the three separate metrics. 
		
		\item \textbf{NIQE} \cite{mittal2012making}: \emph{NIQE is used to evaluate the visual quality of fake images with the real images as the reference and we prefer the small NIQE score.} We train one NIQE model with the 49 real images at each of the 899 angles so we have 899 NIQE models. During evaluation, an average NIQE score is computed for each evaluation angle based on the NIQE model at that angle. Finally, we report the average and standard deviations of the 899 average NIQE scores over the 899 yaw angels (i.e., ``the mean/standard deviation of 899 means"). Note that the NIQE is implemented by the NIQE library in \texttt{MATLAB}. The block size and the sharpness threshold are set to 8 and 0.1 respectively in this and rest experiments.
		
		\item \textbf{Diversity}: \emph{Diversity is used to evaluate the intra-label diversity and the larger the better.} In RC-49, there are 49 chair types. At each evaluation angle, we ask a pre-trained classification -oriented ResNet-34 to predict the chair types of the 200 fake images and an entropy is computed based on these predicted chair types. The diversity reported in Table\ \ref{tab:low_resolution_results} is the average of the 899 entropies over all evaluation angles. 
		
		\item \textbf{Label Score}: \emph{Label Score is used to evaluate the label consistency and the smaller the better.} We ask the pre-trained regression-oriented ResNet-34 to predict the yaw angles of all fake images and the predicted angles are then compared with the assigned angles. The Label Score is defined as the average absolute distance between the predicted angles and assigned angles over all fake images, which is equivalent to the Mean Absolute Error (MAE). Note that, to plot the line graphs, we compute Label Score at each of the 899 evaluation angles.
	\end{itemize}
	
	\subsection{Example $64\times 64$ RC-49 images}\label{supp:rc49_more_example_images}
\begin{figure}[!htbp]
		\centering
		\includegraphics[width=0.7\linewidth]{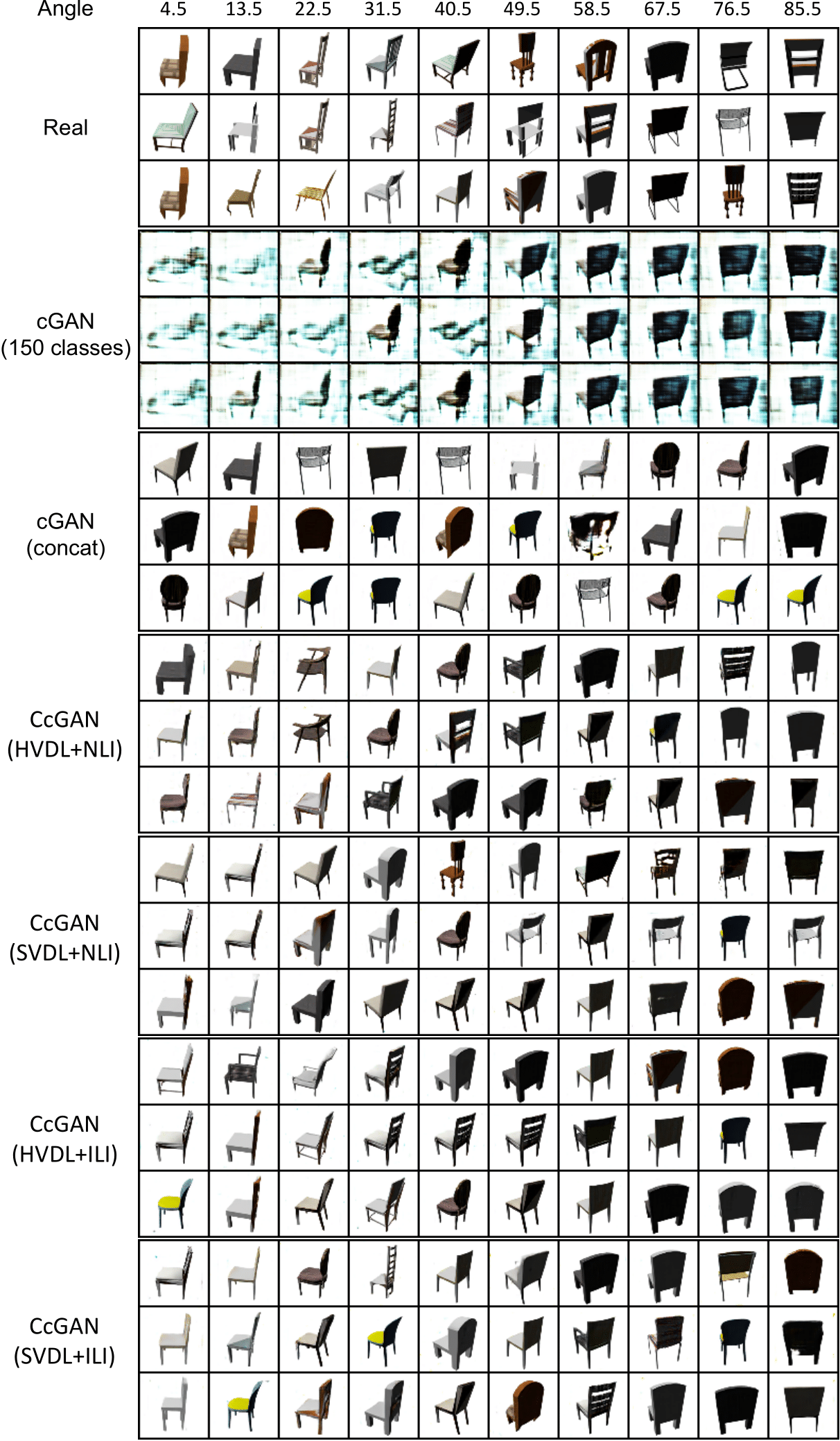}
		\caption{Three RC-49 example images in $64\times 64$ resolution for each of 10 angles: real images and example fake images from cGAN and four proposed CcGANs, respectively. CcGANs produce chair images with \textbf{higher visual quality and more diversity}.}
		\label{fig:rc49_visual_results}
	\end{figure}

	Example RC-49 images are shown in Fig.\ \ref{fig:rc49_visual_results}.
	
	\subsection{Extra experiments}\label{supp:rc49_extra_experiment}	
	
	\subsubsection{Interpolation}\label{supp:rc49_CcGAN_interpolation}
	
\begin{figure}[!htbp]
		\centering
		\includegraphics[width=0.7\textwidth]{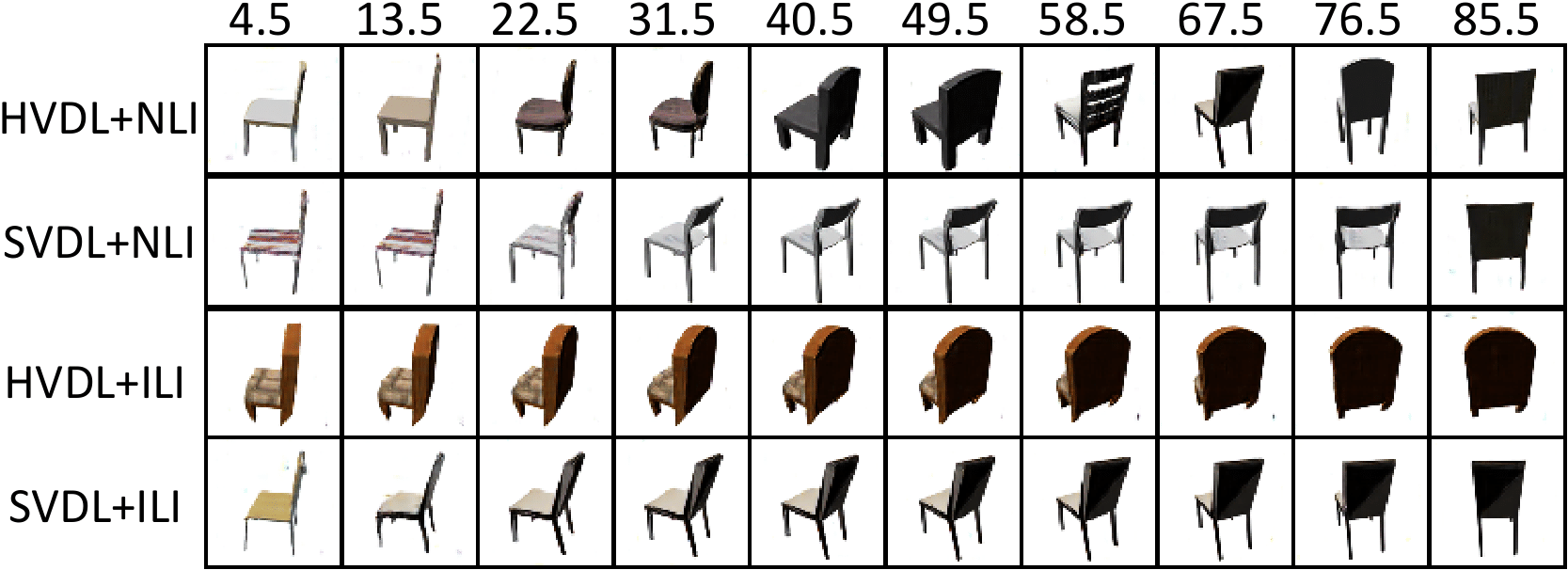}
		\caption{Some example RC-49 fake images from the four CcGAN methods. We fix the noise $\bm{z}$ but vary the label $y$.}
		\label{fig:rc49_fix_z_continuous_label}
	\end{figure}

	In Fig. \ref{fig:rc49_fix_z_continuous_label}, we present some interpolation results of the four CcGAN methods (i.e., HVDL+NLI, SVDL+NLI, HVDL+ILI, and SVDL+ILI). For an input pair $(z, y)$, we fix the noise $z$ but perform label-wise interpolations, i.e., varying label $y$ from 4.5 to 85.5. Clearly, all generated images are visually realistic and we can see the chair distribution smoothly changes over continuous angles. Please note that, Fig. \ref{fig:rc49_fix_z_continuous_label} is meant to show the smooth change of the chair distribution instead of one single chair so the chair type may change over angles. This confirms CcGAN is capable of capturing the underlying conditional image distribution rather than simply memorizing training data.

	\subsubsection{Degenerated CcGAN}\label{supp:rc49_CcGAN_degeneration}
	
	In this experiment, we consider the extreme case of the proposed CcGAN (degenerated CcGAN), i.e., $\sigma\rightarrow 0$ and $\kappa\rightarrow 0$ or $\nu\rightarrow +\infty$. Some examples from a degenerated NLI-based CcGAN are shown in Fig. \ref{fig:rc49_CcGAN_degeneration}. Since, at each angle, the degenerated CcGAN only uses the images at this angle, it leads to the mode collapse problem (e.g, the row in the yellow rectangle) and bad visual quality (e.g., images in the red rectangle) at some angles. 
	
	Note that the degenerated CcGAN is still different from cGAN, since we still treat $y$ as a continuous scalar instead of a class label here and we use the proposed label input method (e.g., NLI) to incorporate $y$ into the generator and the discriminator.

	\subsubsection{cGAN: different number of classes}\label{supp:rc49_cGAN_bin_comparison}
	
\begin{figure}[!htbp]
		\centering
		\begin{minipage}{.5\textwidth}
			\centering
			\includegraphics[width=0.4\linewidth]{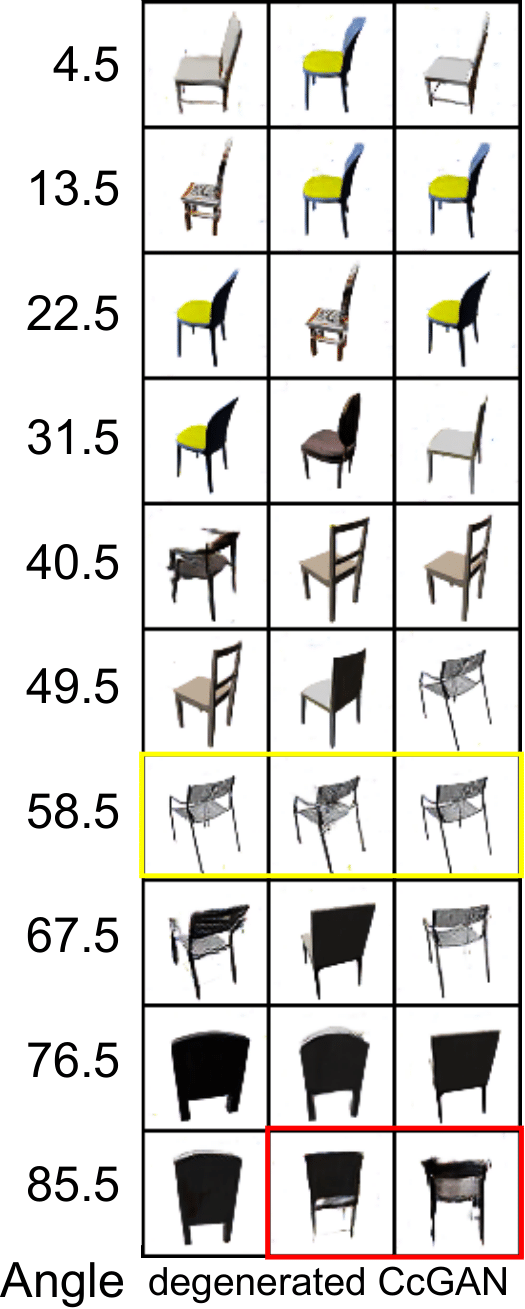}
			\captionsetup{width=0.95\linewidth}
			\captionof{figure}{Some example RC-49 fake images from a degenerated NLI-based CcGAN.}
			\label{fig:rc49_CcGAN_degeneration}
		\end{minipage}%
		\begin{minipage}{.5\textwidth}
			\centering
			\includegraphics[width=0.95\linewidth]{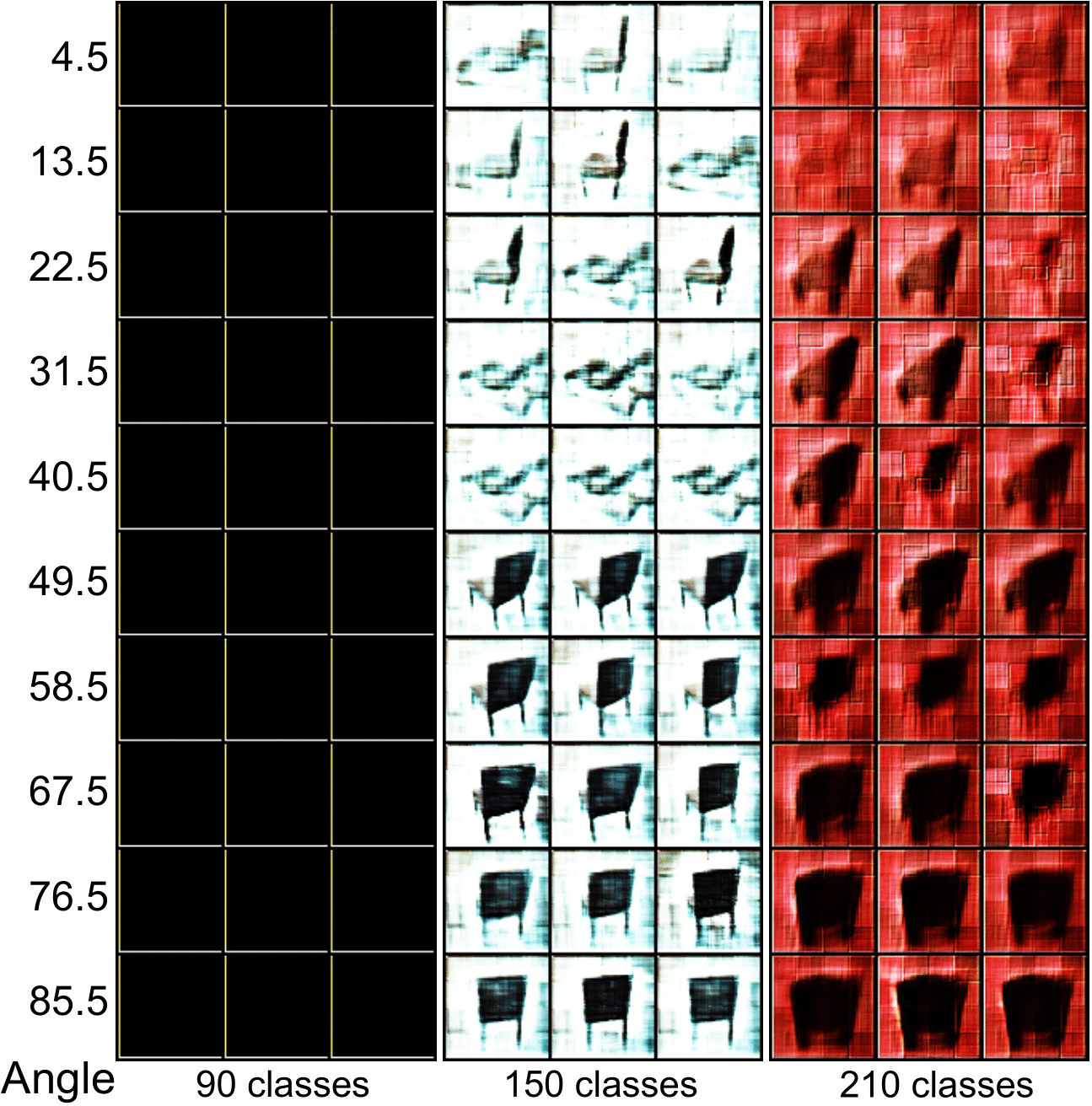}
			\captionsetup{width=0.95\linewidth}
			\captionof{figure}{Example RC-49 fake images from cGAN when we bin the yaw angle range into different number of classes.}
			\label{fig:rc49_cGAN_bin_comparison}
		\end{minipage}
	\end{figure}

	In this experiment, we show that cGAN still fails even though we bin $[0.1,89.9]$ into other number of classes. We experimented with three different bin setting -- grouping labels into 90, 150, and 210 classes, respectively. Experimental results are shown in Fig. \ref{fig:rc49_cGAN_bin_comparison} and we observe all three cGANs completely fail.
	
	\section{More details of the experiment on the Low-resolution UTKFace dataset in Section \ref{sec:utkface}}\label{supp:details_of_utkface}
	
	\subsection{Description of the UTKFace dataset}\label{supp:utkface_data}
	
\begin{figure}[!htbp]
		\centering
		\includegraphics[width=0.5\textwidth]{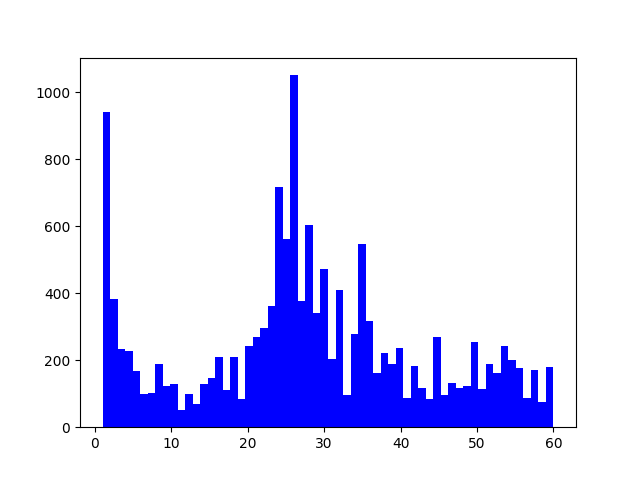}
		\caption{The histogram of the UTKFace dataset with ages varying from 1 to 60.}
		\label{fig:utkface_histogram}
	\end{figure}

	The UTKFace dataset is an age regression dataset \cite{utkface}, with human face images collected in the wild. We use a preprocessed version (cropped and aligned), with ages spanning from 1 to 60. After the data cleaning (i.e., removing images of very low quality or with clearly wrong labels), the number of images left is 14760. These images are then resized to $64\times64$. The histogram of the UTKFace dataset after data cleaning is shown in \ref{fig:utkface_histogram}. 
	
	From Fig.\ \ref{fig:utkface_histogram}, we can see UTKFace dataset is very imbalanced so the samples from the minority age groups are unlikely to be chosen at each iteration during the GAN training. Consequently, cGAN and CcGAN may not be well-trained at these minority age groups. To increase the chance of drawing these minority samples during training, we randomly replicate samples in the minority age groups to ensure that the sample size of each age is more than 200. 
	
	\subsection{Network architectures}\label{supp:utkface_nets}
	The network architectures used in this experiment is similar to those in the RC-49 experiment. Please refer to our codes for more details about the network specifications. 
	
	\subsection{Training setups}\label{supp:utkface_training_setups}
	The cGAN and CcGAN are trained for 40,000 iterations on the training set with the Adam \cite{KingmaB14} optimizer (with $\beta_1=0.5$ and $\beta_2=0.999$), a constant learning rate $10^{-4}$ and batch size 512. The rule of thumb formulae in Section \ref{rmk:rule_of_thumb} are used to select the hyper-parameters for HVDL and SVDL, where we let $m_{\kappa}=1$. 
	
	Please see our codes for more details of the training setups.

	\subsection{Performance measures}\label{supp:utkface_performance}
	
	Similar to the RC-49 experiment, we evaluate the quality of fake images by Intra-FID, NIQE, Diversity, and Label Score. We also train an AE (bottleneck dimension is 512), a classification-oriented ResNet-34, and a regression-oriented ResNet-34 on the UTKFace dataset. Please note that, the UTKFace dataset consists of face images from 5 races based on which we train the classification-oriented ResNet-34. The AE and both two ResNets are trained for 200 epochs with a batch size 256.	
	
	\subsection{Example UTKFace images}\label{supp:utkface_more_example_images}
\begin{figure}[!htbp]
		\centering
		\includegraphics[width=0.7\linewidth]{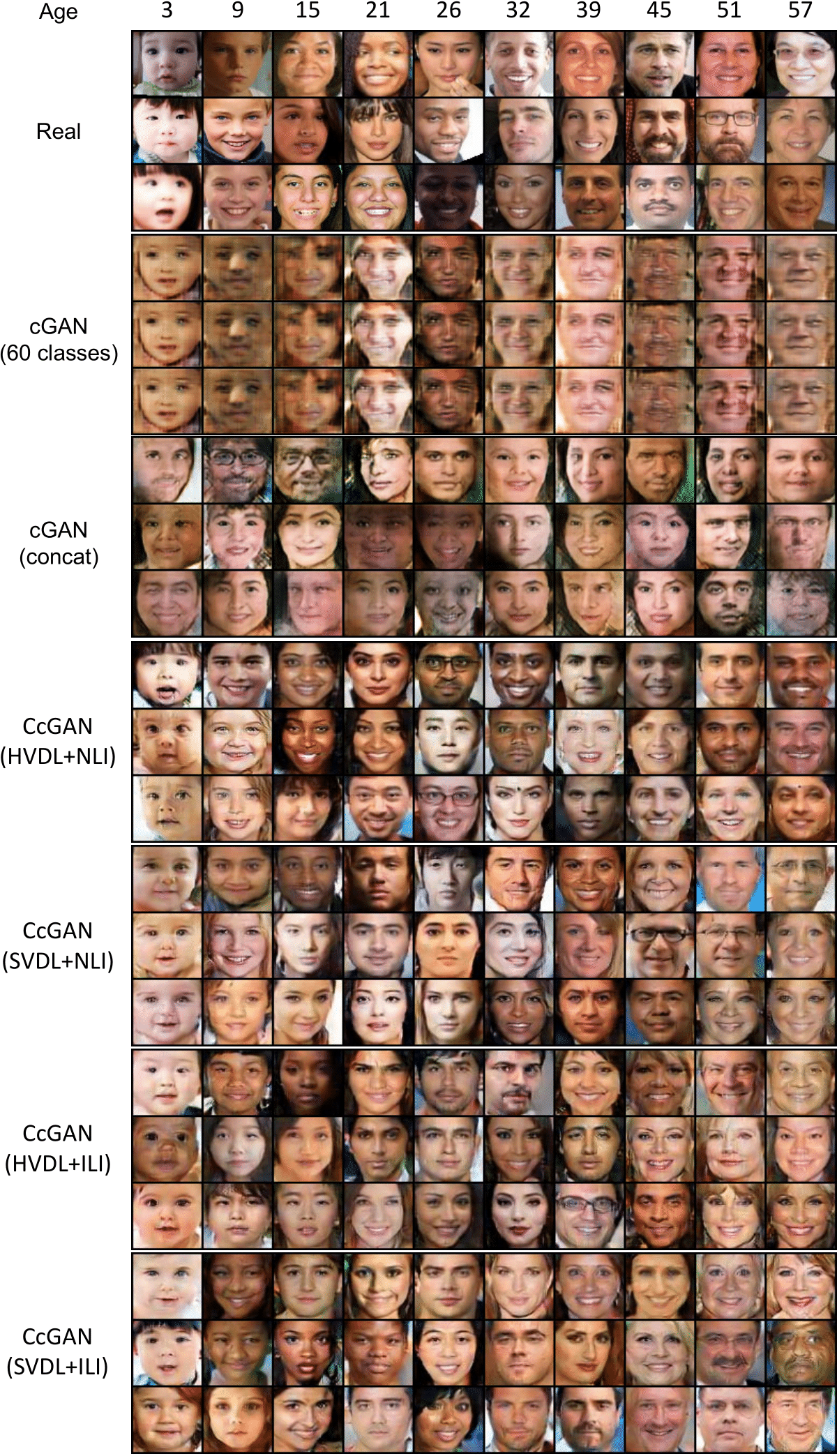}
		\caption{Three UTKFace example images in $64\times 64$ resolution for each of 10 ages: real images and example fake images from cGAN and four proposed CcGANs, respectively. CcGANs produce face images with \textbf{higher visual quality and more diversity}.}
		\label{fig:UTKFace_visual_results}
	\end{figure}

	Example UTKFace images are shown in Fig.\ \ref{fig:UTKFace_visual_results}.
	
	\subsection{Extra experiments}\label{supp:utkface_extra_experiment}

	\subsubsection{Interpolation}\label{supp:utkface_CcGAN_interpolation}
\begin{figure}[!htbp]
		\centering
		\includegraphics[width=0.7\textwidth]{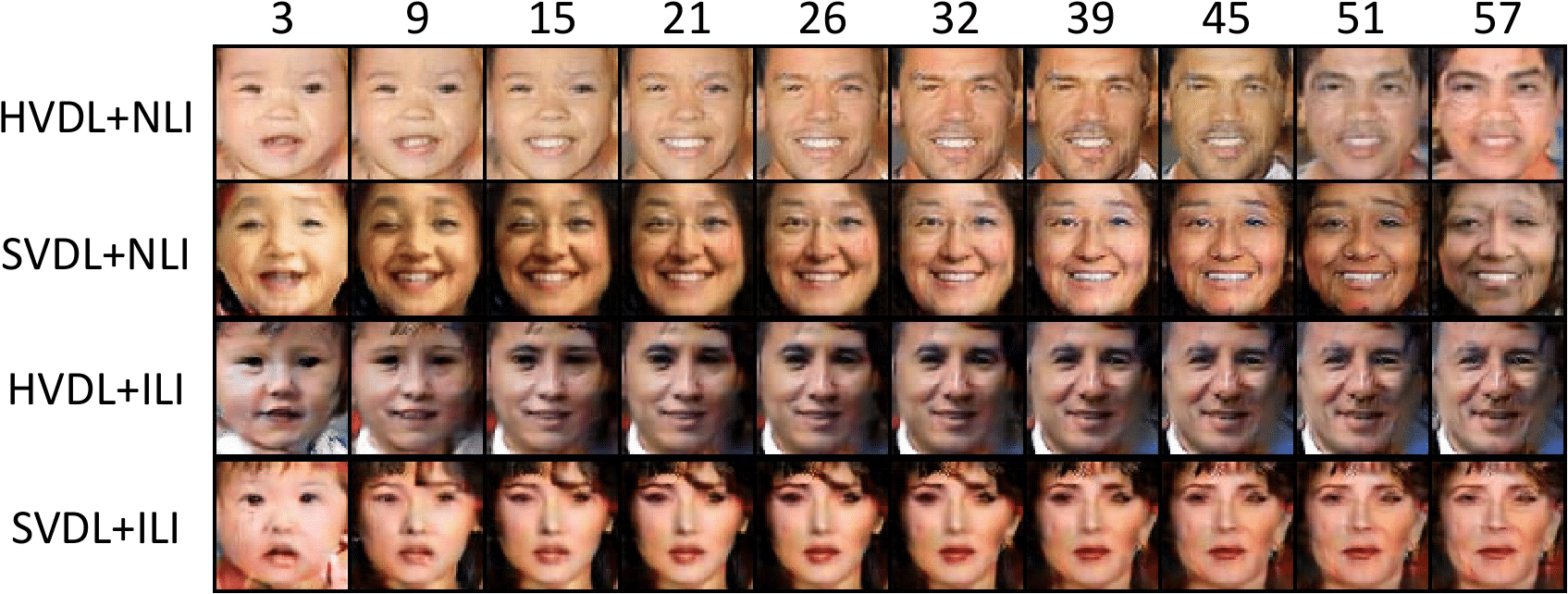}
		\caption{Some examples of generated UTKFace images from the four CcGAN methods. We fix the noise $\bm{z}$ but vary the label $y$ from 3 to 57.}
		\label{fig:UTKFace_fix_z_continuous_label}
	\end{figure}

	To perform label interpolation experiments, we keep the noise vector $z$ fixed and vary label from age 3 to age 57 for the four CcGANs. The interpolation results are illustrated in \ref{fig:UTKFace_fix_z_continuous_label}. As age $y$ increases, we observe the synthetic face gradually becomes older in appearance. This observation convincingly shows that all four CcGANs do not simply memorize or overfit to the training set. Indeed, our CcGANs demonstrate continuous control over synthetic images with respect to ages.     

	\subsubsection{Degenerated CcGAN}\label{supp:utkface_CcGAN_degeneration}
	We consider the extreme cases of the proposed CcGANs on the UTKFace dataset. As shown in Fig. \ref{fig:utkface_CcGAN_degeneration}, the degenerated NLI-based CcGANs fails to generate facial images at some ages (e.g., 51 and 57) because of too small sample sizes.

	\subsubsection{cGAN: different number of classes}\label{supp:utkface_cGAN_bin_comparison}
\begin{figure}[!htbp]
		\centering
		\begin{minipage}{.4\textwidth}
			\centering
			\includegraphics[width=0.55\linewidth]{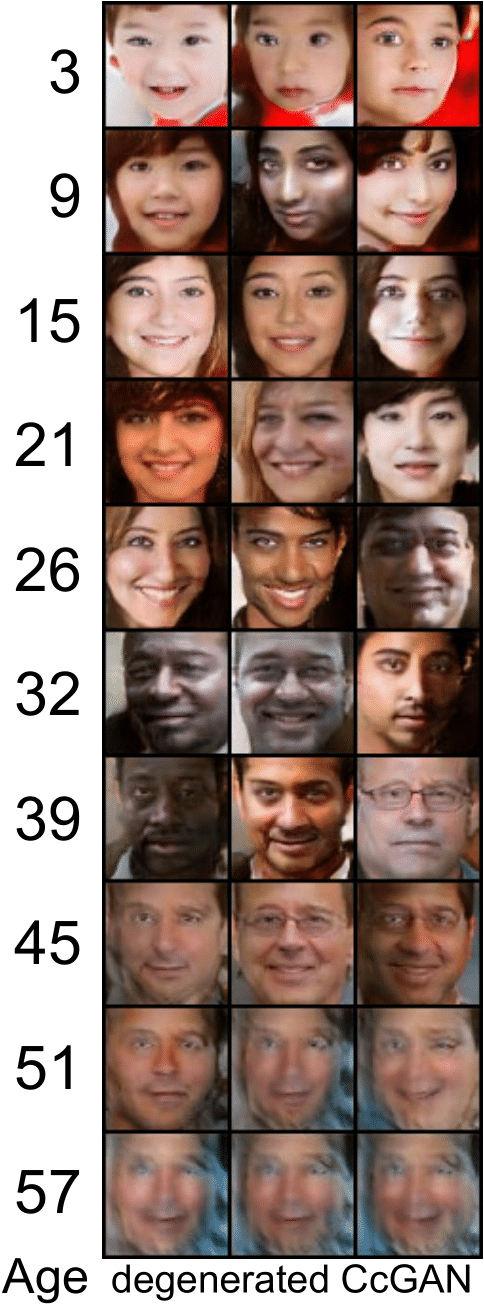}
			\captionsetup{width=0.95\linewidth}
			\captionof{figure}{Some example UTKFace fake images from a degenerated NLI-based CcGAN.}
			\label{fig:utkface_CcGAN_degeneration}
		\end{minipage}%
		\begin{minipage}{.4\textwidth}
			\centering
			\includegraphics[width=0.95\linewidth]{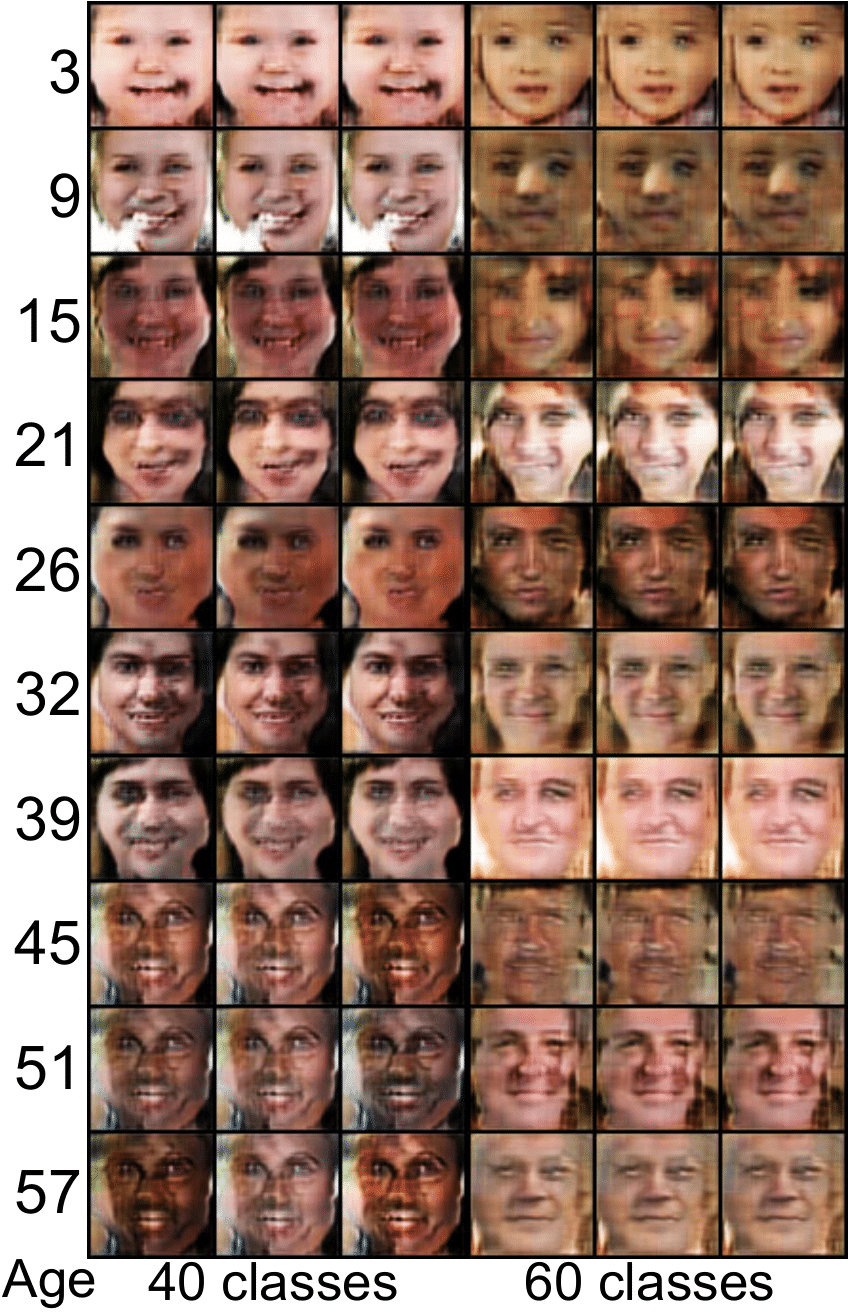}
			\captionsetup{width=0.95\linewidth}
			\captionof{figure}{Example UTKFace fake images from cGAN when we bin the age range into different number of classes.}
			\label{fig:utkface_cGAN_bin_comparison}
		\end{minipage}
	\end{figure}

	In the last experiment, we bin samples into different number of classes based on ground-truth labels, in order to increase the number of training samples at each class. Then we train cGAN using samples from the binned classes. We experimented with two different bin setting, i.e., binning image samples into 60 classes and 40 classes, respectively. The results are reported in Fig.\ref{fig:utkface_cGAN_bin_comparison}. The results demonstrate cGANs consistently fail to generate diverse synthetic images with labels aligned with their conditional information. Moreover, the image quality is much worse than those from the proposed CcGANs. In conclusion, compared with existing cGANs, our proposed CcGANs have substantially better performance in terms of the image quality and diversity.

	\section{More details of the experiment on the Low-resolution Cell-200 dataset in Section \ref{sec:cell200}}\label{supp:details_of_cell200}
	
	\subsection{Description of Cell-200}\label{supp:cell200_data}
	
	Cell-200 is a synthetic image dataset, emulating the colonies of bacterial cells in the view of fluorescence microscope. This dataset contains cell populations with overall number varying between 1 and 200, generated with \cite{simcep}. For each cell population (e.g., 1 to 200), we generate 1000 different synthetic fluorescence microscopic images, with diverse cell variations (e.g., shapes, locations, overlaps and blurring effects). As in \cite{count10}, we set nucleus radius as 5, and image size as $256\times256$. To alleviate computational burden, images in the Cell-200 dataset are then resized to $64\times64$.   
	
	\subsection{Network architectures}\label{supp:cell200_nets}
	The network architectures for cGAN and CcGAN in this experiment are adapted from the famous DCGAN \cite{radford2015unsupervised} architecture. The dimension of the noise $z$ is 128 for NLI-based CcGANs and 256 for ILI-based CcGANs. Please refer to our codes for more details about the network specifications.

	\subsection{Training setups}\label{supp:cell200_training_setups}
	The cGAN and CcGAN are trained for 5000 iterations on the training set with the Adam \cite{KingmaB14} optimizer (with $\beta_1=0.5$ and $\beta_2=0.999$) and a constant learning rate $10^{-4}$. The rule of thumb formulae in Remark \ref{rmk:rule_of_thumb} are used to select the hyperparameters for HVDL and SVDL, where we let $m_{\kappa}=2$. 
	
	For cGAN training, the cell count range $[1,200]$ is split into 100 disjoint intervals, i.e., $[1,3), [3,5), \dots, [197,199), [199,200]$. In this case, cGAN estimates image distribution conditional on these intervals. In Supp. \ref{supp:cell200_cGAN_bin_comparison}, we also compare the performance of cGAN under different splitting schemes.
	
	Please note that we use different batch sizes for cGAN and CcGAN in this experiment. The batch size for cGAN is 512. Differently, for all CcGAN methods in this experiment, the batch size is 512 for the generator but 32 for the discriminator. The reason that we use different batch sizes for the generator and discriminator in CcGAN is based on some observations we got during training. In this experiment, if the generator and the discriminator in CcGAN have the same batch size, the discriminator loss often decreases to almost zero very quickly while the generator loss still maintains at a high level. Consequently, at each iteration, the discriminator can easily distinguish the real and fake images while the generator cannot fool the discriminator and won't improve in the next iteration, which implies a high imbalance between the generator update and the discriminator update. To balance the training of the generator and the discriminator, we deliberately decrease the number of images seen by the discriminator at each iteration to slow down the update of the discriminator so that the generator can catch up. 
	
	Please see our codes for more details of the training setups.
	
	\subsection{Testing setups}\label{supp:cell200_testing_setups}
	We evaluate the trained cGAN and four CcGAN methods on all 200 cell counts (half of them are blinded during training). Each method generates 1,000 images for each cell count, so there are 200,000 fake images from each method. 	
	
	When evaluating the trained cGAN, if a test label $y^\prime$ is unseen in the training set, we just need to find which interval (recall we split $[1,200]$ is split into 100 disjoint intervals) covers this label. Then, we generate samples from the trained cGAN conditional on this interval instead of $y^\prime$.

	\subsection{Performance measures}\label{supp:cell200_performance}
	Similar to the previous two experiments, we evaluate the quality of fake images by Intra-FID, NIQE, and Label Score but Diversity. The Diversity score is not available here because we don't have any class label in Cell-200. An AE with a bottleneck dimension of 512 and a regression-oriented ResNet-34 are pre-trained on the complete Cell-200 dataset (i.e., 200,000 images) to compute the Intra-FID and Label Score respectively. The possibility of lacking class labels in regression-oriented datasets is another reason that we propose to use an AE to compute Intra-FID instead of a classification-oriented CNN. The AE is trained for 50 epochs with a batch size of 256. The regression-oriented ResNet-34 is trained for 200 epochs with a batch size of 256.

	\subsection{Example UTKFace images}\label{supp:cell200_more_example_images}
\begin{figure}[!htbp]
		\centering
		\includegraphics[width=0.7\linewidth]{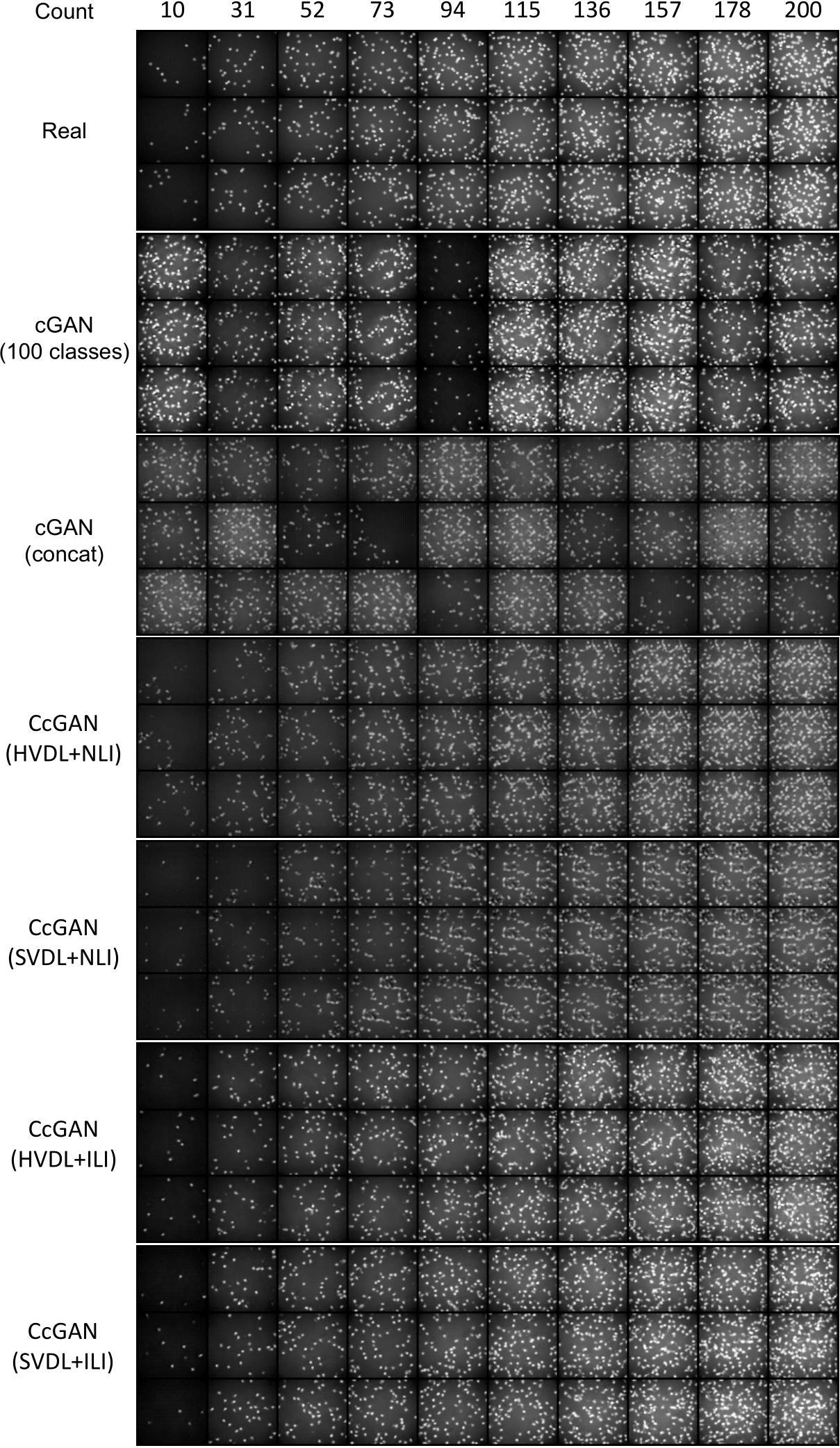} 
		\caption{Three Cell-200 images in $64\times 64$ resolution for each of 10 cell counts absent in the training data: real images and example fake images from cGAN and four proposed CcGANs, respectively. cGAN has severe mode collapse problem in this experiment. Two NLI-based CcGANs do not perform well enough but two ILI-based CcGANs produce images with \textbf{higher visual quality, more diversity, and higher label consistency}.}
		\label{fig:cell200_visual_results}
	\end{figure}

	Example Cell-200 images are shown in Fig.\ \ref{fig:cell200_visual_results}.

	\subsection{Extra experiments}\label{supp:cell200_extra_experiment}
	
	\subsubsection{Interpolation}\label{supp:cell200_CcGAN_interpolation}
\begin{figure}[!htbp]
		\centering
		\includegraphics[width=0.6\textwidth]{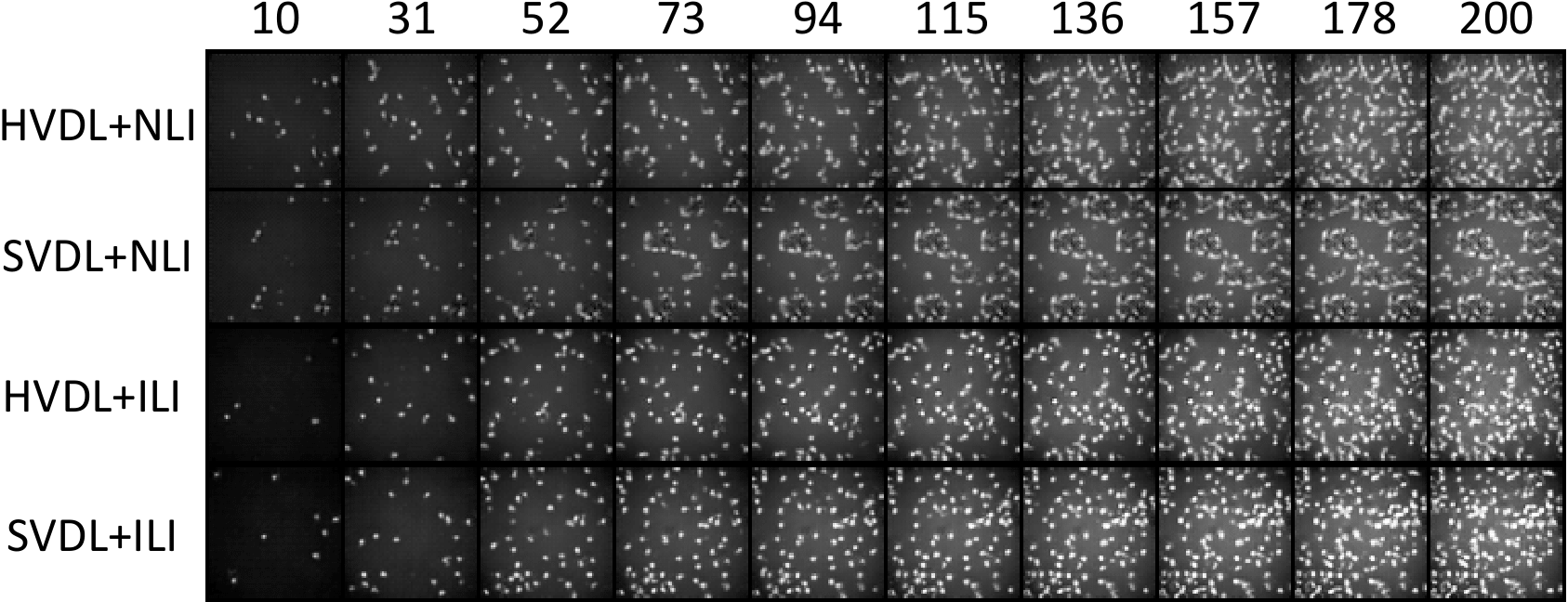}
		\caption{Some examples of generated Cell-200 images from the four CcGAN methods. We fix the noise $\bm{z}$ but vary the label $y$ from 10 to 200.}
		\label{fig:cell200_fix_z_continuous_label}
	\end{figure}

	To perform the label interpolation, we keep the noise vector $z$ fixed and vary label from 10 to 200 for the four CcGANs. The interpolation results are illustrated in \ref{fig:cell200_fix_z_continuous_label}. As cell count $y$ increases, we observe the cells in images become more crowded. This observation convincingly shows that all four CcGANs do not simply memorize or overfit to the training set. Indeed, our CcGANs demonstrate continuous control over synthetic images with respect to cell counts.     

	\subsubsection{cGAN: different number of classes}\label{supp:cell200_cGAN_bin_comparison}
	
\begin{figure}[!htbp]
		\centering
		\includegraphics[width=0.45\textwidth]{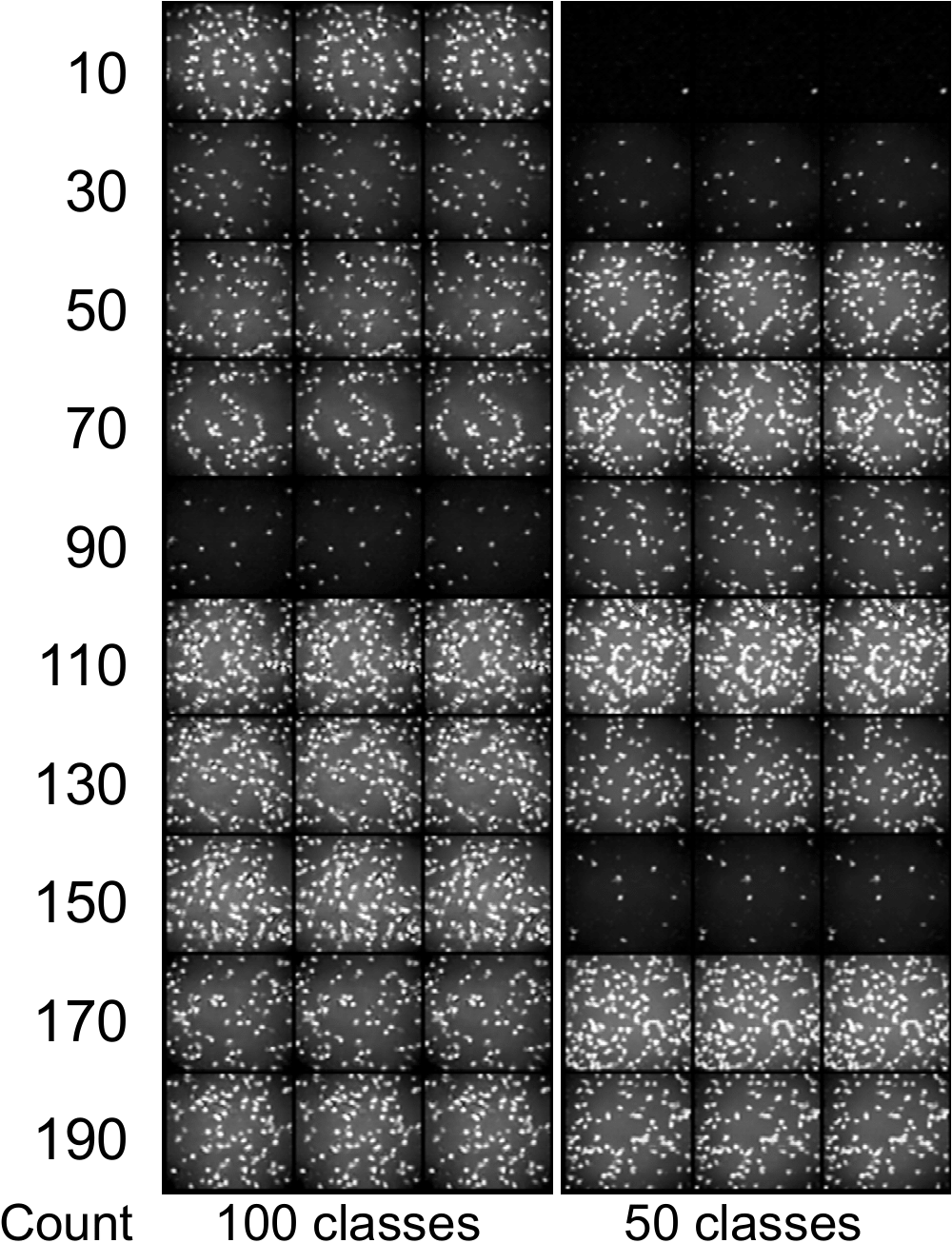}
		\caption{Example Cell-200 fake images from cGAN when we bin the range of cell count into different number of classes.}
		\label{fig:cell200_cGAN_bin_comparison}
	\end{figure}

	In this experiment, we experimented with two different bin setting -- grouping labels into 100 classes and 50 classes, respectively. Experimental results are shown in Fig. \ref{fig:cell200_cGAN_bin_comparison}.  We observe both cGANs fail in this experiment. First of all, cGANs still suffer from the mode collapse problems. Besides, cell counts of generated images do not match those of their given labels.
	
	\section{More details of the experiment on the Low-resolution Steering Angle dataset in Section \ref{sec:steeringangle}}\label{supp:details_of_steeringangle}
	
	\subsection{Description of Steering Angle}\label{supp:steeringangle_data}
	
\begin{figure}[!htbp]
		\centering
		\includegraphics[width=0.5\textwidth]{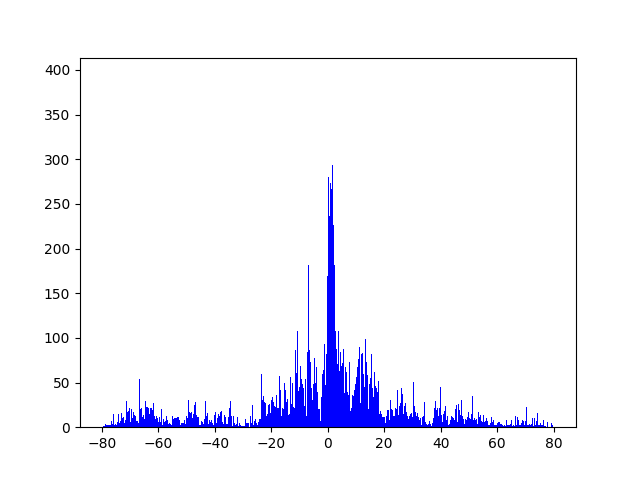}
		\caption{The histogram of the Steering Angle dataset with steering angles varying from $-80^{\circ}$ to $80^{\circ}$. At many angles, we only have 1 or 2 images.}
		\label{fig:steeringangle_histogram}
	\end{figure}

	The \textit{Steering Angle} dataset is a subset of an autonomous driving dataset \cite{steeringangle, steeringangle2}. Steering Angle consists of 12,271 RGB images with 1,904 distinct steering angles ranging from $-80^\circ$ to $80^\circ$. We resize all images to $64\times 64$. The histogram of the steering angles in this dataset is show in Fig.\ \ref{fig:steeringangle_histogram}.
	
	\subsection{Network architectures}\label{supp:steeringangle_nets}
	The network architectures used in this experiment is similar to those in the RC-49 and UTKFace experiments. Please refer to our codes for more details about the network specifications. 
	
	\subsection{Training setups}\label{supp:steeringangle_training_setups}
	
	The cGAN and CcGAN are trained for 20,000 iterations on the training set with the Adam \cite{KingmaB14} optimizer (with $\beta_1=0.5$ and $\beta_2=0.999$) and a constant learning rate $10^{-4}$. The rule of thumb formalue in Remark \ref{rmk:rule_of_thumb} are used to select the hyperparameters for HVDL and SVDL, where we let $m_{\kappa}=5$. 
	
	Please note that, similar to the Cell-200 experiment, we use different batch sizes for the generator and discriminator in four CcGAN methods. The batch size is set to 64 and 512 respectively for the discriminator and generator in CcGAN. Please refer to Supp. \ref{supp:cell200_training_setups} for the reason.
	
	Please see our codes for more details of the training setups.
	
	\subsection{Testing setups}\label{supp:steeringangle_testing_setups}
	
	At the testing stage, we first set 2,000 evenly spaced evaluation labels in $[-80^\circ, 80^\circ]$ and we ask each GAN model to generate 50 images conditional on each evaluation label. 
	
	\subsection{Performance measures}\label{supp:steeringangle_performance}
	
	The quality of generated images from each GAN is evaluated by SFID, NIQE, Diversity, and Label Score.
	
	\begin{itemize}
		\item \textbf{SFID}: To computing SFID, we preset 1,000 SFID centers $[-80^\circ, 80^\circ]$ and let $r_\text{SFID}=2^\circ$. These SFID centers and $r_\text{SFID}=2$ define 1,000 joint SFID intervals. We compute one FID between real and fake images with labels in this interval. We report the mean (i.e., SFID) and standard deviation of these FIDs for each GAN. 
		
		\item \textbf{NIQE} \cite{mittal2012making}: Different from previous three experiments, we train only one NIQE model by using all real images in the Steering Angle dataset since this dataset is highly imbalanced. In the evaluation, we compute one NIQE score for each SFID interval. The reported NIQE score in Table\ \ref{tab:low_resolution_results} is the mean of these NIQE scores.  
		
		\item \textbf{Diversity}: The original autonomous driving dataset \cite{steeringangle, steeringangle2} does not have class labels. To compute Diversity, we manually group the images in Steering Angle into five categories according to their background objects or the types of the road in the images. The five groups are labeled respectively by tree, tree+barrier, bush, bush+barrier, and winding mountain road. Some example images for these five groups are show in Fig.\ \ref{fig:SteeringAngle_five_scenes}. Images in the tree group all have trees in the background. Images in the tree+barrier group all have trees and barriers in the background. Images in the bush group all have bushes in the background. Images in the bush+barrier group all have bushes and barriers in the background. Images in the winding mountain road group all correspond to the scenes of winding mountain roads. A classification-oriented ResNet-34 is trained to classify images from these five groups, and then the Diversity score can be computed based on the entropies of predicted scenes in each SFID interval. 
		
		\begin{figure}[!htbp]
			\centering
			\includegraphics[width=0.6\textwidth]{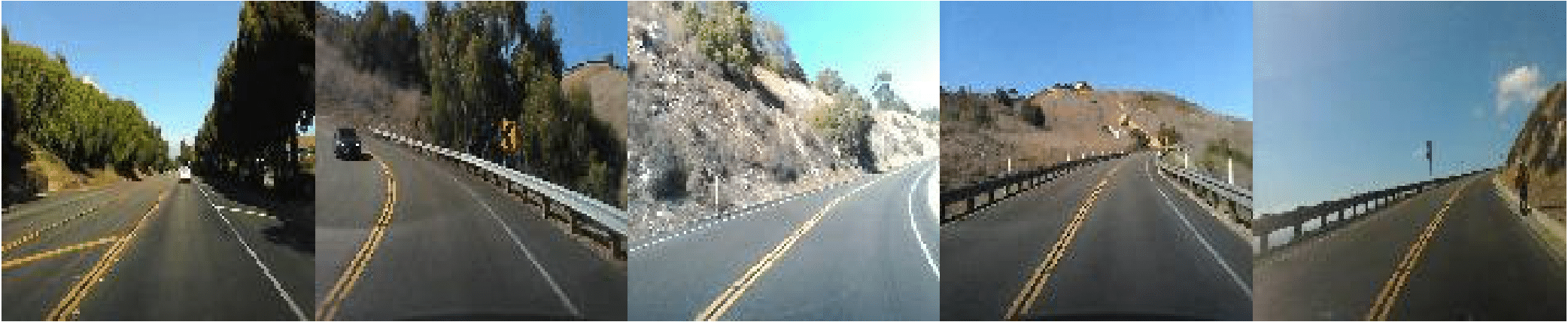}
			\caption{Example Steering Angle images from 5 scenes, i.e., tree, tree+barrier, bush, bush+barrier, and winding mountain road (from left to right).}
			\label{fig:SteeringAngle_five_scenes}
		\end{figure}
		
		\item \textbf{Label Score}: Similar to previous experiments, we pre-train a regression-oriented ResNet-34 to predict the angle for each fake image, and then computes Label Score. 
		
		Please note that, when plotting the line graph of Label Score versus SFID Center in Fig\ \ref{fig:SteeringAngle_line_graphs}, one Label Score is computed for each SFID interval. 
	\end{itemize}

	\subsection{Example Steering Angle images}\label{supp:SteeringAngle_more_example_images}
\begin{figure}[!htbp]
		\centering
		\includegraphics[width=0.7\linewidth]{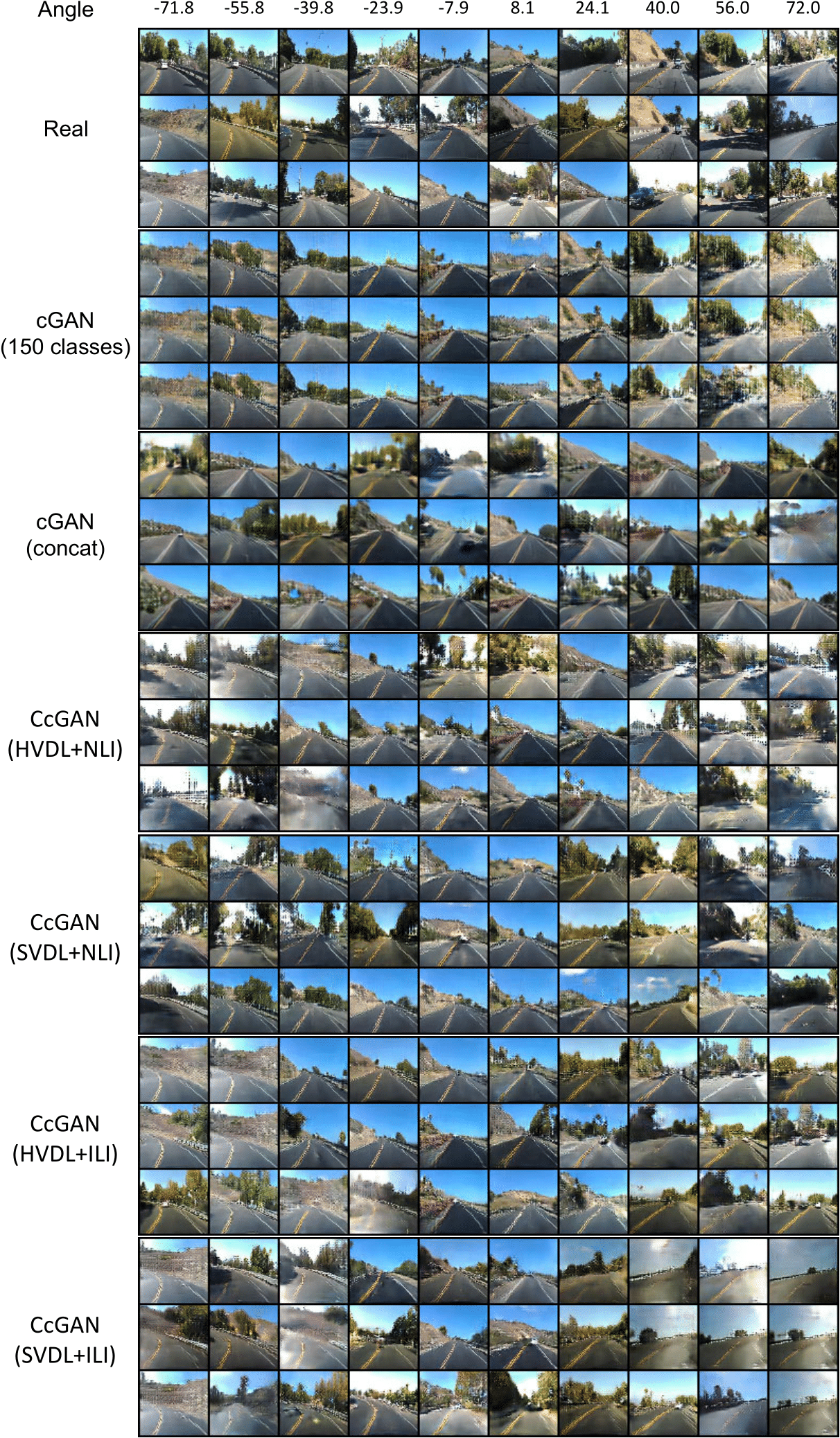} 
		\caption{Three Steering Angle images in $64\times 64$ resolution for each of 10 angles: real images and example fake images from cGAN and four proposed CcGANs, respectively. cGAN has severe mode collapse problem in this experiment. Two NLI-based CcGANs do not work well but two ILI-baesd CcGANs produce images with \textbf{higher visual quality and more diversity}. }
		\label{fig:SteeringAngle_visual_results}
	\end{figure}

	Example Steering Angle images are shown in Fig.\ \ref{fig:SteeringAngle_visual_results}.
	
	\subsection{Extra experiments}\label{supp:steeringangle_extra_experiment}
	
	\subsubsection{Interpolation}\label{supp:steeringangle_CcGAN_interpolation}
	
	In this section, for each CcGAN method, we fix the noise vector $\bm{z}$ but vary the regression label $y$ from $-71.8^\circ$ to $72^\circ$. We can see the road in the image gradually changes from a left turn to a right turn.
	
	\begin{figure}[!ht]
		\centering
		\includegraphics[width=0.6\textwidth]{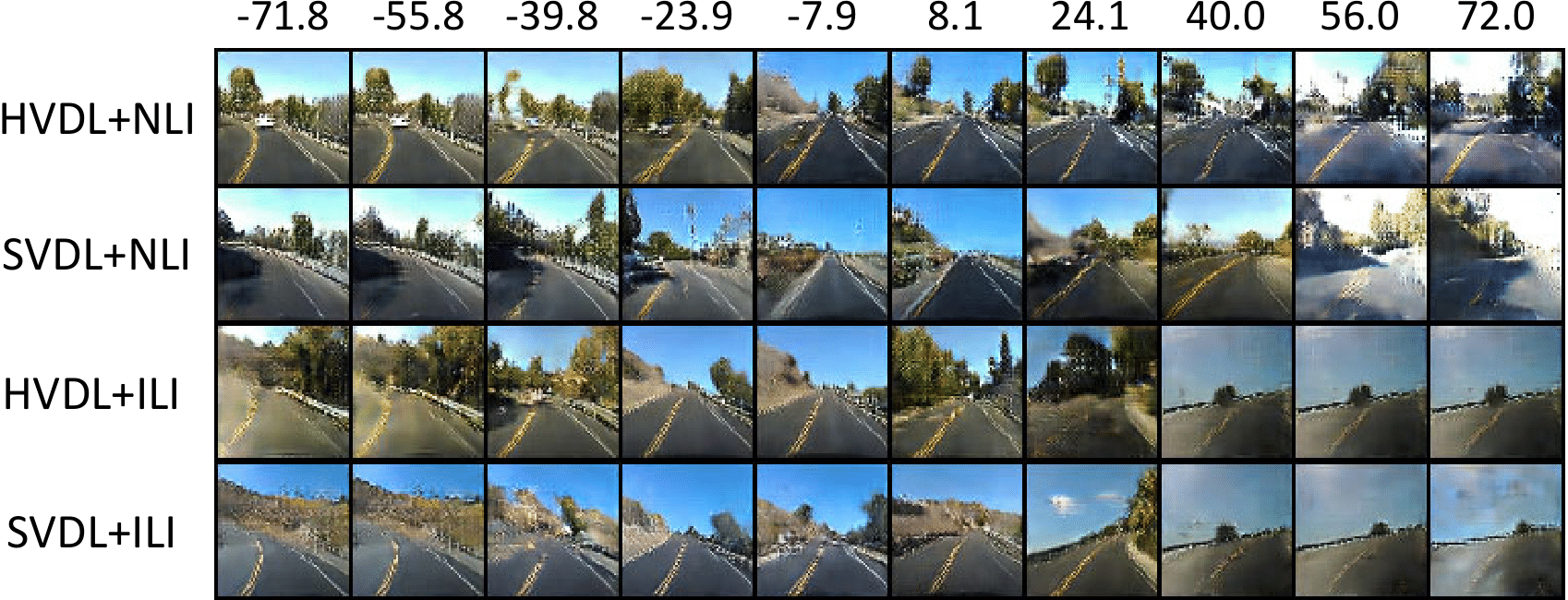}
		\caption{Some examples of generated Steering Angle images from the four CcGAN methods. We fix the noise $\bm{z}$ but vary the label $y$ from $-71.8^\circ$ to $72^\circ$.}
		\label{fig:SteeringAngle_fix_z_continuous_label}
	\end{figure}

	\subsubsection{cGAN: different number of classes}\label{supp:steeringangle_cGAN_bin_comparison}
	
\begin{figure}[!h]
		\centering
		\includegraphics[width=0.8\textwidth]{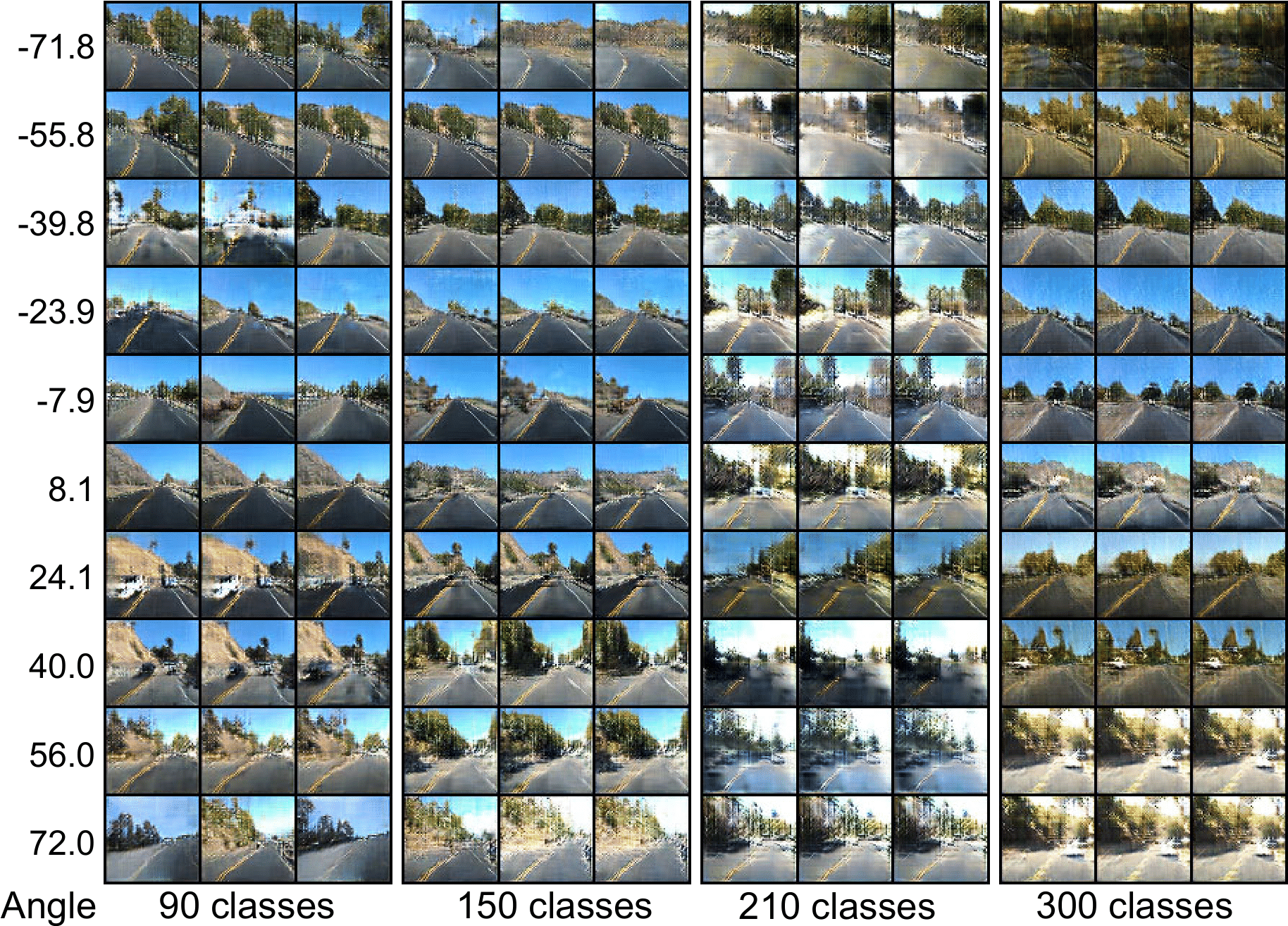}
		\caption{Example Steering Angle fake images from cGAN when we bin the range of steering angles into different number of classes.}
		\label{fig:SteeringAngle_cGAN_bin_comparison}
	\end{figure}

	In this experiment, we experimented with three different bin setting -- grouping labels into 90 classes, 150 classes, and 210 classes, respectively. Experimental results are shown in Fig. \ref{fig:SteeringAngle_cGAN_bin_comparison}.  We observe that different bin settings cannot improve cGAN's performance.
	
	\section{Evaluation Results of Low-resolution Experiments under FID and IS}\label{supp:exp_results_fid_is}
	
	Inception Score (IS) \cite{salimans2016improved} and Fr\'echet Inception Distance \cite{heusel2017gans}, originally proposed for unconditional image generation, are not appropriate overall metrics for our experiment. Consistent with the evaluation of cGANs in \cite{miyato2018cgans,SAGAN-zhang19d,devries2019evaluation}, a conditional generative model in our task needs to be evaluated from three perspectives: (1) visual quality, (2) intra-label diversity (the diversity of images with the same label), and (3) label consistency (whether the labels used as conditions are consistent with the true labels of fake images). (The labels used as conditions are also called assigned labels in our paper.) A conditional generative model's performance in these three perspective partially reflects its conditional density estimation quality. Since IS and FID are developed initially to evaluate images generated from unconditional GANs, they aim to assess the visual quality and \textbf{marginal} diversity of fake images, partially reflecting the marginal density estimation quality. Because computing IS and FID does not need the true and assigned labels of fake images, IS and FID cannot measure intra-label diversity and label consistency. For example, assume we have some fake images with high intra-label diversity and label consistency. We may dramatically degenerate these fake images' intra-label diversity and label consistency by shuffling their assigned labels. However, the IS and FID scores of these fake images won't change because assigned labels are not used in computing IS and FID scores. Furthermore, in our task, a model with high IS and low FID scores may still fail in our task since it may have low intra-label diversity or low label consistency. For example, although cGAN (concat) have better diversity and higher visual quality than cGAN ($K$ classes) does, their label consistency scores are terrible, implying their failure in our task. Therefore, IS and FID are not appropriate overall metrics for our task. To comprehensively evaluate cGAN-generated images, \cite{miyato2018cgans} proposes Intra-FID, which computes FID separately at each of the distinct labels and reports the average FID score. We further extend Intra-FID by SFID to the scenario with insufficient real images. Besides Intra-FID and SFID, we also use three separate metrics (NIQE, Diversity, and Label Score) to evaluate the visual quality, intra-label diversity, and label consistency, respectively. The evaluation in terms of these three individual metrics is often consistent with that based on the overall metric (i.e., Intra-FID or SFID) in our experiments. Therefore, Intra-FID and SFID are taken as the overall metric in our task.
	
	For completeness, we also report them in this appendix. Since our datasets are quite different from ImageNet \cite{deng2009imagenet}, we train a classification ResNet-34 and an autoencoder (the one used for computing Intra-FID) from scratch on our datasets to compute IS and FID respectively. Table \ref{tab:exp_64x64_IS_and_FID} summarizes the evaluation results of all candidate methods on four low-resolution datasets in terms of IS and FID. We can see CcGAN still outperforms two cGANs, even though IS and FID are not appropriate overall metrics for our task.
	
	\begin{table}[!htbp]
		\centering
		\caption{IS and FID scores of all candidates methods in $64\times 64$ experiments.}
		\begin{adjustbox}{width=0.8\textwidth}
			\begin{tabular}{c|cc|cc|cc|cc}
				\toprule
				& \multicolumn{2}{|c|}{RC-49} & \multicolumn{2}{c|}{UTKFace} & \multicolumn{2}{c|}{Cell-200} & \multicolumn{2}{c}{Steering Angle} \\
				\hline
				Method & IS $\uparrow$ & FID $\downarrow$  & IS $\uparrow$ & FID $\downarrow$ & IS $\uparrow$ & FID $\downarrow$  & IS $\uparrow$ & FID $\downarrow$ \\
				\hline
				cGAN ($K$ classes) & $2.382$ & 1.066 & $2.636$ & 0.963 & -     & 30.086 & $2.572$ & 0.976 \\
				cGAN (concat) & $11.440$ & 0.295 & $3.103$ & 0.465 & -     & 37.689 & $3.251$ & 0.255 \\
				CcGAN (HVDL+NLI) & $14.730$ & 0.285 & $3.328$ & 0.114 & -     & 40.279 & $3.587$ & 0.316 \\
				CcGAN (SVDL+NLI) & $19.425$ & 0.207 & $3.307$ & 0.087 & -     & 51.318 & $3.968$ & \textbf{0.212} \\
				CcGAN (HVDL+ILI) & $17.992$ & 0.213 & $3.256$ & \textbf{0.056} & -     & 3.263 & $\bm{4.592}$ & 0.327 \\
				CcGAN (SVDL+ILI) & $\bm{20.173}$ & \textbf{0.197} & $\bm{3.382}$ & 0.142 & -     & \textbf{1.684} & $4.439$ & 0.331 \\
				\bottomrule
			\end{tabular}%
		\end{adjustbox}
		\label{tab:exp_64x64_IS_and_FID}%
	\end{table}%

	\section{Comparison against state of the art cGANs}\label{supp:comparison_against_sota}
	All class-conditional GANs and CcGAN are trained for 30,000 iterations with batch size 256. Except CR-BigGAN and ReACGAN, when training all cGANs, we update the discriminator network twice in each iteration. 
	
	To implement the class-conditional SNGAN, we use the vanilla cGAN loss~\cite{mirza2014conditional} instead of the hinge loss~ \cite{lim2017geometric} because hinge loss results in mode collapse in this setting. 
	
	To implement CR-BigGAN~\cite{Zhang2020Consistency} and ReACGAN~\cite{reacgan2021}, we borrow codes of StudioGAN from \url{https://github.com/POSTECH-CVLab/PyTorch-StudioGAN}. The training setups for CR-BigGAN and ReACGAN are mainly based on the configuration file of StudioGAN designed for $64\times 64$ TinyImageNet. Please note that ReACGAN~\cite{reacgan2021} is published more than one year after our submission of CcGAN to ICLR 2021. 
	
	DiffAugment~\cite{zhao2020differentiable} with the strongest transformation combination (Color + Translation + Cutout) is also used in some GAN training including BigGAN+DiffAugment (class-conditional), SAGAN+DiffAugment (CcGNA), and BigGAN+DiffAugment (CcGAN).
	
	Other setups are similar to Supp.~\ref{supp:details_of_rc49}. Please see our code in ``.\textbackslash RC-49\textbackslash RC-49\_64x64\_extra" for more details. 
	
	\section{More details of the high-resolution experiments in Section \ref{sec:experiment_hd}}\label{supp:details_of_experiment_hd}
	
	\subsubsection{Reformulated hinge loss}\label{supp:reformulated_hinge_loss}
	
	Our CcGAN (SVDL+ILI) in the high-resolution experiments is trained with a reformulated hinge loss shown as follows.
	\begin{equation}
		\label{eq:SVDL_hinge}
		\begin{aligned}
			\widehat{\mathcal{L}}^{\text{SVDL}}(D) = & - \frac{C_7}{N^r}\sum_{j=1}^{N^r}\sum_{i=1}^{N^r}\mathbb{E}_{\epsilon^r\sim\mathcal{N}(0,\sigma^2)}\left[W_3\cdot\min( 0, -1+D(\bm{x}_i^r, y_j^r+\epsilon^r) ) \right]\\
			&- \frac{C_8}{N^g}\sum_{j=1}^{N^g}\sum_{i=1}^{N^g}\mathbb{E}_{\epsilon^g\sim\mathcal{N}(0,\sigma^2)}\left[ W_4\cdot \min( 0,-1-D(\bm{x}_i^g, y_j^g+\epsilon^g) ) \right],
		\end{aligned}
	\end{equation}
	where $\epsilon^r\triangleq y-y_j^r$, $\epsilon^g\triangleq y-y_j^g$,
	\begin{align}
		W_3=\frac{w^r(y_i^r,y_j^r+\epsilon^r)}{\sum_{i=1}^{N^r}w^r(y_i^r,y_j^r+\epsilon^r)},\quad
		W_4=\frac{w^g(y_i^g,y_j^g+\epsilon^g)}{\sum_{i=1}^{N^g}w^g(y_i^g,y_j^g+\epsilon^g)}, \nonumber
	\end{align}
	and $C_7$ and $C_8$ are some constants.

	\subsubsection{High-resolution RC-49}\label{supp:rc-49_hd_setups}
	
\begin{figure}[!htbp]
		\centering
		\includegraphics[width=1\textwidth]{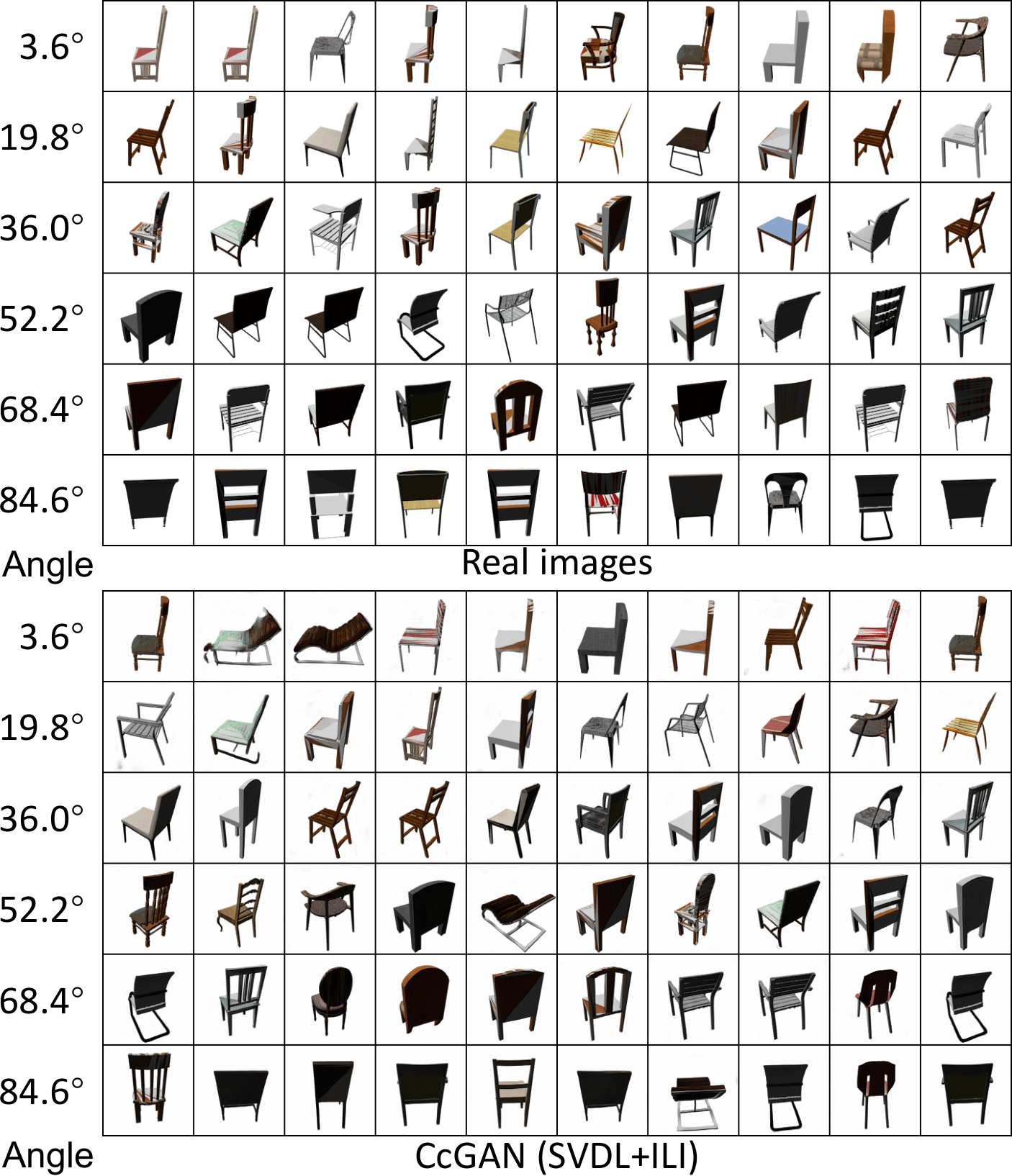}
		\caption{Some example real RC-49 images and fake RC-49 images from CcGAN (SVDL+ILI) in the $128 \times 128$ resolution. We can see CcGAN can generate visually realistic, diverse and label consistent images.}
		\label{fig:rep_RC-49_HD_overall_comparision_part_1}
	\end{figure}

\begin{figure}[!htbp]
		\centering
		\includegraphics[width=1\textwidth]{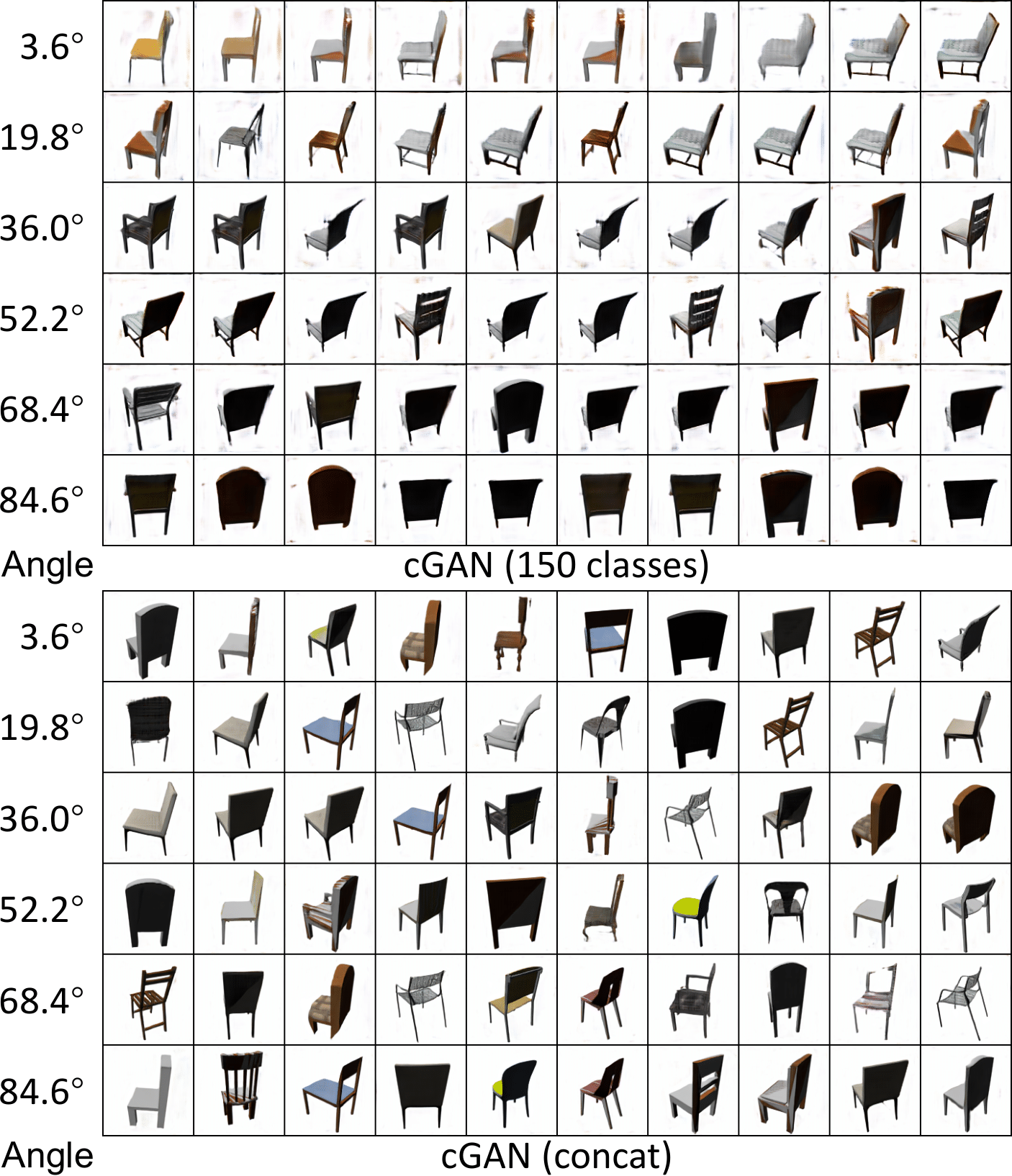}
		\caption{Some example fake RC-49 images from cGAN (150 classes) and cGAN (concat) in the $128 \times 128$ resolution. They show two types of failures of conventional cGANs. cGAN (150 classes) has high label consistency but low visual quality and low intra-label diversity. cGAN (concat) has high intra-label diversity and fair visual quality but low label consistency.}
		\label{fig:rep_RC-49_HD_overall_comparision_part_2}
	\end{figure}

	In the high-resolution experiment, we test CcGAN (SVDL+ILI), cGAN (150 classes), and cGAN (concat) on RC-49 with two resolutions, i.e., $128\times 128$ and $256\times 256$. We use SAGAN \cite{SAGAN-zhang19d} as the backbone for all candidates. We also use hinge loss \cite{lim2017geometric} to train cGAN (150 classes) and cGAN (concat), and Eq.\ \eqref{eq:SVDL_hinge} to train CcGAN (SVDL+ILI). DiffAugment \cite{zhao2020differentiable} with the strongest transformation combination (Color + Translation + Cutout) is also used in all GAN training. DiffAugment substantially alleviates the mode collapse problem of cGAN ($K$ classes) on RC-49. When training each candidate GAN, at each iteration, we update the discriminator twice while update the generator once. For $128\times 128$ experiment, the batch size is 256. For $256\times 256$ experiment, the batch size is set 128. The rest experimental setups are consistent with the low-resolution experiment. Some example images in the $128\times 128$ resolution for this experiment are shown in Figs.\ \ref{fig:rep_RC-49_HD_overall_comparision_part_1} and \ref{fig:rep_RC-49_HD_overall_comparision_part_2}.

	\subsubsection{High-resolution UTKFace}\label{supp:utkface_hd_setups}
	
\begin{figure}[!htbp]
		\centering
		\includegraphics[width=1\textwidth]{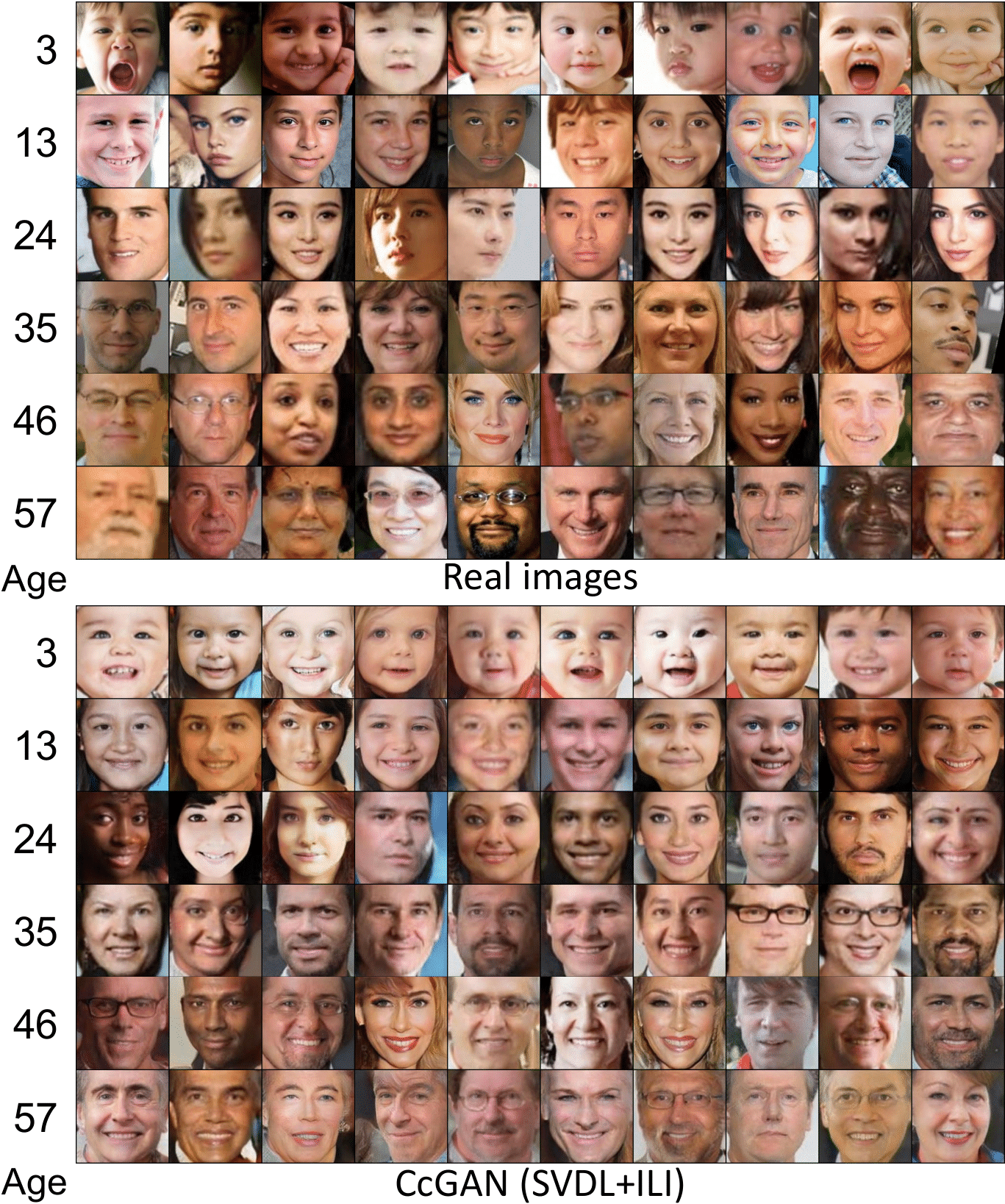}
		\caption{Some example real UTKFace images and fake UTKFace images from CcGAN (SVDL+ILI) in the $192\times 192$ resolution. We can see CcGAN can generate visually realistic, diverse and label consistent images.}
		\label{fig:rep_UTKFace_HD_overall_comparision_part_1}
	\end{figure}

\begin{figure}[!htbp]
		\centering
		\includegraphics[width=1\textwidth]{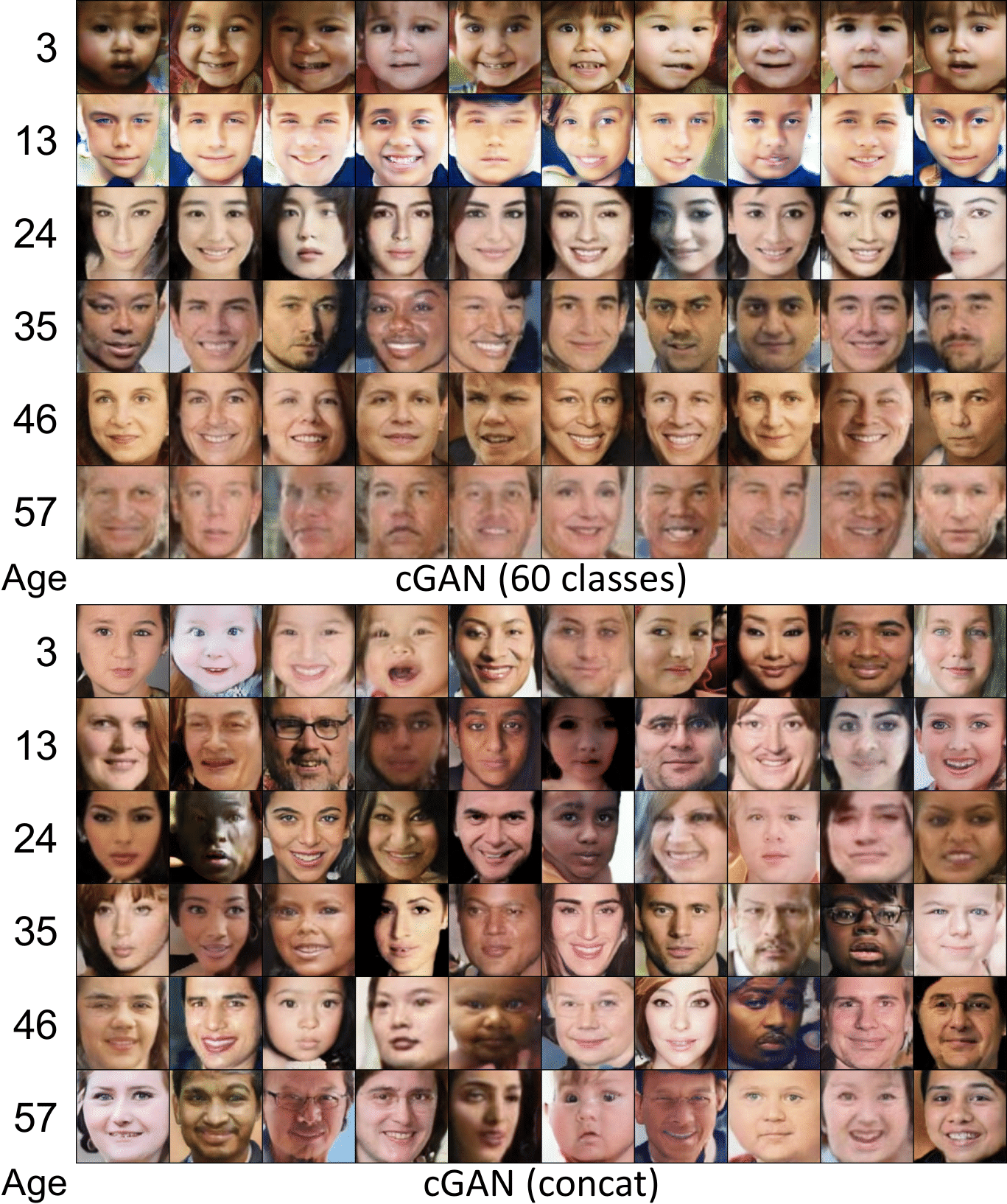}
		\caption{Some example fake UTKFace images from cGAN (60 classes) and cGAN (concat) in the $192\times 192$ resolution. They show two types of failures of conventional cGANs. cGAN (60 classes) has high label consistency but low visual quality (e.g., last row) and low intra-label diversity (e.g., second row only has boys). cGAN (concat) has high intra-label diversity and moderate visual quality but low label consistency (e.g., first row has many adults).}
		\label{fig:rep_UTKFace_HD_overall_comparision_part_2}
	\end{figure}

	In the high-resolution experiment, we test CcGAN (SVDL+ILI), cGAN (60 classes), and cGAN (concat) on UTKFace with two resolutions, i.e., $128\times 128$ and $192\times 192$. We use SAGAN \cite{SAGAN-zhang19d} as the backbone for all candidates. We also use hinge loss \cite{lim2017geometric} to train cGAN (150 classes) and cGAN (concat), and Eq.\ \eqref{eq:SVDL_hinge} to train CcGAN (SVDL+ILI). DiffAugment \cite{zhao2020differentiable} with the strongest transformation combination (Color + Translation + Cutout) is also used in all GAN training. DiffAugment substantially alleviates the mode collapse problem of cGAN ($K$ classes) on UTKFace. When training each candidate GAN, at each iteration, we update the discriminator four times while update the generator once. For $128\times 128$ experiment, the batch size is 256. For $192\times 192$ experiment, the batch size is set 96. We also use $\nu=900$ for the CcGAN training. The rest experimental setups are consistent with the low-resolution experiment. Some example images in the $192\times 192$ resolution for this experiment are shown in Figs.\ \ref{fig:rep_UTKFace_HD_overall_comparision_part_1} and \ref{fig:rep_UTKFace_HD_overall_comparision_part_2}.

	\subsubsection{High-resolution Steering Angle}\label{supp:steering_angle_hd_setups}
	
\begin{figure}[!htbp]
		\centering
		\includegraphics[width=1\textwidth]{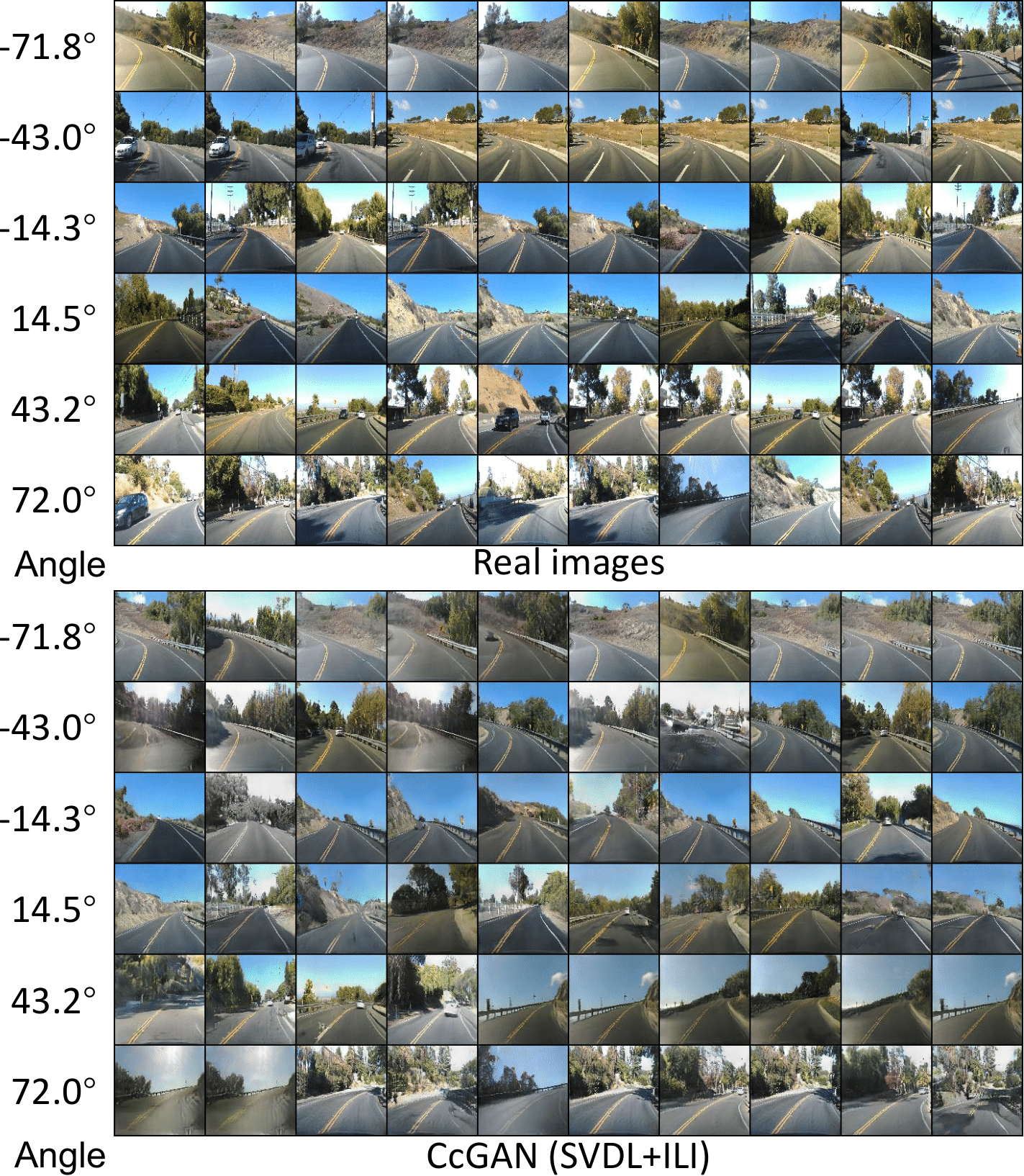}
		\caption{Some example real Steering Angle images and fake Steering Angle images from CcGAN (SVDL+ILI) in the $128 \times 128$ resolution. We can see CcGAN can generate visually realistic, diverse and label consistent images.}
		\label{fig:rep_SteeringAngle_HD_overall_comparision_part_1}
	\end{figure}

\begin{figure}[!htbp]
		\centering
		\includegraphics[width=1\textwidth]{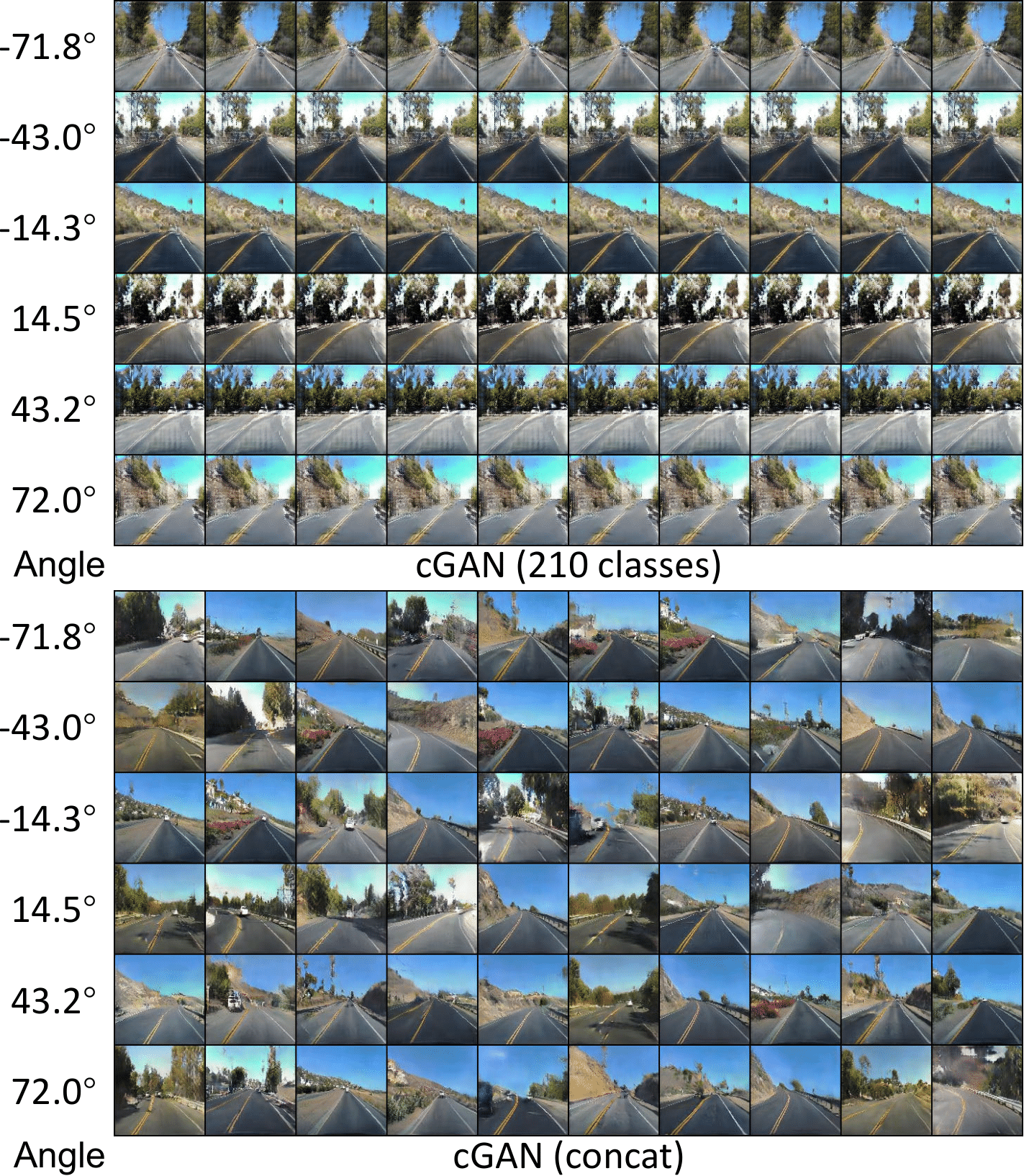}
		\caption{Some example fake Steering Angle images from cGAN (210 classes) and cGAN (concat) in the $128 \times 128$ resolution. They show two types of failures of conventional cGANs. cGAN (150 classes) has high label consistency but fair visual quality and low intra-label diversity. cGAN (concat) has high intra-label diversity and moderate visual quality but low label consistency.}
		\label{fig:rep_SteeringAngle_HD_overall_comparision_part_2}
	\end{figure}

	In the high-resolution experiment, we test CcGAN (SVDL+ILI), cGAN (210 classes), and cGAN (concat) on Steering Angle in $128\times 128$ resolution. We use SAGAN \cite{SAGAN-zhang19d} as the backbone for all candidates. We also use hinge loss \cite{lim2017geometric} to train cGAN (150 classes) and cGAN (concat), and Eq.\ \eqref{eq:SVDL_hinge} to train CcGAN (SVDL+ILI). DiffAugment \cite{zhao2020differentiable} with the strongest transformation combination (Color + Translation + Cutout) is also used in all GAN training. In this experiment, even with DiffAugment, cGAN ($K$ classes) still has the mode collapse problem. When training each candidate GAN, at each iteration, we update the discriminator twice while update the generator once. The batch size is set 256. The rest experimental setups are consistent with the low-resolution experiment. Some example images in the $128\times 128$ resolution for this experiment are shown in Figs.\ \ref{fig:rep_SteeringAngle_HD_overall_comparision_part_1} and \ref{fig:rep_SteeringAngle_HD_overall_comparision_part_2}.

\end{document}